\documentclass[twoside,11pt]{article}

\usepackage{jmlr2e}

\usepackage{graphicx,subfigure}
\usepackage{pstricks,pst-node,pst-tree}
\usepackage{subfloat}
\usepackage{booktabs}
\usepackage{lscape}
\usepackage{rotating}
\usepackage{tabls}
\usepackage{color}
\usepackage{subfigure}
\usepackage{url}
\usepackage{amsmath}
\usepackage{color}
\usepackage{bigstrut}
\usepackage{longtable}
\usepackage{supertabular}
\usepackage{longtable,pdflscape}
\usepackage{tabu}
\usepackage{longtable}
\usepackage{pdflscape}

\usepackage[colorinlistoftodos,textwidth=3cm]{todonotes}

\usepackage{ntheorem}

\def\|{\,|\,}

\def\obj{{\bf x}}

\def\eqref#1{Eq~\ref{#1}}

\def\P{{\rm P}}

\def\obj{{\bf x}}

\def\data{\ifmmode \mathcal D\else$\data$\fi}
\def\labels{\ifmmode \mathcal L\else$\labels$\fi}
\def\card{\ifmmode \mathcal X\else$\card$\fi}

\def\k{\ifmmode \|\!\mathcal{Y}\!\| \else$\k$\fi}

\def\train{\ifmmode \mathcal T\else$\train$\fi}
\def\test{\ifmmode \mathcal U\else$\train$\fi}
\def\model{\ifmmode \mathcal M\else$\model$\fi}

\def\P{{\rm P}}
\def\ANDE^#1{\mathop{{\rm A}#1{\rm DE}}}

\def\eqref#1{Eq~\ref{#1}}

\def\ANN{{\textrm{ANN}^0}}
\def\ANNd{{\textrm{ANN}^0 \textrm{(d)}}}

\def\dLC{{ \mathop{ \textrm{LC}^{\textrm{d}}}}}
\def\LC{{ \mathop{ \textrm{LC}}}}

\def\s{{\bf s}}

\def\|{\,|\,}

\def\obj{{\bf x}}

\jmlrheading{01}{2017}{1-28}{01/17; Revised 00/00}{00/00}{Nayyar~A.\ Zaidi, Yang.\ Du, Geoffrey~I.\ Webb}

\ShortHeadings{Discretizing Quantitative Attributes in Linear Classifiers}{Zaidi, Du, Webb}
\firstpageno{1}

\begin{document}

\title{On the Effectiveness of Discretizing quantitative Attributes in Logistic Regression for Big Datasets}
\title{Discretizing of quantitative Attributes in Logistic Regression for Big Datasets}
\title{On the Effectiveness of Discretization in Logistic Regression for Big Datasets}
\title{Discretization in Logistic Regression for Big Datasets}

\title{On the Effectiveness of Discretizing quantitative Attributes in a Linear Classifier for Big Datasets}
\title{On the Effectiveness of Discretization in a Linear Classifier for Big Datasets}
\title{Discretization in Linear Classifiers for Big Datasets}
\title{On the Effectiveness of Discretization in Linear Classifiers for Big Datasets}
\title{Discretization, Linear Classifiers and Big Data}
\title{On the Effectiveness of Discretizing Quantitative Attributes in a Linear Classifier for Big Datasets}
\title{On the Effectiveness of Discretizing Quantitative Attributes in Linear Classifiers}

\author{\name Nayyar~A.~Zaidi \email nayyar.zaidi@monash.edu \\
\name Yang Du \email ydu32@student.monash.edu \\  
\name Geoffrey~I.~Webb \email geoff.webb@monash.edu \\  
       \addr Faculty of Information Technology \\ Monash University \\ VIC 3800, Australia.      
        }

\editor{xxxxxx xxxxxx}

\maketitle

\begin{abstract}

Learning algorithms that learn linear models often have high representation bias on real-world problems. In this paper, we show that this representation bias can be greatly reduced by discretization. 
Discretization is a common procedure in machine learning that is used to convert a quantitative attribute into a qualitative one. 
It is often motivated by the limitation of some learners to qualitative data. Discretization loses information, as fewer distinctions between instances are possible using discretized data relative to undiscretized data. In consequence, where discretization is not essential, it might appear desirable to avoid it. 
However, it has been shown that discretization often substantially reduces the error of the linear generative Bayesian classifier naive Bayes. 
This motivates a systematic study of the effectiveness of discretizing quantitative attributes for other linear classifiers.
In this work, we study the effect of discretization on the performance of linear classifiers optimizing three distinct discriminative objective functions --- logistic regression (optimizing negative log-likelihood), support vector classifiers (optimizing hinge loss) and a zero-hidden layer artificial neural network (optimizing mean-square-error). 
We show that discretization can  greatly increase the accuracy of these linear discriminative learners by reducing their representation bias, especially on big datasets.
We substantiate our claims with an empirical study on $42$ benchmark datasets.

\end{abstract}


\section{Introduction} \label{sec_intro}

One of the many factors that affect the error of a learning system is its \emph{representation bias} \citep{vanderPutten2004}, or, as it is also called, its \emph{hypothesis language bias} \citep{Mitchell1980}.  We define representation bias herein as the minimum loss of any model in the space of models available to the learner. It is clearly desirable in the general case to use a space of models that minimizes representation bias for a given problem.
 Learning algorithms that use linear models, such as logistic regression (LR) \citep{Murphy:2012:ML} and support vector classifiers (SVC) \citep{liblinear}, are very popular, possibly in part due to their lending themselves to convex optimization.
In this paper we argue that learning algorithms that learn linear models often have high representation bias on real-world problems, and that often this bias can be reduced by discretization. 
\iftrue We illustrate this in Figure~\ref{fig:simple-example} which shows on simple synthetic data how a linear classifier cannot create an accurate classifier on the numeric data but can when the data are discretized using simple univariate discretization.

\begin{figure}
	\centering   
	\hspace{-0.3in}
	\subfigure[Perceptron on original data]{\label{fig_2dperceptron}\includegraphics[width=80mm,height=55mm]{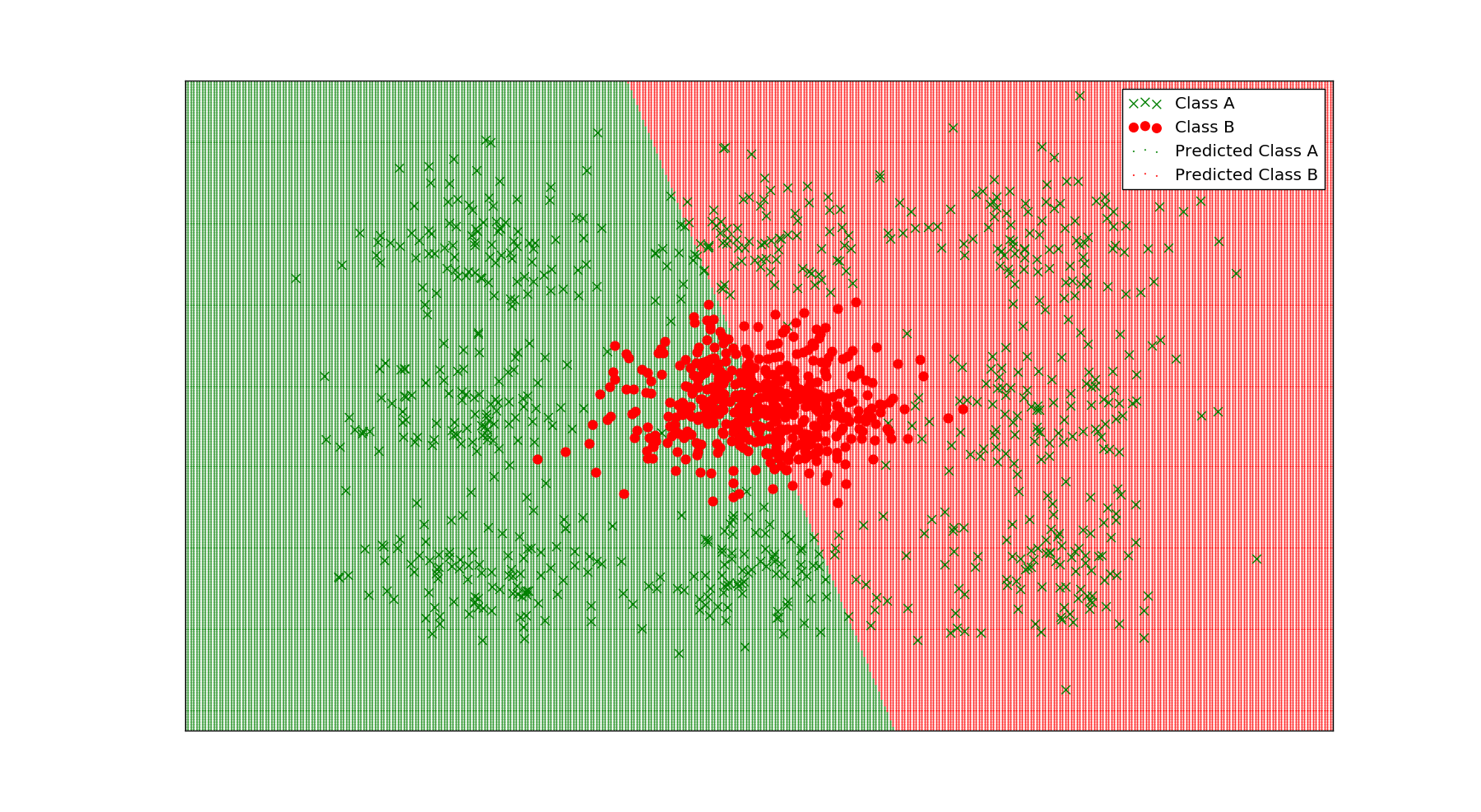}} \hspace{-0.2in}
	\subfigure[Perceptron trained on discretized data. $3$-bin equal frequency discretization is used.]{\label{fig_2dperceptronDisc}\includegraphics[width=80mm,height=55mm]{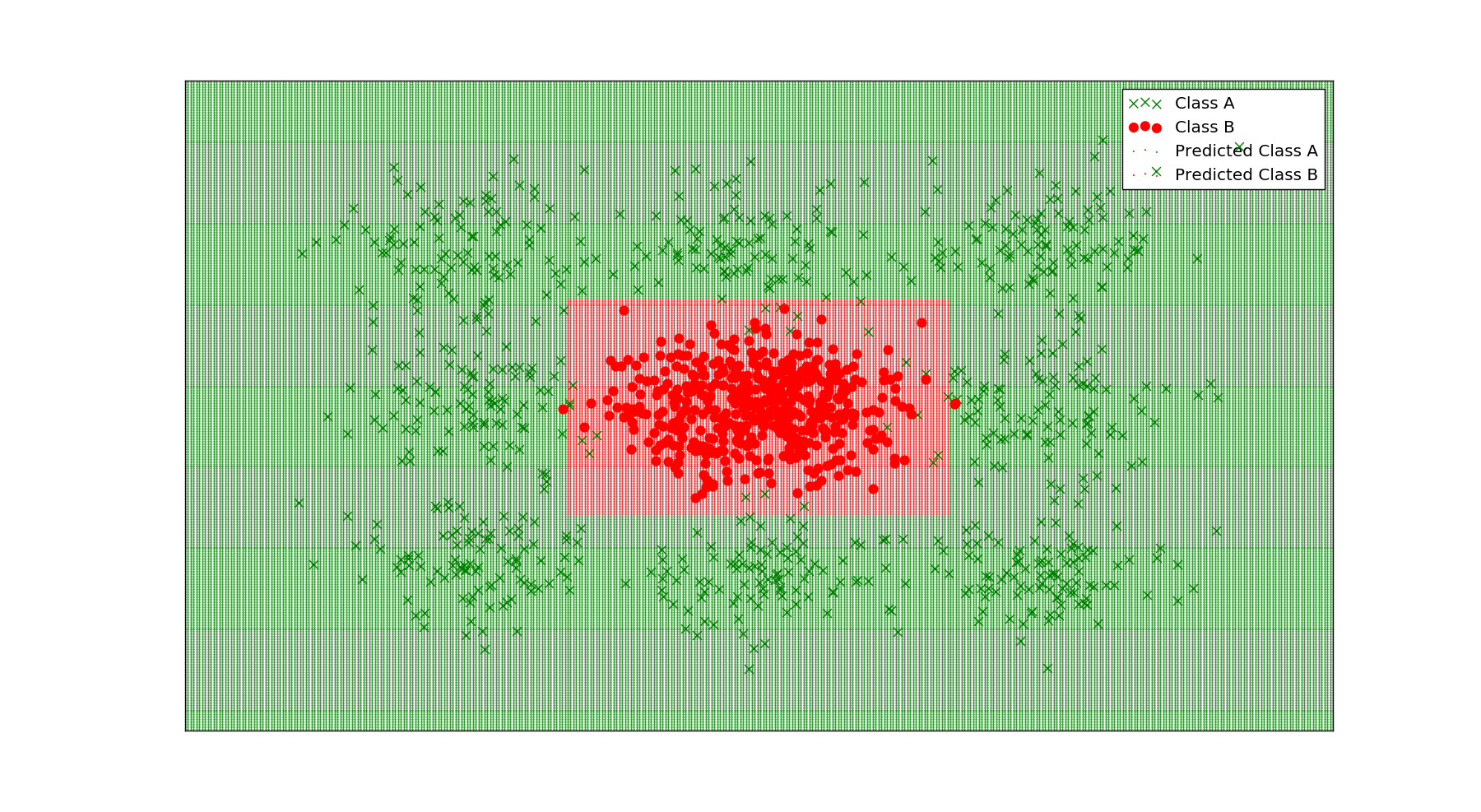}}
	\caption{Illustration of the effectiveness of a perceptron after EMD discretization on  simple two-dimensional contrived data.}
	\label{fig:simple-example}
\end{figure}

\else
 In the following, we will illustrate this point using two simple contrived examples but before that, we will explain one of the oldest linear classifiers that emerged in machine learning research -- the \emph{Perceptron}.

A perceptron is a linear classifier that is guaranteed to converge only when the data is linearly separable~\citep{Rosenblatt1957}. Also, if there are multiple hyper-planes that separate the data, a perceptron is guaranteed to find one of those hyper-planes (note that it might not be the best hyper-plane). However, if the data is not separable, some sort of termination criteria needs to be specified, otherwise, the algorithm will never terminate. 
A Perceptron follows a simple training algorithm: it starts with the zero weights that is $\mathbf{\beta} = 0$. It goes through the dataset one by one. For every data point $\obj$, if the sign of its label (denoted as $y$) is the same as that of $\beta^T \obj$, nothing is done. However, if the signs differ, the weights are updated by the following simple rule: $\beta_i = \beta_i + \eta y x_i$, where $\eta$ is the learning rate (it has similar interpretation as the learning rate of an SGD optimizer).
Perceptron algorithm spiked a lot of research in artificial intelligence, but interest in it was soon lost due to its linear nature. The advent of back-propagation algorithm led to a renewed interest in the field of machine learning sparking a lot of interest in artificial neural networks (ANN). While back-propagation is one way of constructing non-linear decision boundaries, application of feature transformation (or kernel trick) to perceptrons led to a simple kernel perceptron algorithm of~\cite{Aizerman1964}, that successfully could model non-linear decision boundaries. However, it remained less famous through-out 60's, 70's and 80's. 
It is only in early 90's, kernel trick became extremely popular with the introduction of support vector machines.
We conjecture that the reason kernel perceptron algorithm did not became popular is due to its proneness to over-fitting. Early datasets were small, simply taking some of the quadratic combination of features would have resulted in more features than the data points. However, for datasets of today, because of their sheer scale, such transformation should not be any problem~\citep{Sonnenburg2010,zaidi2016ALR}.

Let us discuss the working of a linear classifier (perceptron in this example) on the simplest problem that is not linearly separable. Figure~\ref{fig1} plots the univariate data. The feature (X-axis) is the age and the class is the `susceptibility to some disease' (Y-axis). It can be seen that as the disease affects mostly middle-aged people, the relationship between age and the disease is not linear. Figure~\ref{fig_1dperceptron} shows the decision boundary obtained by training the perceptron algorithm which does a poor job in modeling the non-linearity \footnote{The number of iterations of Perceptron are set to $1000$ in all the experiments in this section. No other terminating criterion is used.}. 
As explained earlier, one can transform the data or use ANN back-propagation algorithm with at least one hidden layer to model this non-linearity. We will discuss these two approaches in the context of second example (multi-variate data).
But first, let us discuss the role of discretization. One can see by plotting the data that there are three distinct boundaries that exists.
Figure~\ref{fig_1dperceptronDisc} shows the decision boundary of the perceptron after data is discretized. In a nutshell, we used $3$-bin equal-frequency discretization, transformed the data and then trained a linear classifier in this transformed space. As can be seen, the resulting perceptron can model the non-linearity perfectly.
\begin{figure}[t] 
\centering   
\hspace{-0.3in}
\subfigure[Perceptron on original data]{\label{fig_1dperceptron}\includegraphics[width=80mm,height=55mm]{images/figure1_1d.png}} \hspace{-0.2in}
\subfigure[Perceptron trained on discretized data. $3$-bin equal frequency discretization is used.]{\label{fig_1dperceptronDisc}\includegraphics[width=80mm,height=55mm]{images/figure2_1dDisc.png}}
\caption{Illustration of the effectiveness of discretization on a simple one-dimensional contrived data.}
\label{fig1}
\end{figure}

Figure~\ref{fig2} illustrates the similar concept in multivariate case. It can be seen that data is not linearly separable -- class B is surrounded by class A. Figure~\ref{fig_2dperceptron} shows the decision boundary obtained by training a simple Perceptron. As in the univariate case, it can be seen that it does a poor job in distinguishing between the two classes. We trained an ANN with one-hidden layer (two nodes) on this dataset. Figure~\ref{fig_2dANN} shows the power of ANN in modeling the non-linearities present in this dataset \footnote{Note, the parameters of an ANN are trained with back-propagation using quasi-Newton (LFBGS) optimization algorithm.}.
Another way of modeling non-linear decision surfaces is using the data transformation. We appended an extra dimension to the data that is $X_1X_2$ and trained the perceptron in this transformed space. The resulting decision boundary is shown in Figure~\ref{fig_2dkernelperceptron}. The non-linear decision surfaces can be seen \footnote{Note our goal with this example is to illustrate that non-linear decision boundaries can be obtained via feature transformation and not the quality of the decision boundary for this problem. Perhaps one need cubic or even higher-order features to obtain the perfect decision surface in this case.}.
Figure~\ref{fig_2dperceptronDisc} shows the perceptron trained after $3$-bin discretization.
Just like the univariate case, it can be seen that the resulting perceptron does produce non-linear decision boundaries and can distinguish between the two classes extremely well. 
\begin{figure}[hbt] 
\centering   
\hspace{-0.3in}
\subfigure[Perceptron on original data]{\label{fig_2dperceptron}\includegraphics[width=80mm,height=55mm]{images/figure2d2_Perceptron.png}} \hspace{-0.2in}
\subfigure[One-hidden layer with two nodes Artificial Neural Network.]{\label{fig_2dANN}\includegraphics[width=80mm,height=55mm]{images/figure2d2_ANN}}

\hspace{-0.3in}
\subfigure[Kernel Perceptron]{\label{fig_2dkernelperceptron}\includegraphics[width=80mm,height=55mm]{images/figure2d2_KPerceptron}} \hspace{-0.2in}
\subfigure[Perceptron trained on discretized data. $3$-bin equal frequency discretization is used.]{\label{fig_2dperceptronDisc}\includegraphics[width=80mm,height=55mm]{images/figure2d2_Disc}}
\caption{Illustration of the effectiveness of perceptron after discretization on a simple two-dimensional contrived data by comparing its decision boundaries with that of ANN, Kernel-Perceptron and perceptron on original data.}
\label{fig2}
\end{figure}
\fi
 
\begin{figure}[t] 
	\centering   
	\hspace{-0.3in}
	\subfigure[Perceptron on original data]{\label{fig3a}\includegraphics[width=80mm,height=55mm]{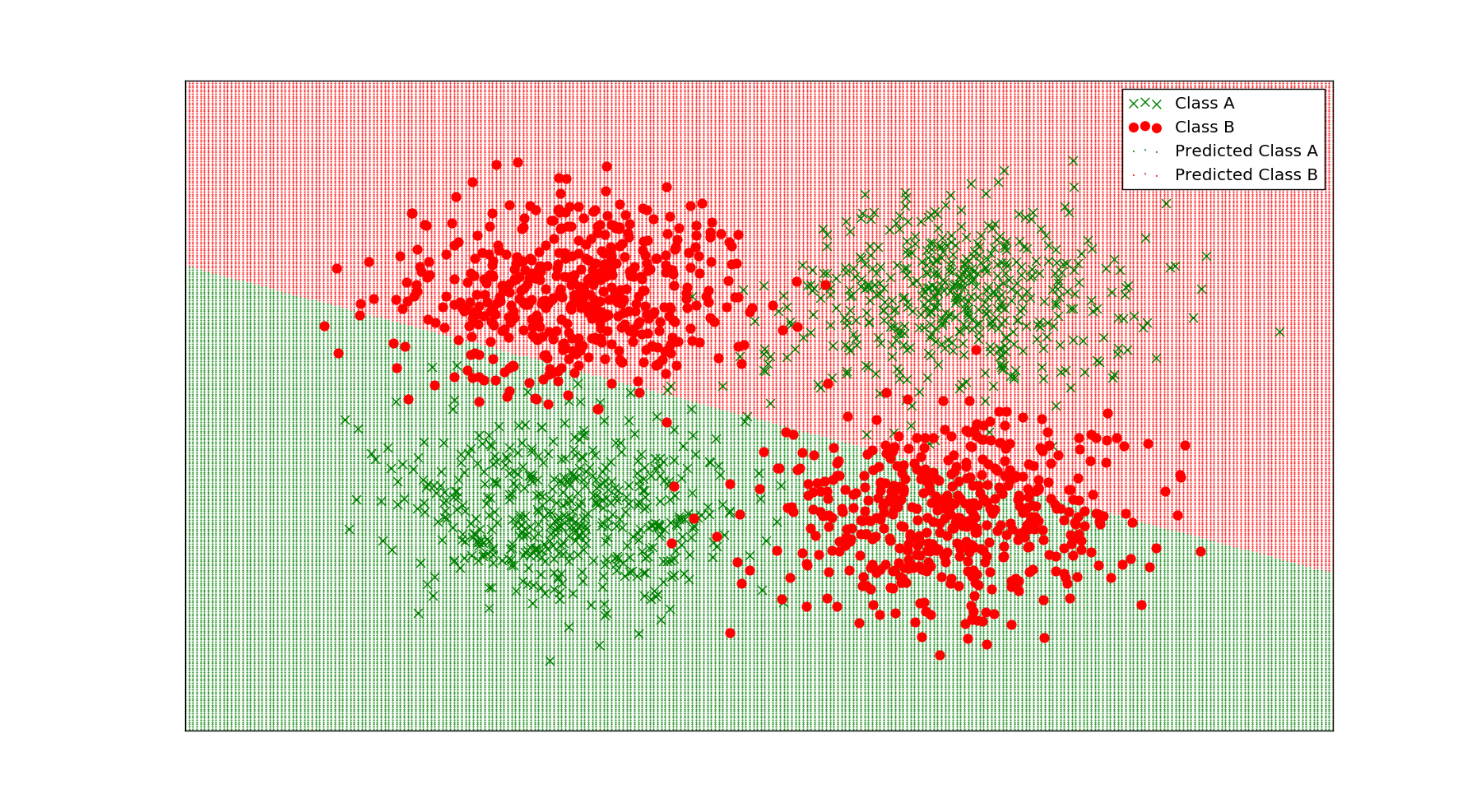}} \hspace{-0.2in}
	\subfigure[Perceptron trained on discretized data. $4$-bin equal frequency discretization is used.]{\label{fig3d}\includegraphics[width=80mm,height=55mm]{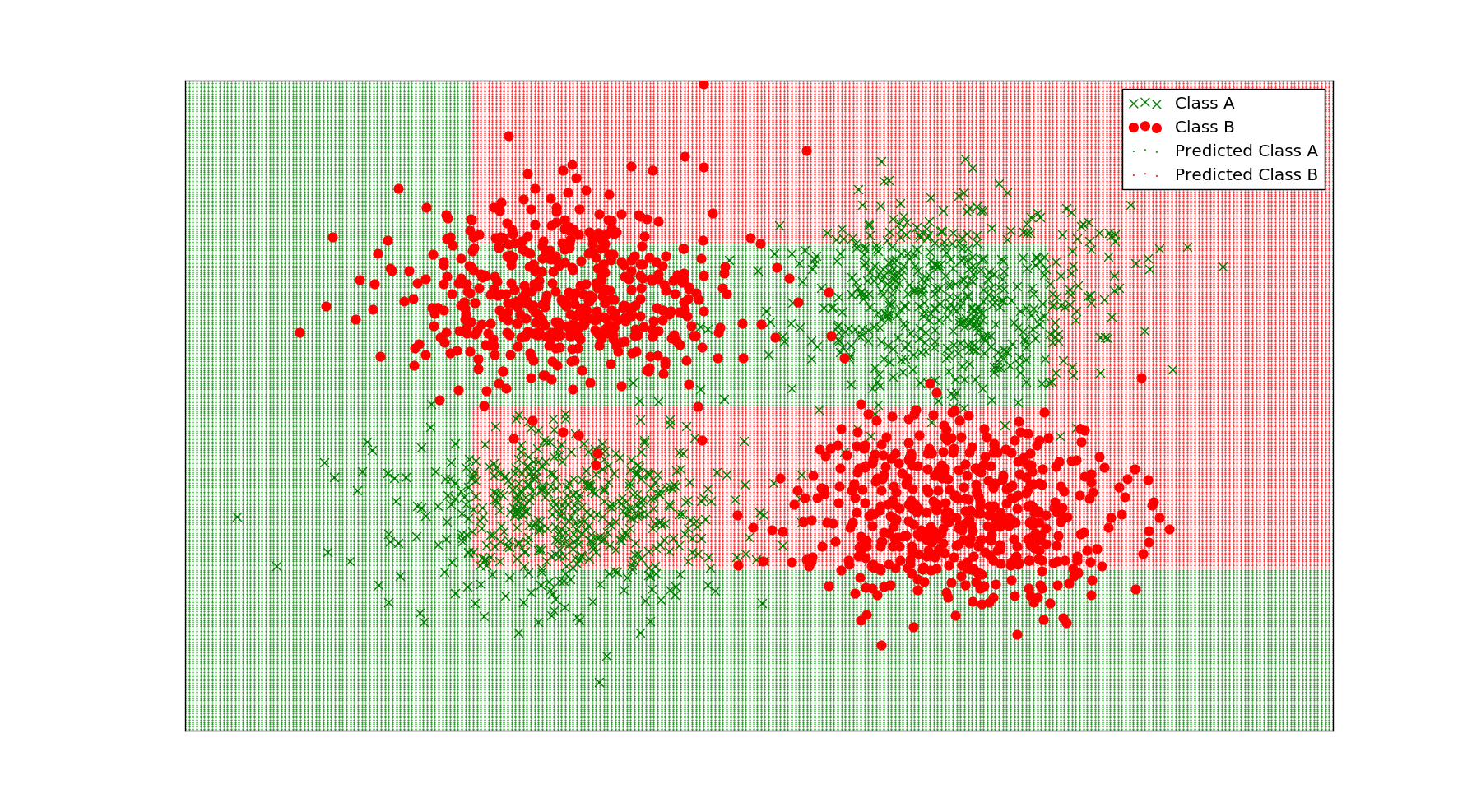}}
	\caption{Comparison of decision boundaries with and without discretization on a simple two-dimensional X-OR problem.}
	\label{fig:XOR}
\end{figure}

There are two important observations that motivate this work:
\begin{itemize}
\item A linear classifier with discretization is not linear with respect to the original data.
\item Contrary to what might be thought given that discretization loses information, a linear classifier with discretization can reduce representation bias and may consequently reduce error.
\end{itemize}
Note that we do not claim that discretization is the only useful feature transformation or a substitute for other approaches to creating non-linear classifiers, such as multi-layer perceptrons. The simple AND problem illustrated in Figure~\ref{fig:simple-example} is a case where  where discretization is as effective as any other method for obtaining non-linear decision surfaces. However, there are problems where discretization will not be this effective. 
One example is the X-OR problem illustrated in Figure~\ref{fig:XOR}.

Note also that we do not claim that discretization always reduces representation bias.
\iftrue
However, linear models make a strong implicit assumption, that the data are well modeled by a linear decision boundary.  As illustrated in Figure~\ref{fig:simple-example} discretization can overcome this assumption. The startling result that we reveal is that doing so is often useful in practice.  We show that discretization often reduces the bias of a linear classifier and that this reduction in bias frequently results in lower error when the data quantity is sufficiently large to minimize overfitting.
\else
\todo[inline]{\small Geoff, can you please do this? Restate the following in terms of discriminative linear classifiers}If a learner makes inappropriate assumptions about the data, it will be biased. Of course, the fewer  the inappropriate assumptions, the lower the bias. 
Let us take an example of a learner that can be biased -- one that creates linear models. Two well-known examples  are naive Bayes (NB) and logistic regression (LR). They can not take into account complex multi-variate interactions in the data. 
The models created by these two learners are equivalent in form and only differ in terms of the objective function that they optimize when parameterizing the model, specifically, log-likelihood for NB (generative) and conditional-log-likelihood for LR (discriminative). 
Discretization for naive Bayes has been relatively well-studied. It is widely accepted as the best default method for handling numeric data \citep{Yang2009} because it allows better estimates of likelihood than alternatives.

It will be interesting to see if discretization can help alleviate the effects of some of the strong linear assumptions of NB and LR, by making them non-linear to handle some of non-linear interactions present in various datasets -- the topic that we address in this paper.
However, in this work, we will be mainly interested in studying the effect of discretization on discriminative linear classifiers that is LR and its variants.

The underlying bias of an LR due to its linear nature has been investigated recently by~\cite{zaidi2016ALR}. They show that a higher-order LR taking into account quadratic and cubic interactions is much lower-biased than a linear LR.
A similar trend can be observed in the context of Bayesian Networks where it is shown that a higher value of $k$ for k-dependence estimators (KDB) also leads to lower bias~\citep{ana14a}. 

It has been claimed that reducing bias is important when learning from large datasets because variance will decrease as data quantity increases resulting in bias dominating error \citep{brain1999effect}.
Our goal in this paper is to investigate whether discretization can reduce LR's bias, and hence reduce its error, especially on big datasets.

The main contributions of this paper are as follows:
\begin{itemize}
\item We show that discretization can actually relaxes the strong linear assumption of a linear LR which results in reducing its bias. This results in significantly better accuracy on all datasets.
\item We show that a trend similar to LR exists for other linear classifiers based on Hinge and Mean-square-loss.
\end{itemize}
\fi

The rest of this paper is organized as follows. 
Some preliminary background and terminology is given in Section~\ref{subsec_terminology}.  
We discuss discretization in general in Section~\ref{subsec_discretization}. 
Linear classifiers based on Conditional Log-Likelihood (CLL) Hinge Loss (HL) and Mean-square-error loss are discussed in Sections~\ref{subsec_CLL},~\ref{subsec_HLoss} and~\ref{subsec_MSE} respectively. 
Optimization strategies for training these linear classifiers are discussed in Section~\ref{subsec_os}.
An overview of related work is given in Section~\ref{sec_rw}.
Experimental results are given in Section~\ref{sec_experiments}. 
We conclude in Section~\ref{sec_conclusions} with pointers to directions for future work.

\begin{table}[t] \scriptsize \center
\begin{tabular}{|p{2.8cm}|p{9.0cm}|}
\hline
Notation & Description \\
\hline

$I$								& Number of qualitative attributes\\
$K$								& Number of quantitative attributes\\
$n$								& Total number of attributes, $n = I + K$\\
$N$								& Number of data points in $\data$\\

$\data = \{ \obj^{(1)},\ldots,\obj^{(N)} \}$ 	& Data consisting of $N$ objects \\
$\labels = \{ y^{(1)},\ldots,y^{(N)}\}$		& Labels of data points in $\data$ \\

$\tilde \P(e)$						& Actual probability of event $(e)$\\

$\P(e)$							& Probability of event $e$\\
$\P(e\|g)$							& Conditional probability of event $e$ given $g$\\

$\obj = \langle x_0, x_1,\ldots,x_n \rangle$& An object ($n$-dimensional vector) and $\obj \in \data$ \\

$Y$								& Random variable associated with class label \\
$y$								& $y \in Y$. Class label for object. Same as $x_0$ \\
$C$								& $|Y|$, Number of classes\\

$X_i$							& Random variable associated with qualitative attribute $i$ \\
$x_i$							& Actual value that $X_i$ takes. \\
$|X_i|$							& (Applicable only to qualitative attributes) Number of values of attribute $X_i$\\

$\beta$							& LR parameter vector to be optimized \\

$\beta_{y,i}$						& LR parameter associated with quantitative attribute $i$ for class $y$ \\
$\beta_{y,k,j}$						& LR parameter associated with qualitative attribute $k$ for class $y$ taking value $j$ \\

$\beta_{y,0}$						& LR intercept term for class $y$. \\

$\lambda$							& Regularization parameter \\

\hline
\end{tabular}
\caption{List of symbols.}
\label{tab_notation}
\end{table}

\section{Background} \label{sec_notation} 

\subsection{Terminology} \label{subsec_terminology}
In machine learning and data mining research, there exists variation in the terminology when it comes to characterizing the nature of an attribute (or feature). For example, `continuous vs. discrete', `numeric vs.\ categorical' and `quantitative vs.\ qualitative'. We believe that the `quantitative vs.\ qualitative' distinction is best suited for our study in this paper and hence, this is used throughout the paper.

Qualitative attributes are the attributes on which arithmetic operations can not be applied. The values of a qualitative attribute can be placed or categorized in distinct categories. Sometimes there exist a meaningful rank among these categories, resulting in distinction of ordinal and nominal among quantitative attributes. For example, \emph{Student Grade: \{HD, D, C, P, F\}} and \emph{Pool Depth: \{Very Deep, Deep, Shallow\}} are ordinal attributes, while \emph{Marital Status: \{Married, Never-married, Divorced, Widow, Widower\}} and \emph{Nationality: \{Australian, American, British\}} are nominal attributes.

Quantitative attributes, on the hand, are the attributes on which arithmetic operations can be applied. They can be both discrete and continuous. For example, \emph{Number of Children} is a discrete-quantitative attribute (values determined by counting), whereas \emph{Temperature} is a continuous-quantitative attribute (values determined by measuring).

A list of the various symbols used in this work is given in  Table~\ref{tab_notation}.

\subsection{Discretization} \label{subsec_discretization}

Discretization is a common process in machine learning that is used to convert a quantitative into a qualitative attribute~\citep{Liu2002,Garcia2013}. The need for discretization originates from the facts that some classifiers can only handle,  and some others sometimes to operate better with qualitative attributes. The process involves finding cut-points within the range of the quantitative attribute and to group values into intervals based on these cut-points. This removes the ability to distinguish between data points falling in the same interval. Therefore, discretization entails information loss. 

Discretization methods can be categorized into two categories: \emph{Supervised} and \emph{Un-supervised}. In the unsupervised case, class information is not used during cut-point determination process. 
Popular approaches are equal-frequency and equal-width discretization. Equal-width discretization (EWD) divides the quantitative attribute's range (maximum value $x_i^{max}$ and minimum value $x_i^{min}$) into $k$ equal-width intervals where $k$ is provided by the user. Each interval will have a width of $w = \frac{x_i^{max} - x_i^{min}}{k}$. Equal-frequency discretization (EFD), on the other hand, divides the sorted values of a quantitative attribute such that each interval has approximately $k$ number of data points. Each interval will contain $N/k$ data points. It is also important that data points with identical value are placed in the same interval, therefore, in practice, each interval will have slightly different number of data points. Choosing EWD or EFD and the number of bins is problem specific and can have huge impact on the overall performance of any model. Of course, choosing a large $k$ will result in less information loss, but can result in over-fitting on small datasets. 

Supervised discretization methods, on the other hand, utilize the class information of the data point to better define the cut-points.  For example, state-of-the-art discretization technique Entropy-Minimization Discretization (EMD) sorts the quantitative attribute's values and then finds the cut-point such the information gain is maximized across the splits~\citep{Kohavi1996}. The technique is applied recursively on the successive splits and the minimum-description-length (MDL) criterion is used to determine when to stop splitting.


\subsection{Linear Classifier - CLL} \label{subsec_CLL}

A Logistic Regression classifier optimizes the conditional log-likelihood (CLL) which is defined as:
\begin{eqnarray} \label{eq_cll}
\textrm{CLL}(\beta) & = & \sum_{l = 1}^{N} \log \P(y^{(l)}\|\obj^{(l)}),
\end{eqnarray}
where
\begin{eqnarray} \label{eq_probs}
\P(y^{(l)}\|\obj^{(l)}) & = & \frac{\exp(\beta_{y,0} + \beta_y^T \obj^{(l)})}{\sum_{c=1}^{C} \exp(\beta_{y,0} + \beta_c^T \obj^{(l)})}.
\end{eqnarray}
The term $\beta_{y,0} + \beta_y^T \obj^{(l)}$ is expanded as: $\beta_{y,0} x_0^{(l)} + \beta_{y,1} x_1^{(l)} + \beta_{y,2} x_2^{(l)} + \dots + \beta_{y,n} x_n^{(l)}$, where $x_0$ can be assumed to be $1$ for all data points. 
Since the objective function as defined in Equation~\ref{eq_cll}, is linear in $x$, it is a linear classifier. 

Equation~\ref{eq_probs} leads to a multi-class softmax objective function. Since, a set of parameters are learned for each class, we have made this distinction explicit with subscript $y$ in parameter notation, that is, $\beta_{y,j}$ denotes a parameter for class $y$ and attribute $j$.
Typically, an LR minimizes the negative of the CLL known as negative log-likelihood (NLL), which is defined as:
\begin{eqnarray} \label{eq_num}
\textrm{NLL}(\beta) & = & - \sum_{l = 1}^{N} \left( \left(\beta_{y,0} + \beta_y^T \obj^{(l)}\right) - \log \left( \sum_{c=1}^{C} \exp(\beta_{c,0} + \beta_c^T \obj^{(l)}) \right) \right).
\end{eqnarray}
Note, in the following, for simplicity, we will drop the superscript $(l)$ notation.
It should be noted that many software libraries for multi-class LR are either based on implementing multi-class (softmax) objective function of Equation~\ref{eq_num} or they optimize a more simpler binary objective function of the following form:
\begin{eqnarray} \label{eq_numbin}
\textrm{NLL}(\beta) & = & \sum_{l = 1}^{N} \left( \log \left( 1 + \exp(\beta_{0} + \beta^T \obj^{(l)}) \right) \right),
\end{eqnarray}
and solve a one-versus-all classification problem.
Note that in the case of binary classifiers, there is only one set of parameters for the two classes as oppose to $C$ set of parameter that needed to be optimized for the softmax case. At classification time, one needs to apply $C$ different trained LR classifiers and choose one with the highest probability. 

Nonetheless, optimizing a standard LR with NLL based either on Equation~\ref{eq_num} or Equation~\ref{eq_numbin} requires substantial input manipulation, i.e., appending $1$ to all data points and then, converting qualitative attributes using one-hot-encoding. For example, a qualitative attribute $X_j$ taking values $\{a,b,c\}$, will be converted into three attributes $X_j, X_{j+1}, X_{j+2}$, each taking values either $0$ or $1$. 
An alternative to manipulating the input is to modify the model and optimize the following objective function instead:
\begin{eqnarray} \label{eq_numdisc}
\textrm{NLL}(\beta) & = & - \sum_{l = 1}^{N}  \left( \left( \beta_{y,0} + \sum_{i=1}^I \beta_{y,i} x_i + \sum_{k=1}^K \sum_{j=1}^{|X_k|} \beta_{y,k,j} \mathbf{1}_{X_k = x_j} \right) \right.  \\
& & - \left. \log \left( \sum_{c=1}^{C} \exp(\beta_{c,0} + \sum_{i=1}^I \beta_{c,i} x_i + \sum_{k=1}^K \sum_{j=1}^{|X_k|} \beta_{c,k,j} \mathbf{1}_{X_k = x_j}) \right) \right). \nonumber
\end{eqnarray} 
Note that the models expressed in~Equation~\ref{eq_num} and~\ref{eq_numdisc} are exactly equivalent and will lead to the same results. The only difference is that the model in Equation~\ref{eq_num} requires converting all qualitative attributes into quantitative ones using one-hot-encoding, whereas the model in Equation~\ref{eq_numdisc} does not.
Equation~\ref{eq_numdisc} can be simplified even further -- for datasets with only qualitative attributes, and including only terms that are not canceled out, we have:
\begin{eqnarray} \label{eq_disc}
\textrm{NLL}(\beta) & = & - \sum_{l = 1}^{N}  \left( \beta_{y,0} + \sum_{k=1}^K \beta_{y,k,j} \mathbf{1}_{X_k = x_j, Y = y} \right.  \nonumber \\
& & - \left. \log \left( \sum_{c=1}^{C} \exp(\beta_{c,0} + \sum_{k=1}^K \beta_{c,k,j} \mathbf{1}_{X_k = x_j, Y = c}) \right) \right).
\end{eqnarray}
Instead of converting qualitative attributes into quantitative ones and using the model of Equation~\ref{eq_num}, one can convert quantitative attributes into qualitative ones using discretization methods as discussed in Section~\ref{subsec_discretization} and use the model of Equation~\ref{eq_disc}.
It can be seen that with Equation~\ref{eq_num}, the number of parameters optimized are: $(C - 1) + (C - 1)n$.
Whereas, with Equation~\ref{eq_disc}, $(C - 1) + (C - 1) \sum_{i=1}^{n}|X_i|$ parameters are optimized. 
Since the two models are not equivalent, this will result in different training time, speed and rate of convergence and of course, classifications.



\subsection{Linear Classifier -- Hinge Loss} \label{subsec_HLoss} 

Hinge Loss (HL) is widely used as an alternative to CLL and has been the basis of Support Vector Machines. A classifier optimizing either a Hinge Loss objective function or its variant is a linear classifier and is known as the Support Vector Classifier (SVC). Here we define L2-Loss HL as:
\begin{eqnarray} \label{eq_hl}
\textrm{HL}(\beta) & = & \sum_{l = 1}^{N} \textrm{max}(0, 1 - y\beta^T \obj)^2.
\end{eqnarray}
An alternative is L1-Loss HL which is equal to: $ \sum_{l = 1}^{N} \textrm{max}(0, 1 - y\beta^T \obj)$. In this work, we will focus only on the L2-Loss. 
In practice, a penalty term is also added for regularizing the objective function as:
\begin{eqnarray} \label{eq_hlreg}
\textrm{HL}(\beta) & = & \frac{1}{2}||\beta^T \beta||^2 + \lambda \sum_{l = 1}^{N} (\textrm{max}(0, 1 - y\beta^T \obj))^2,
\end{eqnarray}
where $\lambda$ is the regularization parameter. 
We will discuss the gradient and Hessian of this objective function later in Section~\ref{subsec_os}.

\subsection{Linear Classifier -- Mean-Square-Error} \label{subsec_MSE} 

Another linear classifier is based on optimizing the Mean-Square-Error (MSE) objective function and is defined as:
\begin{eqnarray} \label{eq_mse}
\textrm{MSE}(\beta) & = & \frac{1}{2} \sum_{l = 1}^{N} \sum_{c = 1}^{C} (\tilde{\P}(c\|\obj) - \P(c\|\obj))^2,
\end{eqnarray}
where $\P(c\|\obj)$ is given in Equation~\ref{eq_probs} and $\tilde{\P}(c\|\obj$) is the actual probability of class $c$ given data instance $\obj$. This will be a vector of size $C$ with all zeros except at the location of the label of $\obj$, where it will be 1 (assuming there are no duplicate data points in the dataset). The objective function of Equation~\ref{eq_mse} is similar to that optimized by artificial neural-networks (ANN). However, in ANN, $\P(c\|\obj)$ is defined in terms of multiple layers. We can interpret Equation~\ref{eq_mse} as the objective function of a zero-layer ANN.

\subsection{Optimization} \label{subsec_os}

There is no closed form solution to optimizing the negative log-likelihood, hinge loss and mean-square-error objective function, and, therefore, one has to resort to iterative minimization procedures such as gradient descent or quasi-Newton.
%
An iterative optimization procedure generates a sequence $\{\beta^k\}_{k=1}^{\infty}$ converging to an optimal solution.
At every iteration $k$, the following update is made: $\beta^{k+1} = \beta^{k} + \s^k$, where $\s^k$ is the search direction vector. 
The following equation plays the pivotal role as it holds the key to obtain $\s^k$ by solving a system of linear equations: 
\begin{eqnarray} \label{eq_NewtonSearch}
\nabla^2 f(\beta^k) \s^k = - \nabla f(\beta^k),
\end{eqnarray} 
where $f$ is the objective function that we are optimizing.
There are two very important issues that must be addressed when solving for search direction vector using Equation~\ref{eq_NewtonSearch}~\citep{nocedal2006numerical}. 
First, it can be infeasible to explicitly compute and store the Hessian, especially on high-dimensional data. 
Second, the solution obtained using Equation~\ref{eq_NewtonSearch}, does not guarantee  convergence.
There are three main strategies for addressing the first issue:
\begin{itemize}
\item Consider $\nabla^2 f(\beta^k)$ to be an identity matrix -- in this case, $\s^k = - \nabla f(\beta^k)$. This leads to a family of algorithms known as first-order methods such as Gradient Descent, Coordinate Descent, etc.
\item Do not compute $\nabla^2 f(\beta^k)$ directly, but approximate it from the information present in $\nabla f(\beta^k)$ instead. 
This property is useful for large scale problems where we cannot store the Hessian matrix.
This leads to approximate second-order methods known as quasi-Newton algorithms, for example, L-BFGS which, is considered to be the most efficient algorithm (de-facto standard) for training LR.
\item Third, use standard `direct algorithms' for solving a system of linear equations such as Gaussian elimination to solve for $\s^k$. Or, use any one of the iterative algorithm such as conjugate gradient, etc.
For large datasets, generally iterative methods are preferable over direct methods, as the former requires computing the whole Hessian matrix. The optimization method now has two layers of iterations. An outer layer of iteration to update $\beta^k$, and an inner layer of iterations to find Newton direction $\s^k$. In practice, one can only use an approximate Newton direction in early stages of the outer iterations. This method is known as `Truncated Newton method'~\citep{nash2000}.
\end{itemize}
It should be noted that these methods differ in terms of the speed-of-convergence, cost-per-iteration, iterations-to-convergence, etc. 
For example, Coordinate Descent updates one component of $\beta$ at every iteration, so the cost-per-iteration is very low, but iterations-to-convergence will be very high. On the other hand, Newton methods, will have high cost-per-iteration, but very low number of iterations-to-convergence.
The three methods described above are all affected by the scaling of the axis. Therefore, scaling quantitative attributes or converting quantitative  into qualitative attributes will effect the speed and the quality of the convergence.

The second issue can be addressed by adjusting the length of the Newton direction. For that, two techniques are mostly used -- line search and trust region.
Line search methods are standard in optimization research. We can modify Equation~\ref{eq_NewtonSearch} as $\nabla^2 f(\beta^k) \s^k = - \eta^k \nabla f(\beta^k)$, where $\eta^k$ is known as the step-size. Standard line searches obtain an optimal step-size as a solution to the following sub-optimization problem: $\eta^k = \textrm{argmin}_\eta f(\beta^k + \eta \s^k)$.
Trust-region methods, unlike line search, are relatively new in optimization research. 
Trust-region methods first find a region around the current solution -- in this region, a quadratic (or linear) model is used to approximate the objective function. The step size is determined based on the goodness of fit of the approximate model. 
If a significant decrease in the objective function is achieved with a forward step, the approximated model is a good representative of the original objective function and vice-versa. The size of the (trust) region is specified as a spherical area of size $\Delta_k$. The convergence of the algorithm is guaranteed by controlling the size of the region which (in each iteration) is proportional to the reduction in the value of objective function in the previous iteration.

In the following we will define the gradient and Hessian of the  three objective functions  conditional log-likelihood, hinge loss and mean square error. Note, we only define the gradient and the Hessian for qualitative attributes here. 
For softmax CLL, $\nabla f(\beta^k)$ and $\nabla^2 f(\beta^k)$ can be written as:
\begin{eqnarray} \label{eq_lrgrad}
\frac{\partial  \textrm{NLL} (\beta)}{\partial \beta_{y',i}} & =  & \sum_{l=1}^{N} (\mathbf{1}_{y=y'} - \P(y'|\obj)) x_i, \nonumber
\end{eqnarray}
\begin{eqnarray} \label{eq_lrhess}
\frac{\partial^2 \textrm{NLL} (\beta)}{\partial \beta_{y',i} \partial \beta_{y'',j}} &  =  & - \sum_{l=1}^{N} (\mathbf{1}_{y'=y''} - \P(y'|\obj)) \P(y''|\obj) x_i x_j. \nonumber
\end{eqnarray}
For Hinge-loss, the (sub-) gradients can be written as:
\begin{eqnarray} \label{eq_hlgrad}
\frac{\partial  \textrm{HL} (\beta)}{\partial \beta_{i}} & \!\!\! = \!\!\! & 2 \sum_{l=1}^{N'} (y\beta^T\obj - 1) y x_i, \nonumber
\end{eqnarray}
where $N'$, are the instances for which $y\beta^T\obj < 1$ is true. Similarly the Hessian can be written as:
\begin{eqnarray} \label{eq_hlhess}
\frac{\partial^2 \textrm{HL} (\beta)}{\partial \beta_{i} \partial \beta_{j}} &  =  & 2 \sum_{l=1}^{N'} x_i x_j \nonumber
\end{eqnarray}
For MSE, one can write the gradients as:
\begin{eqnarray} \label{eq_msegrad}
\frac{\partial  \textrm{MSE} (\beta)}{\partial \beta_{j,x_j,k}} & =  & \sum_{l=1}^{N} \sum_{c=1}^{C} (\mathbf{1}_{y=c} - \P(c|\obj)) (\mathbf{1}_{k=c} - \P(k|\obj)) x_i, \nonumber
\end{eqnarray}
and the Hessian can be written as:
\begin{eqnarray} \label{eq_msehess}
\frac{\partial^2 \textrm{MSE} (\beta)}{\partial \beta_{j,x_j,k} \partial \beta_{j',x_{j'},k'}} & \!\!\!\!\! = \!\!\!\!\! & - \sum_{l=1}^{N} \left[ \right.
(-1)(\mathbf{1}_{k'=c} - \P(k'|\obj)) \P(c|\obj) \mathbf{1}_{j=j'} + \nonumber \\
& & (\mathbf{1}_{y=c} - \P(c|\obj)) (-1) (\mathbf{1}_{k'=k} - \P(k'|\obj)) P(k|\obj) \mathbf{1}_{j=j'} \P(c|\obj) + \nonumber \\
& & \left. (\mathbf{1}_{y=c} - \P(c|\obj)) (\mathbf{1}_{k=c} - \P(k|\obj)) (\mathbf{1}_{k'=c} - \P(k'|\obj)) \P(c|\obj) \mathbf{1}_{j=j'} \right] x_i x_j
\nonumber
\end{eqnarray}

\section{Related Work} \label{sec_rw} 

Discretization is often motivated by a need to adapt data for a model that cannot handle quantitative attributes. 
In Statistics and many of its related and applied branches (such as epidemiology, medical research and consumer marketing), it goes by names of `dichotomization' and `categorization' (where the two techniques differ as the former splits the measurement scale into two while the later can have more than two categories) -- and has been examined in many studies \citep{Irwin2003,MacCallum2002,Greenland1995}.
However, in most of these studies a majority opinion is against the use of dichotomization -- and for categorization, it is advised to be used with caution. 
The main reason cited for this is that dichotomization and categorization lead to information loss since the variability among the members of the group is subsumed. For example,~\cite{Altman2006} write: 
\begin{quote}  ... Firstly, much information is lost, so the statistical power to detect a relation between variable and patient outcome is reduced ... and considerable variability may be subsumed within each group. Individuals close to but on opposite sides of cut-point are characterized as being very different rather than very similar ... \end{quote}
In practical machine learning, the common practice is to discretize an attribute only if necessary (i.e., if a model expects categorical attributes).
An exception is for Bayesian classifiers, where it is common practice to discretize numeric attributes \citep{Yang2009}.

The ambivalence towards discretization is understandable. Obviously, the quality (and sometime quantity) of data is the key to training accurate models and hence getting good results. In many cases, the data are the result of costly and time-consuming efforts (for example in breast cancer research where there are several stake-holders involved just to obtain a few attributes of the data). Losing some of the data (or more precisely, losing some distinction among the instances) due to discretization should be undesirable. However, a number of motivations for discretization have been put forward:
\begin{itemize}
	\item Discretization can lead to simplification of statistical analysis. For example, if a quantitative attribute is split on the median, then one can compare the two groups based on $t$, $\chi^2$ or some other test to estimate the difference between the two groups. This may  ease  interpretation and presentation of results ~\citep{Altman2006}.
	\item If there is error in the measurement scale, discretization can improve the performance of the model by reducing the  contamination~\citep{Flegal1991,Kupper1991,Fung1994,Shentu2010}.
	\item 
	In many domains there exist pre-defined (or standard) thresholds to convert a quantitative to a qualitative scale.
	In these cases, a discretized attribute might better represent the task at hand as it will be more interpretable or have distinct significance. For example, in medical research doctors might better interpret blood-pressure as high and low rather than on a numeric scale. 
	\item A discretized attribute might be better utilized than the quantitative attribute by the learning system. 
	For example, consider a classifier that relies on estimation of conditional probabilities such as $\P(x_i\|y)$. If $X_i$ is quantitative, $x_i$ can take infinite many values and if the number of training samples are small, reliable estimation of $\P(x_i\|y)$ from the data is not possible. A common approach is to impose a parametric model to estimate the value of $\P(x_i\|y)$ based on this model in which case the accuracy will depend on the appropriateness of the parametric model selected. Discretization can obviate this problem. By converting a quantitative attribute $X_i$ into a qualitative one $X^*_i$, the probabilities will take the form of $\P(x^*_i\|y)$ which may be reliably estimated from the data as there will be many $x_i$ values falling into the same interval \citep{Yang2009}.
	\item The final reason for discretization has to do with overcoming a model's assumptions. It might be the case that discretization help avoid some strong assumption that the learner makes about the data. If those assumptions are correct, discretization will have a negative impact, but if those assumptions are false, discretization may lead to better results \citep{Altman1994}.  It is this final motivation that we examine herein.
\end{itemize}

The effect of discretization on various classification algorithms such as naive Bayes, Support Vector Machines and Random Forest is discussed in~\cite{Lustgarten2008}. On many biomedical datasets, it is shown that discretization can greatly improve the performance of the learning algorithm. The role of discretization as feature selection technique is also explored. On various contrived datasets, \cite{Maleki2009} studied the effect of discretization on the precision and recall of various classification methods.

The effectiveness of discretization for naive Bayes classifier is relatively well studied~\citep{Hsu2000, Dougherty1995, Yang2009}.~\cite{Dougherty1995} conducted an empirical study of naive Bayes with four well-known discretization methods and found that all the discretization methods result in significantly reducing error relative to a naive Bayes that assumes a Gaussian distribution for the continuous variables.
~\cite{Hsu2000} attributes this to the \emph{perfect aggregation} property of Dirichlet distributions. 
In naive Bayes settings, a discretized continuous distribution is assumed to have a categorical distribution with Dirichlet priors. 
The perfect aggregation property of Dirichlet implies that we can learn the class-conditional probability of the discretized interval with arbitrary accuracy. 
It is also shown that there exists a \emph{partition independence assumption}, by virtue of that, Dirichlet parameters corresponding to a certain interval depend only on the area below the curve of the probability distribution function, but is independent of the shape of the curve in that interval.

\section{Experiments} \label{sec_experiments}

In this section, we compare the performance of linear classifier with discretized linear classifier on various datasets from the UCI repository~\citep{UCIrepository}. 

We denote linear classifier optimizing the conditional log-likelihood as LR, a linear classifier optimizing the Hinge loss as SVC (support vector classifier) and a linear classifier optimizing the mean-square-error as $\ANN$ (artificial neural network with zero hidden layers) -- their discrete counterparts are denoted as LR(d), SVC(d) and $\ANNd$ respectively. 

In the remainder of this paper, when discussing results, we will collectively refer to LR, SVC and $\ANN$ as linear classifiers and denote them by $\LC$. 
We will collectively refer to LR(d), SVC(d) and $\ANNd$ as discretized linear classifiers and denote them by $\dLC$.

The details of datasets used in this work are given in Appendix~\ref{app_datasets}.
For discretized linear classifiers, different supervised and unsupervised discretization techniques were considered. Since, this is not a comparative study on the relative efficacies of various discretization techniques for linear classifiers, we only report results with supervised entropy-based discretization of~\cite{Fayyad1992}, which we found gives better results than other discretization methods such as equal-frequency, equal-width, etc.

Each algorithm is tested on each dataset using either $5$ or $10$ rounds of $2$-fold cross validation. 

During the presentation of results, we split our datasets into two categories -- \emph{Big} and \emph{Little}. The \emph{Big} category comprises of datasets with more than $100,000$ instances and the \emph{Little} category comprises of the remaining datasets with $<100,000$ instances.

We compare four different metrics: 0-1 Loss, RMSE, Bias and Variance. We also compare training-time, testing time, and rate of convergence.
As discussed in Section~\ref{sec_intro}, the reason for performing bias-variance estimation is that it provides insights into how the learning algorithm might be expected to perform with varying amounts of data. We expect low variance algorithms to have relatively low error for small data and low bias algorithms to have relatively low error for large data~\citep{Brain2002}.
There are a number of different bias-variance decomposition definitions.
In this research, we use the bias and variance definitions of~\cite{KohaviWolpert1996} together with the repeated cross-validation bias-variance estimation method proposed by~\cite{WebbMultiBoosting2000}.

We report Win-Draw-Loss (W-D-L) results when comparing the 0-1 Loss, RMSE, bias and variance of two models. 
A two-tail binomial sign test is used to determine the significance of the results. Results are considered significant if $p \leq 0.05$ and shown in bold. 

For hinge-loss, a dataset with more than two classes was transformed into a binary dataset. Data points belonging to the majority class were assigned to class $A$ and the remaining data points were assigned to class $B$.

Missing values of the quantitative attribute were replaced with the mean of the attribute values whereas missing values of the qualitative attribute were treated as a distinct attribute value.

Quantitative attributes were also normalized between 0 and 1, as this is often recommended for gradient-based optimization methods.

Three optimization methods -- gradient descent, quasi-Newton, Trust-region based Newton method (TRON) were used. We found TRON to be converging relatively faster than the other methods. Therefore, in the following, we report results with TRON optimization only.
However, it is worth mentioning that a similar pattern of results was seen between $\LC$ and $\dLC$ for the other optimization methods.

\subsection{Comparison of the Accuracy of $\dLC$ and $\LC$}

\begin{figure}[t] 
\centering
\hspace{-0.20in}
{\includegraphics[width=55mm,height=50mm]{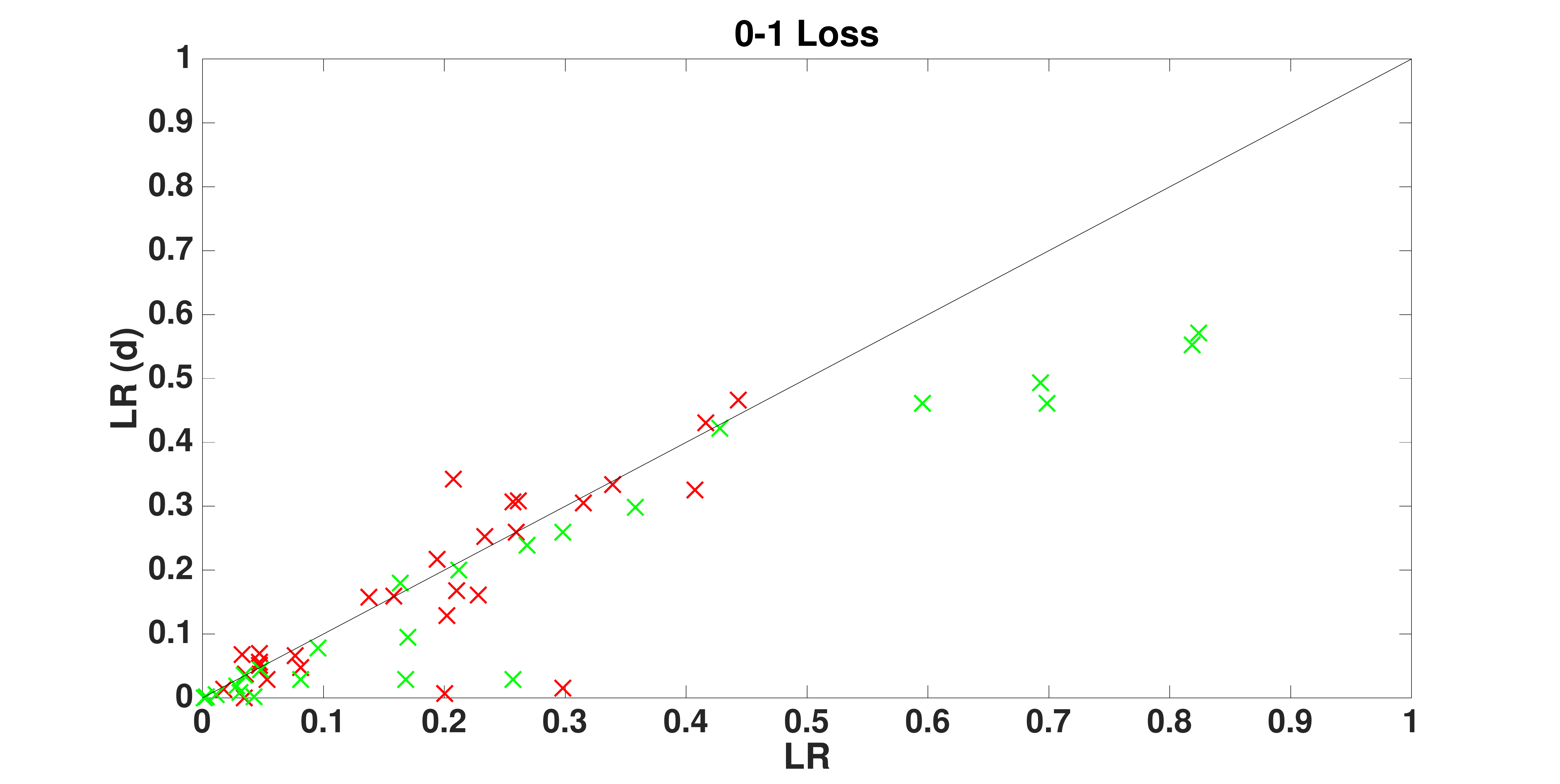}}\hspace{-0.20in}
{\includegraphics[width=55mm,height=50mm]{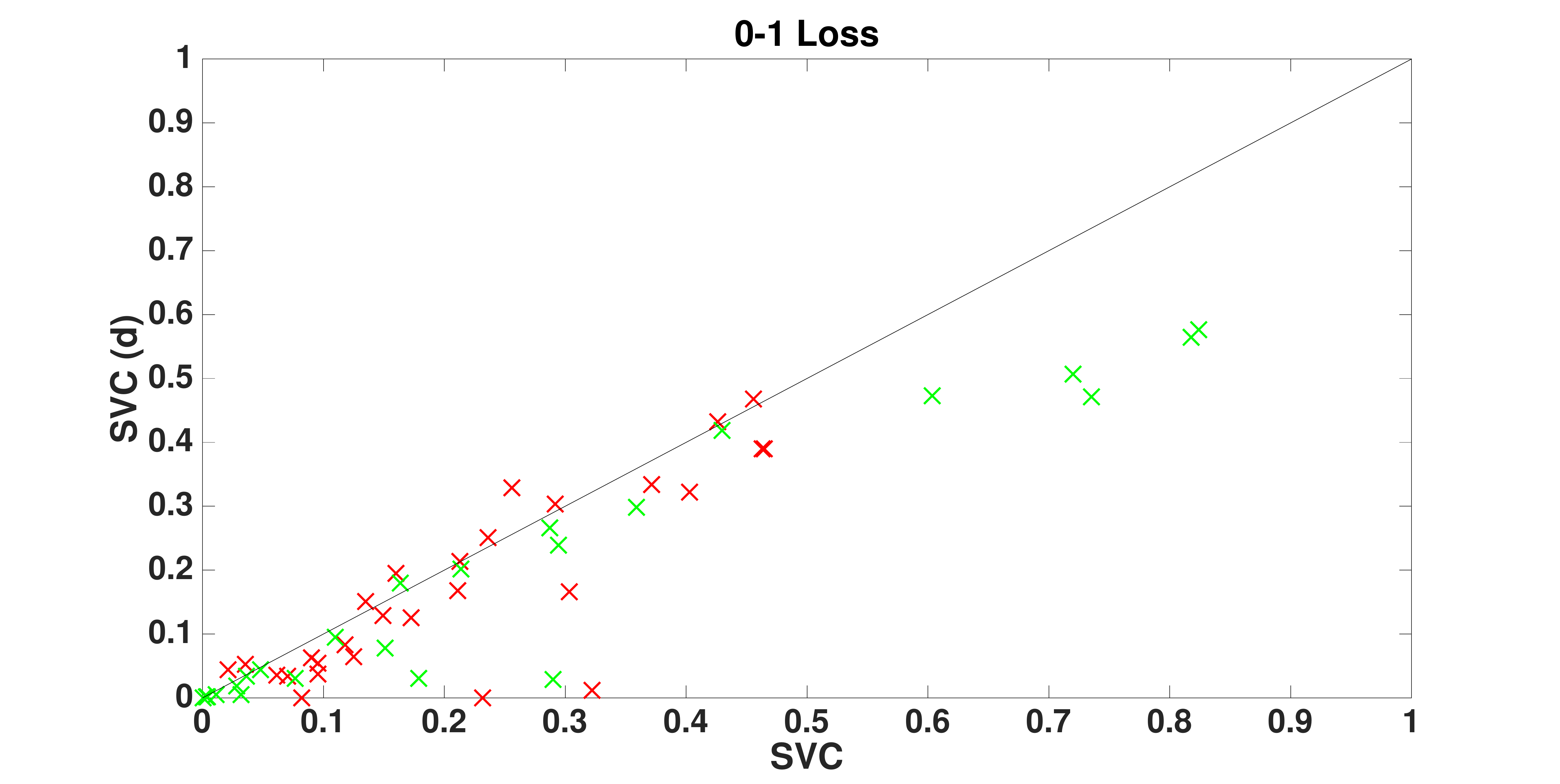}}\hspace{-0.20in}
{\includegraphics[width=55mm,height=50mm]{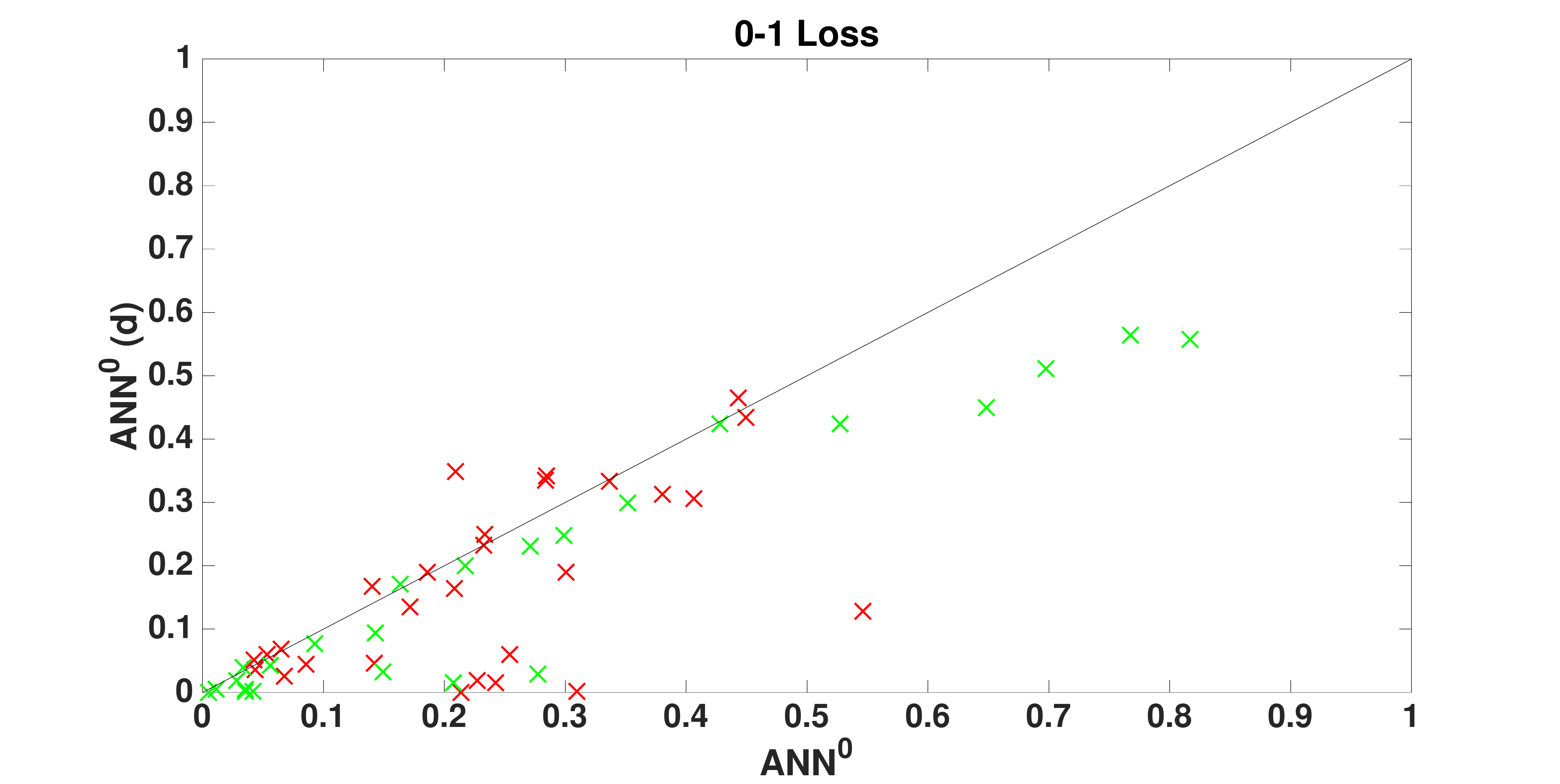}}\hspace{-0.20in}
\caption{\small 
Comparative scatter of \texttt{0-1 Loss} results for linear and discretized-linear classifiers. Linear classifiers are on the X-axis whereas discretized-linear classifiers are on the Y-axis. 
For points below the diagonal line, discretized-linear classifiers win. 
Results on $Big$ datasets are shown in green, whereas results on $Little$ datasets are shown in red.}
\label{fig_LRvsdLR_01Loss}
\end{figure}

\begin{figure}[t] 
\centering
\hspace{-0.20in}
{\includegraphics[width=55mm,height=50mm]{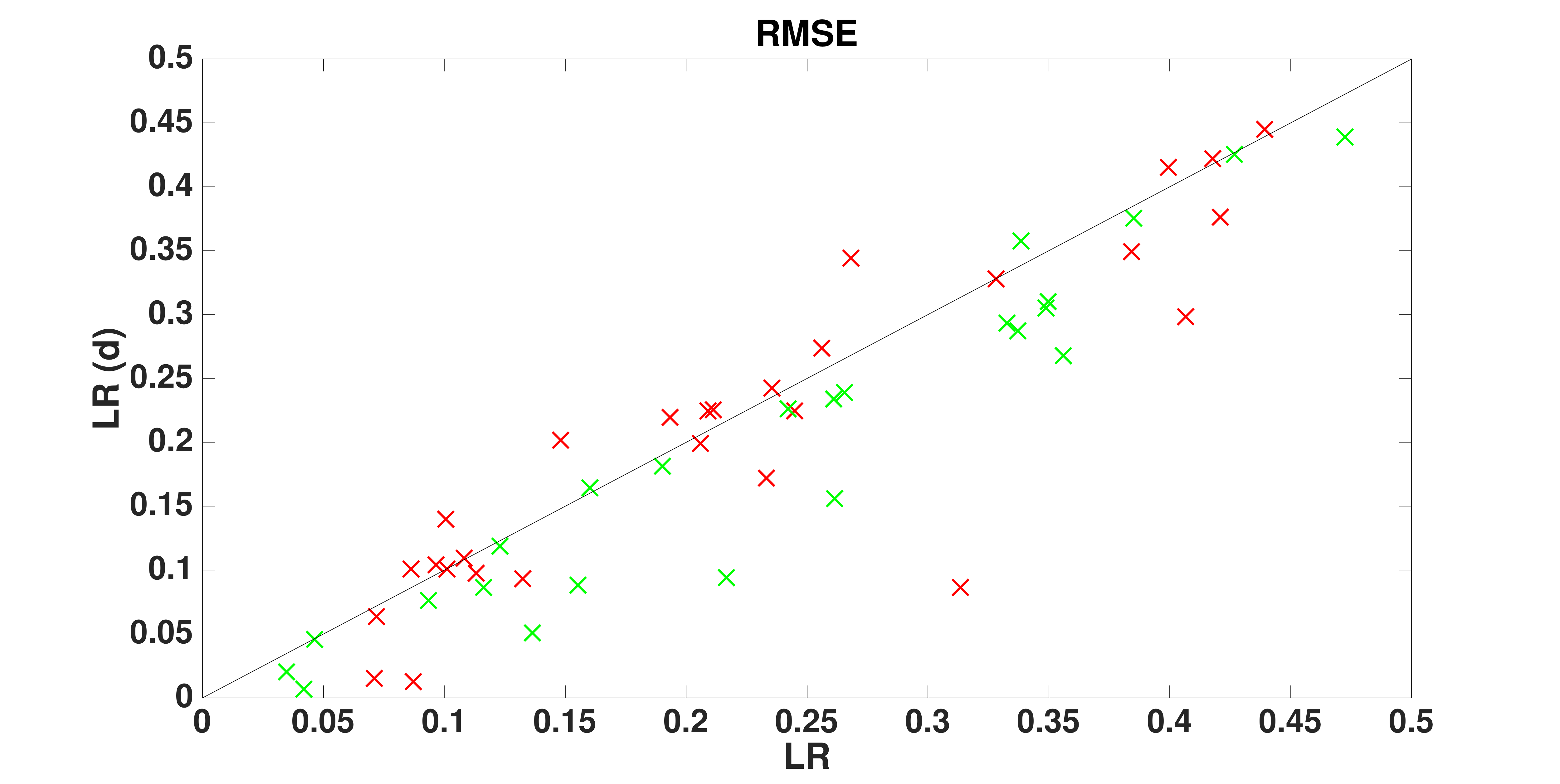}}\hspace{-0.20in}
{\includegraphics[width=55mm,height=50mm]{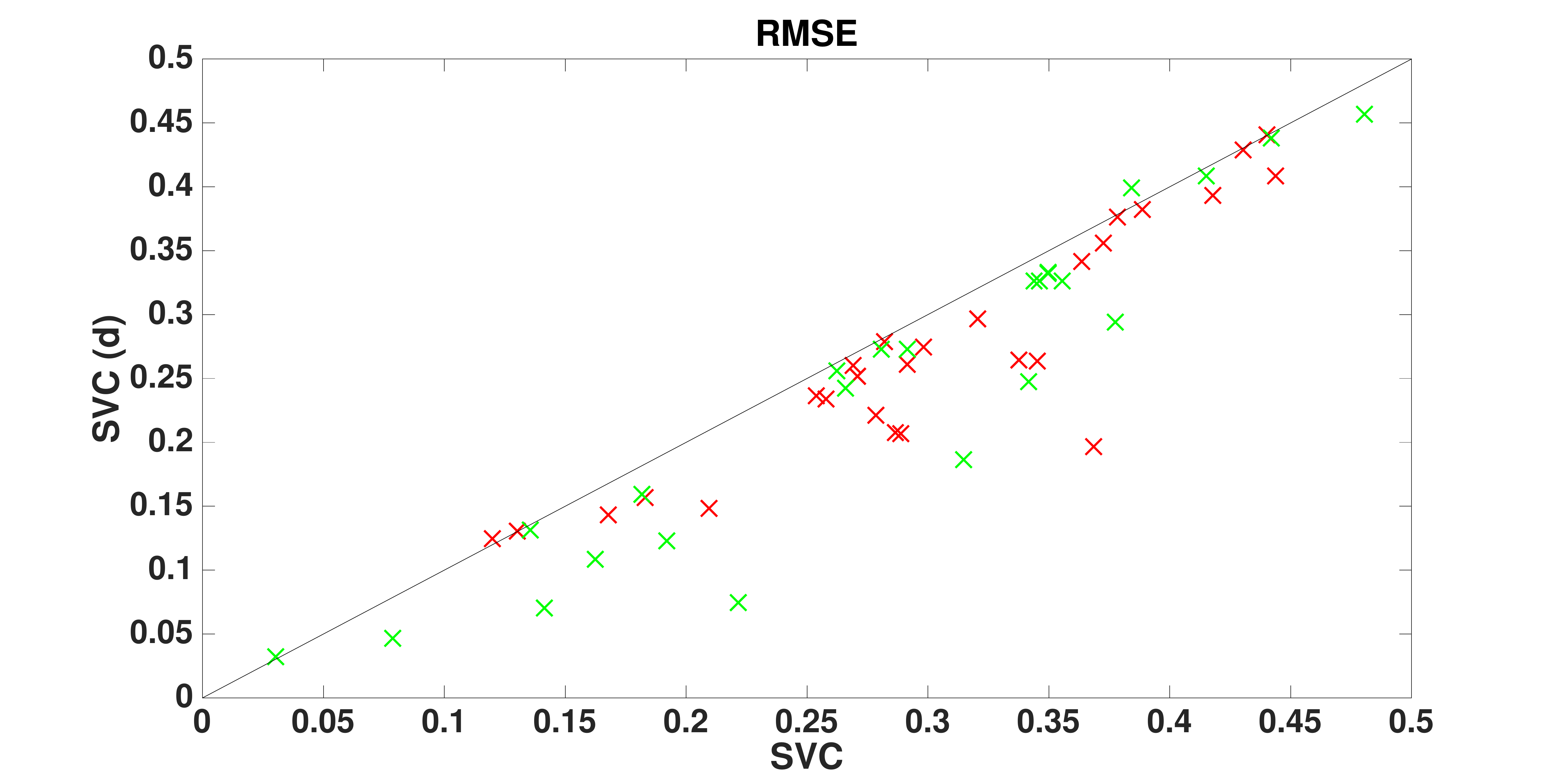}}\hspace{-0.20in}
{\includegraphics[width=55mm,height=50mm]{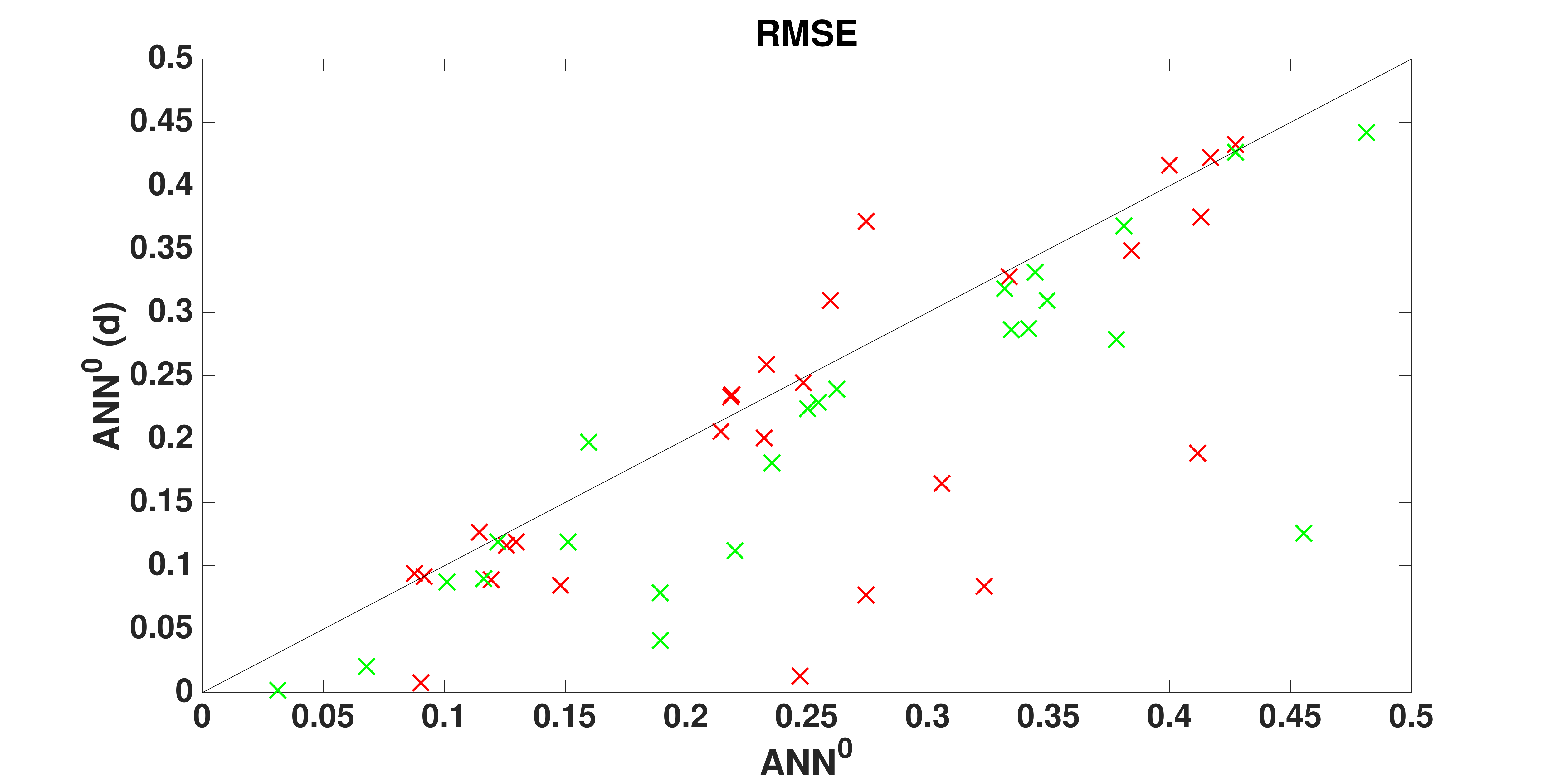}}\hspace{-0.20in}
\caption{\small 
Comparative scatter of \texttt{RMSE} results for linear and discretized-linear classifiers. Linear classifiers are on the X-axis whereas discretized-linear classifiers are on the Y-axis. 
For points below the diagonal line, discretized-linear classifiers win. 
Results on $Big$ datasets are shown in green, whereas results on $Little$ datasets are shown in red.}
\label{fig_LRvsdLR_RMSE}
\end{figure}

In this section, we compare the accuracy of $\dLC$ and $\LC$ in terms of their 0-1 Loss and RMSE on $52$ datasets. Results are shown in Figures~\ref{fig_LRvsdLR_01Loss} and~\ref{fig_LRvsdLR_RMSE}. 
It can be seen that the three $\dLC$ classifiers result in much better accuracy than their corresponding $\LC$. 
In the scatter plots, results on \emph{Big} datasets are shown in green dots, whereas results on \emph{Little} datasets are shown in red dots.
It can be seen that, on almost all $\emph{Big}$ datasets, $\dLC$ leads to higher accuracy (most green-dots are below the diagonal line). 
It can also be seen that some of the differences are substantial -- this shows the effectiveness of discretization on LR, SVC and $\ANN$ especially for big datasets.

A comparison of the win-draw-loss between the two models is given in Table~\ref{tab_dLRvsLR}. It can be seen that on big datasets, $\dLC$ wins on all except on $2$ datasets -- very promising result. This proves our hypothesis that on big datasets, discretization leads to low-bias non-linear classifier resulting in far superior results than a linear classifier with no discretization. 
On small datasets, discretization is significantly effective for SVC and non-significantly effective for $\ANN$.
Note that on small datasets, LR(d) and LR leads to similar performances with $13$ wins and $14$ losses for 0-1 Loss and $12$ wins and $15$ losses for RMSE. 
However, one should take into account that the scale of LR(d) wins is much higher than that of LR. This can be seen from the spread of red-dots in the left-most plots of Figures~\ref{fig_LRvsdLR_01Loss} and~\ref{fig_LRvsdLR_RMSE}.
\begin{table}[t] \scriptsize
\centering
\tabcolsep=4.0pt\renewcommand{\arraystretch}{1.5}
\begin{tabular}{p{2cm}cccccc}
\cline{1-7}
& \multicolumn{2}{c}{\bf LR(d) vs. LR} & \multicolumn{2}{c}{\bf SVC(d) vs. SVC} & \multicolumn{2}{c}{\bf $\ANNd$ vs. $\ANN$} \\
\cmidrule {2-3}\cmidrule (l){4-5}\cmidrule (l){6-7}
& W-D-L& $p$& W-D-L& $p$& W-D-L& $p$ \\
\cline{1-7}
&\multicolumn{6}{c}{All Datasets} \\
\cline{1-7}
0-1 Loss   &35/1/16& \bf 0.011	& 39/1/12 & \bf $<$0.001	& 41/1/10 & \bf $<$0.001\\
RMSE	&34/1/7& \bf 0.016	& 39/1/12 & \bf $<$0.001 	& 47/1/4 & \bf $<$0.001\\
\cline{1-7}
&\multicolumn{6}{c}{Big Datasets} \\
\cline{1-7}
0-1 Loss   &22/0/2& \bf $<$0.001	& 22/0/2 & \bf$<$0.001	& 23/0/1 & \bf $<$0.001\\
RMSE	&22/0/2& \bf$<$0.001	& 22/0/2 & \bf$<$0.001 	& 22/0/2 & \bf $<$0.001\\
\cline{1-7}
&\multicolumn{6}{c}{Small Datasets} \\
\cline{1-7}
0-1 Loss   &13/1/14&  1.000	& 17/1/10 &  0.247	& 18/1/9 & 0.087\\
RMSE	&12/1/15&  0.701	& 17/1/10 &  0.247 	& 25/1/2 & \bf $<$0.001\\
\cline{1-7}
\end{tabular}
\caption{\small Win-Draw-Loss comparison of \texttt{0-1 Loss} and \texttt{RMSE} of LR(d) vs.\ LR,  SVC(d) vs.\ SVC and $\ANNd$ vs.\ $\ANN$. Significant results are shown in bold.} 
\label{tab_dLRvsLR}
\end{table}

\subsection{Comparison of the Bias and Variance}

\begin{figure}[t] 
\centering
\hspace{-0.20in}
{\includegraphics[width=55mm,height=50mm]{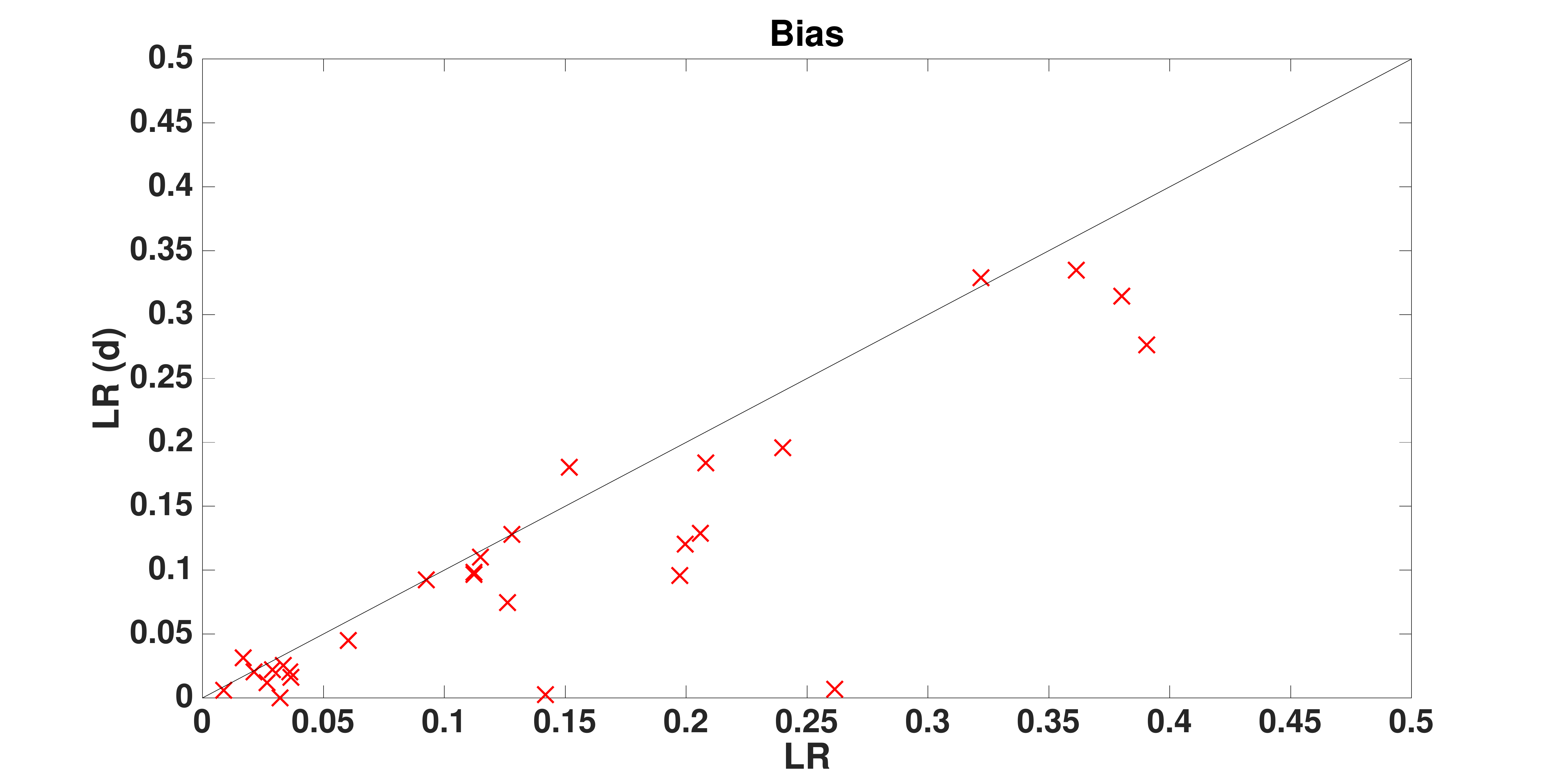}}\hspace{-0.20in}
{\includegraphics[width=55mm,height=50mm]{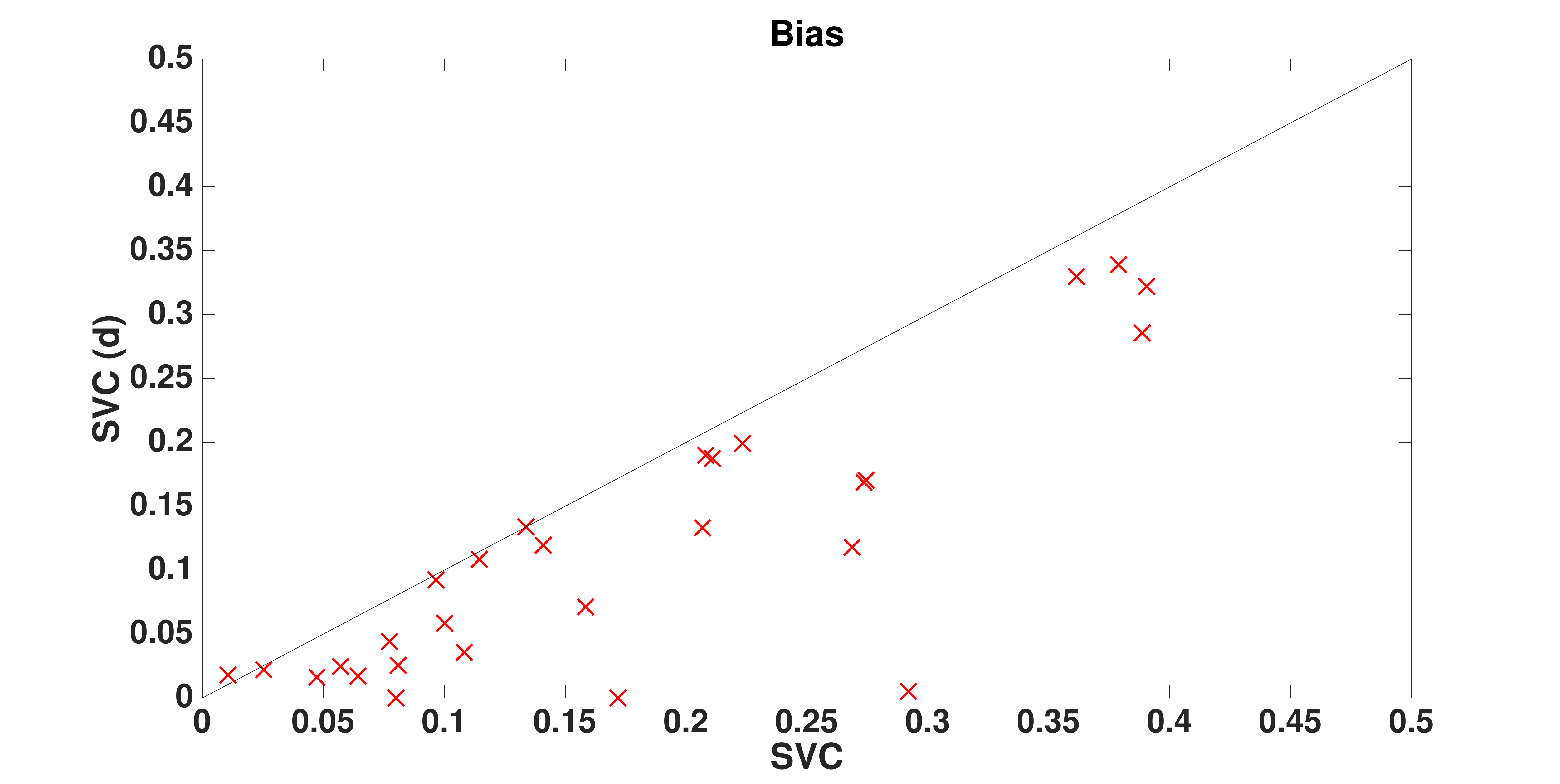}}\hspace{-0.20in}
{\includegraphics[width=55mm,height=50mm]{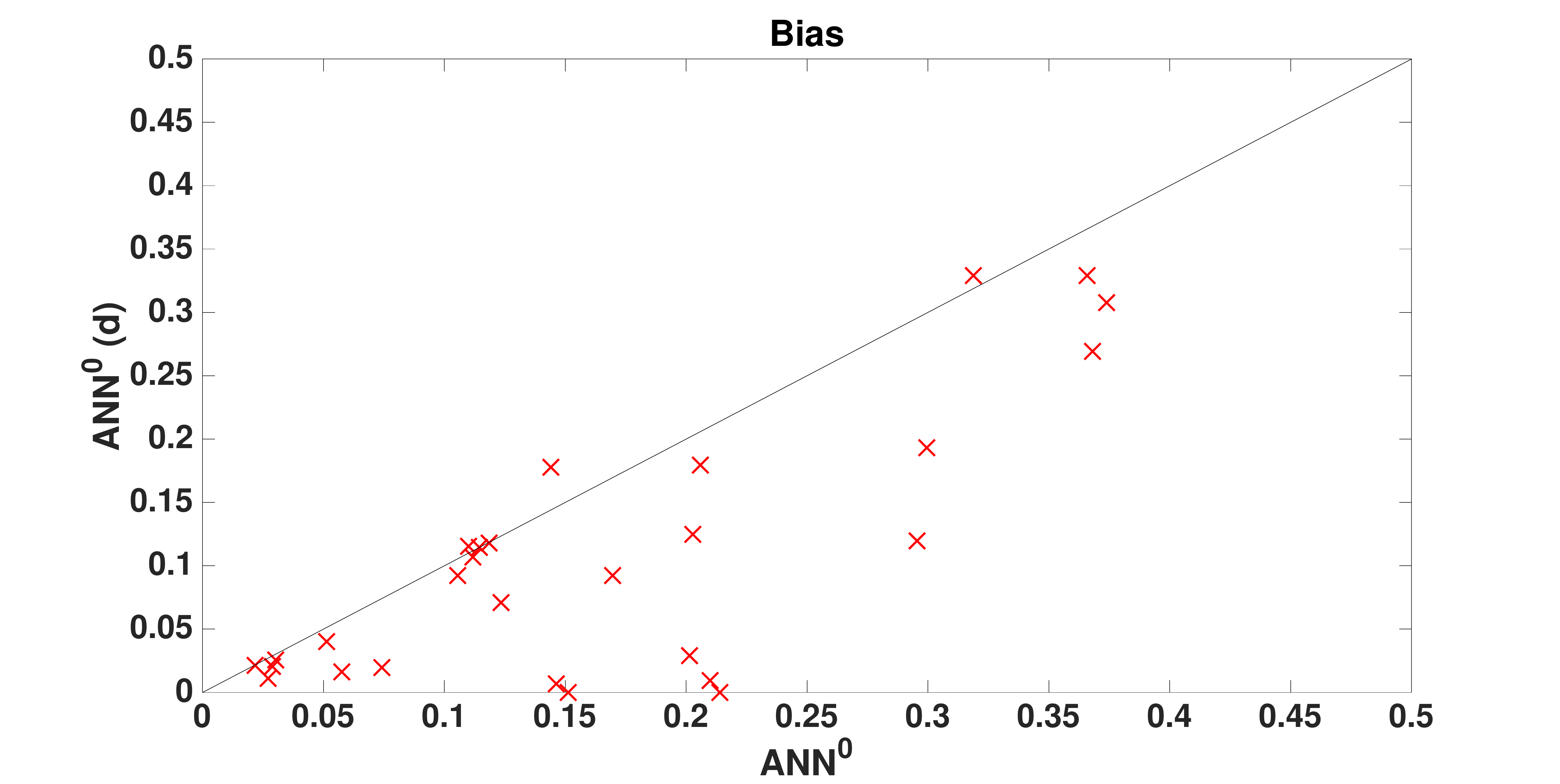}}\hspace{-0.20in}
\vspace{-0.15in}
\caption{\small 
Comparative scatter of \texttt{Bias} results for $\LC$ and $\dLC$ classifiers. $\LC$ are on the X-axis whereas $\dLC$ are on the Y-axis. 
For points below the diagonal line, $\dLC$ win.
}
\label{fig_LRvsdLR_Bias}
\end{figure}

\begin{figure}[t] 
\centering
\hspace{-0.20in}
{\includegraphics[width=55mm,height=50mm]{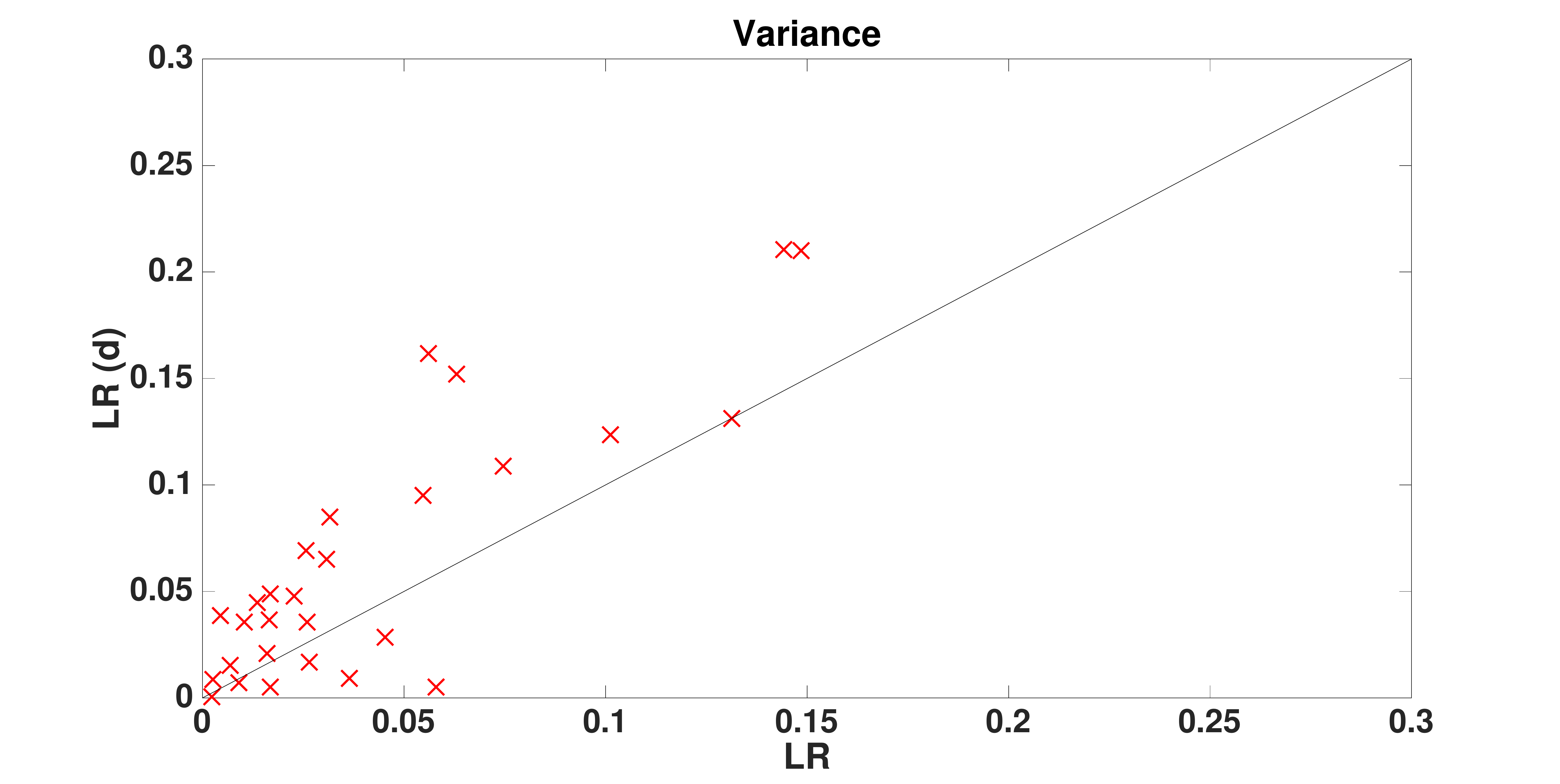}}\hspace{-0.20in}
{\includegraphics[width=55mm,height=50mm]{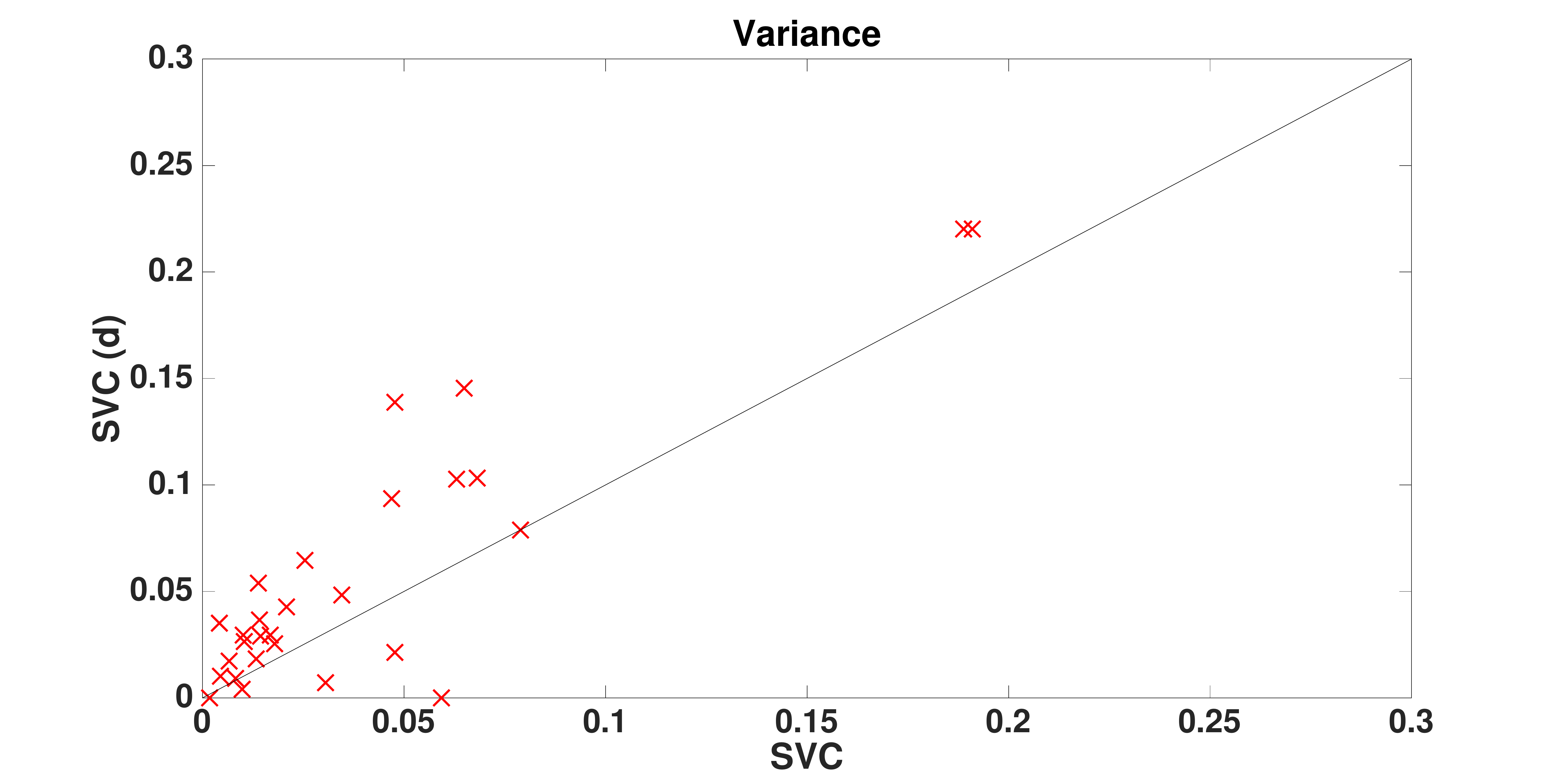}}\hspace{-0.20in}
{\includegraphics[width=55mm,height=50mm]{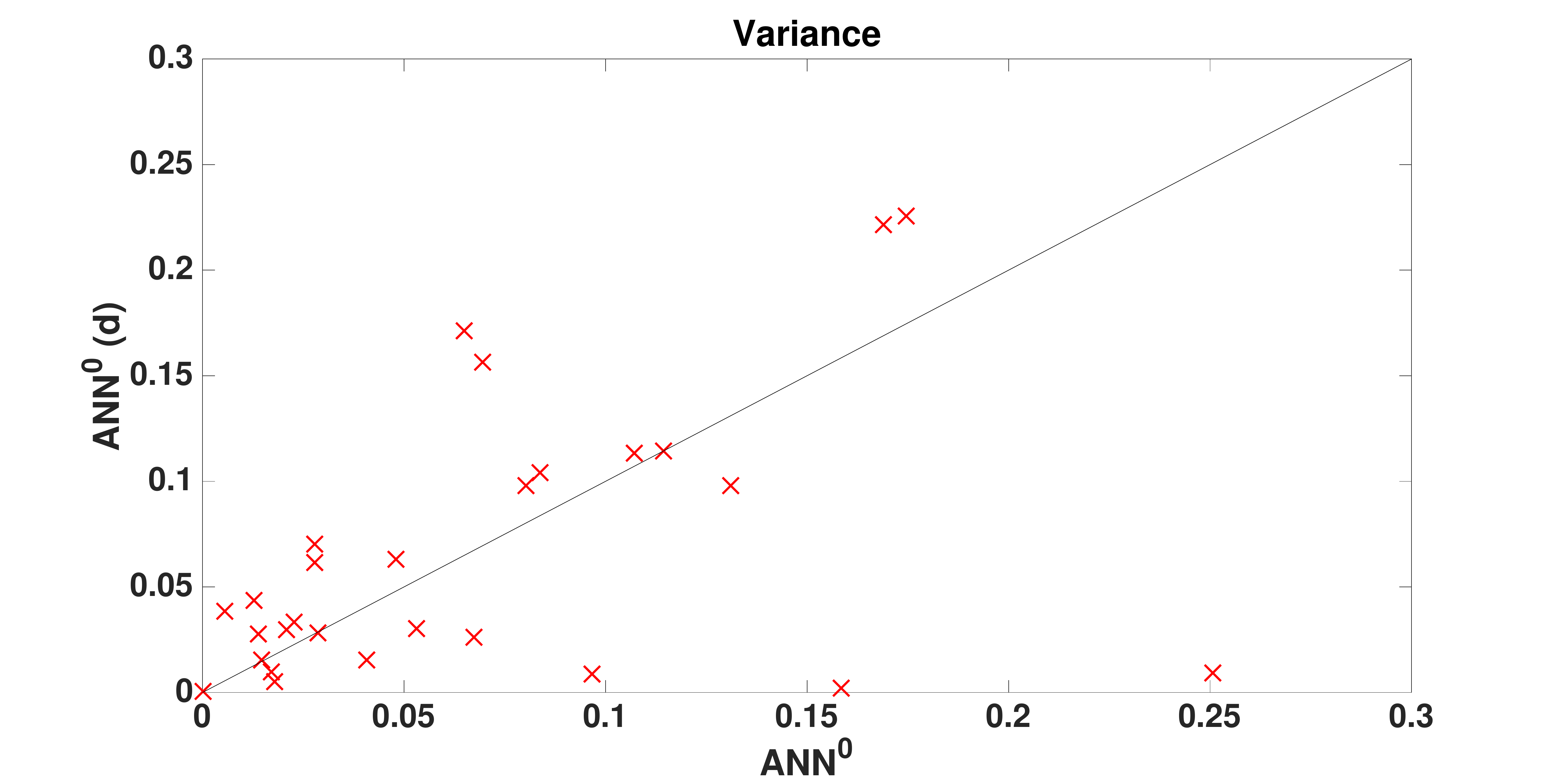}}\hspace{-0.20in}
\vspace{-0.15in}
\caption{\small 
Comparative scatter of \texttt{Variance} results for $\LC$ and $\dLC$ classifiers. $\LC$ are on the X-axis whereas $\dLC$ classifiers are on the Y-axis. 
For points below the diagonal line, $\dLC$ classifiers win. 
}
\label{fig_LRvsdLR_Var}
\end{figure}

Figures~\ref{fig_LRvsdLR_Bias} and~\ref{fig_LRvsdLR_Var} present scatter plots of the bias and variance of $\LC$ and $\dLC$ classifiers. 
It can be seen that the three $\dLC$ classifiers lead to low-bias and high-variance models. 
Note that we present a bias-variance analysis on only \emph{Little}  datasets. This is because the software we have for obtaining bias-variance estimates is single threaded and could not benefit from the high-performance environment in which most of our experiments were run.
As a result it was not feasible to run these experiments on the larger datasets. 
Nonetheless, results confirm our hypothesis that $\dLC$ classifiers tend to have lower bias than $\LC$. 

\subsection{Comparison of the Convergence Curves of $\dLC$ and $\LC$}

As training of both $\dLC$ and $\LC$ classifiers are based on iterative optimization algorithms, they produce a sequence of values as part of their training, i.e., of their objective function which (should ideally) decrease with successive iterations until convergence. 
A technique that leads to the global minimum faster (steeper curve) and in fewer iterations (shorter curve) is desirable.
Note that  $\LC$ and $\dLC$ have different models (and parameterizations) and, therefore, the optimization space for the two problems is also very different.
In the following, let us compare the convergences of $\LC$ and $\dLC$ on some sample datasets. A similar trend was observed on all datasets, here we report results on nine representative datasets only.

\begin{figure}[t]
\centering
\hspace{-0.1in}
\includegraphics[width=50mm,height=40mm]{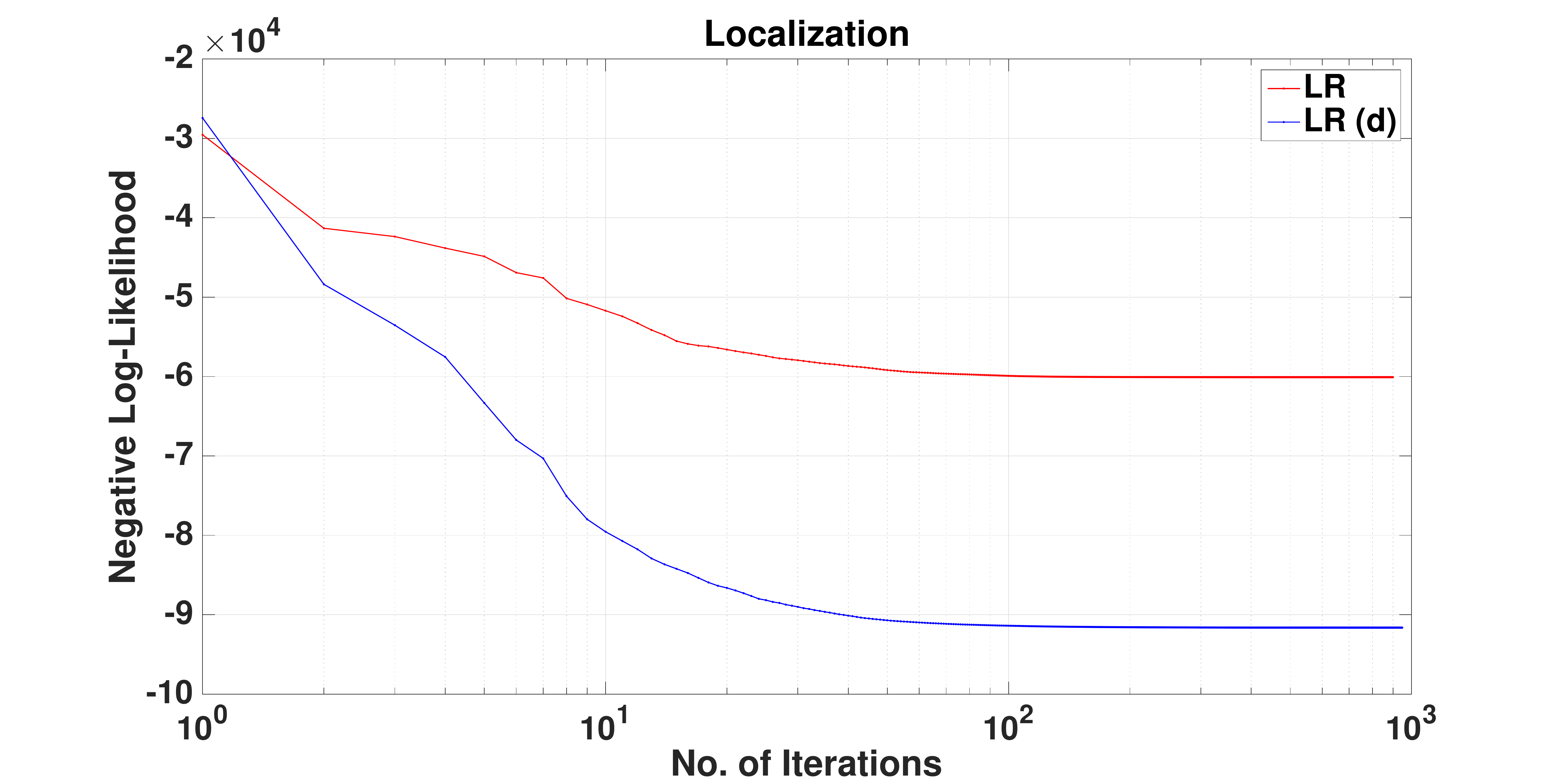}
\includegraphics[width=50mm,height=40mm]{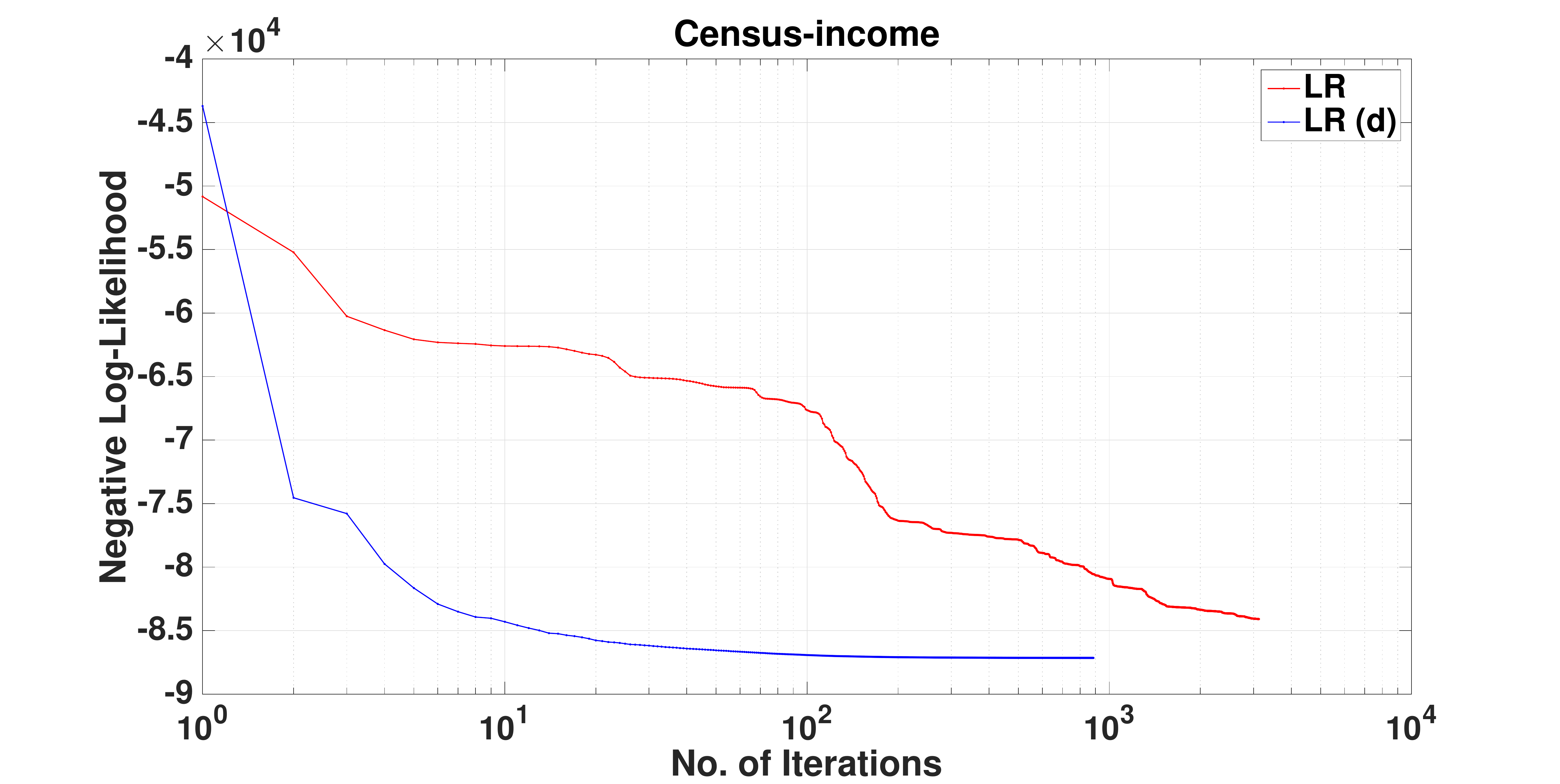}
\includegraphics[width=50mm,height=40mm]{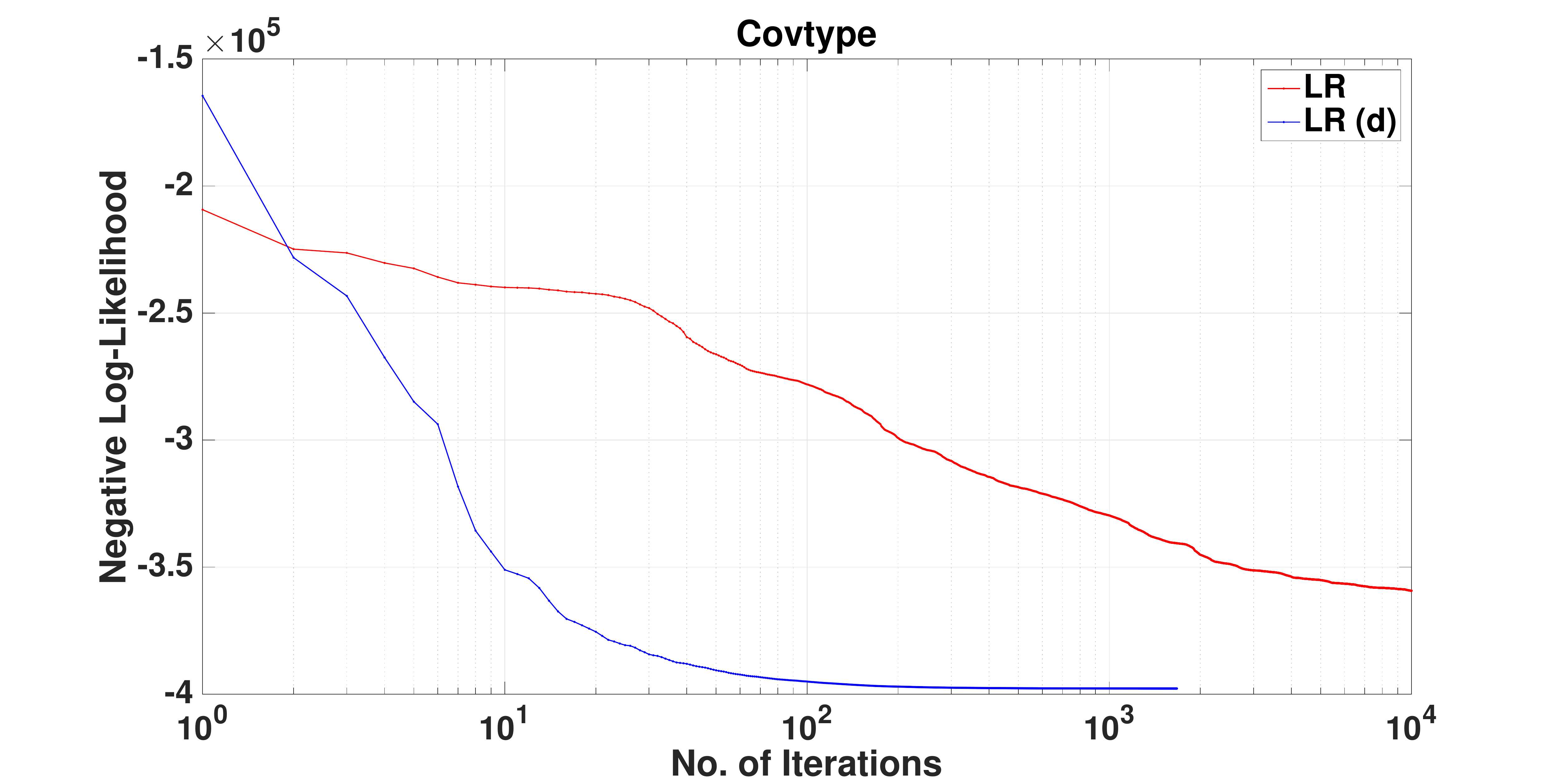}

\includegraphics[width=50mm,height=40mm]{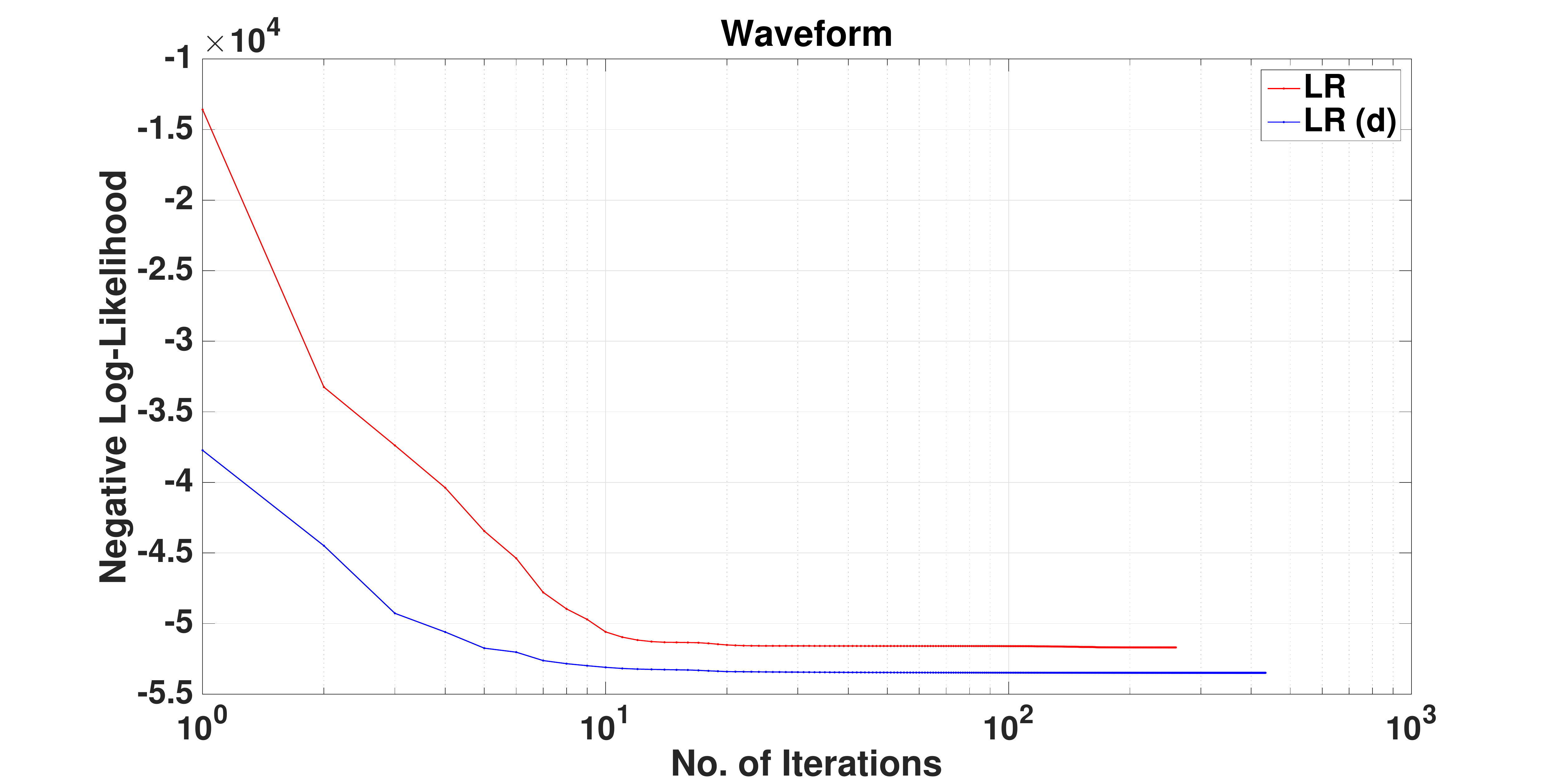}
\includegraphics[width=50mm,height=40mm]{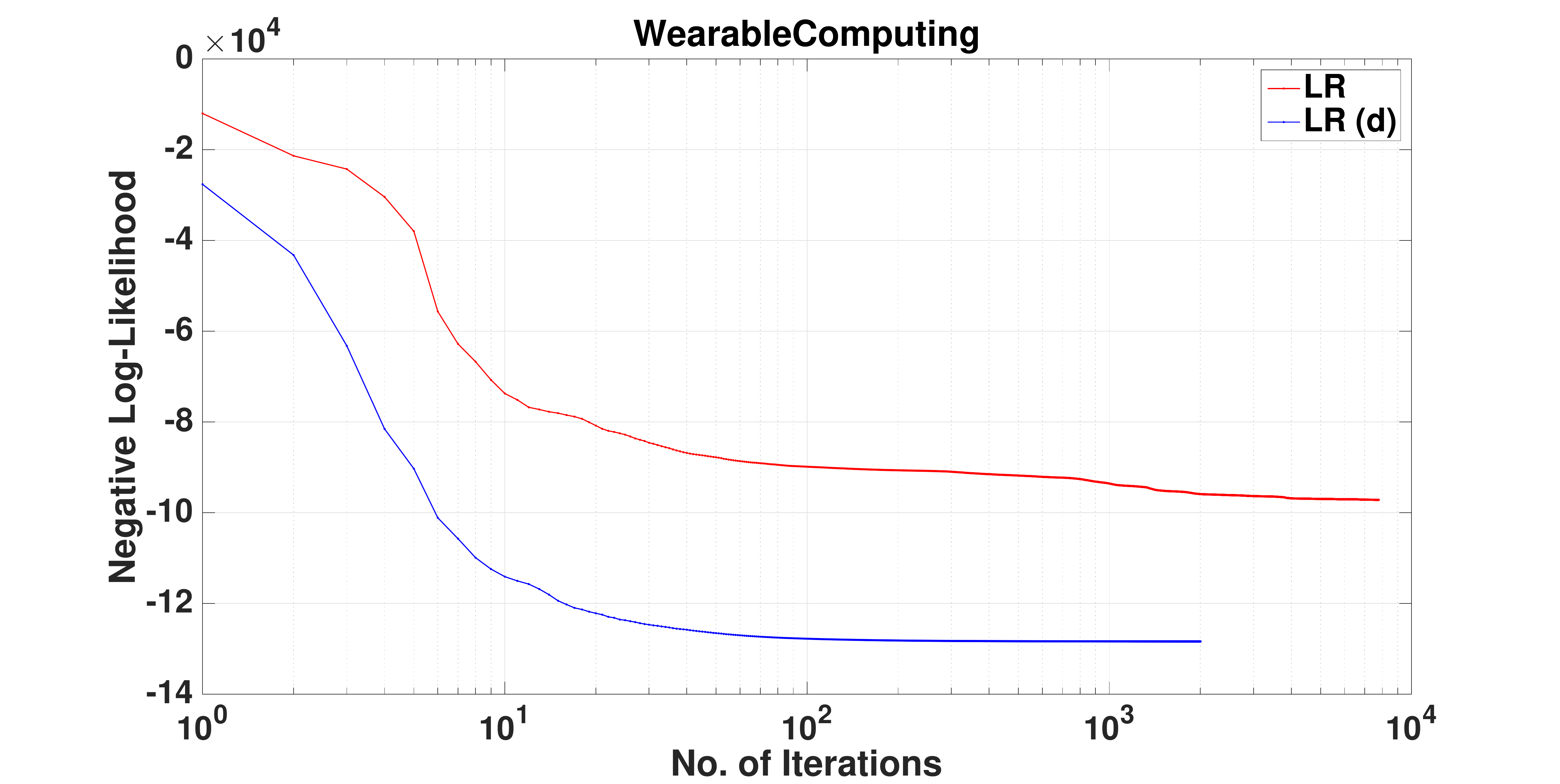}
\includegraphics[width=50mm,height=40mm]{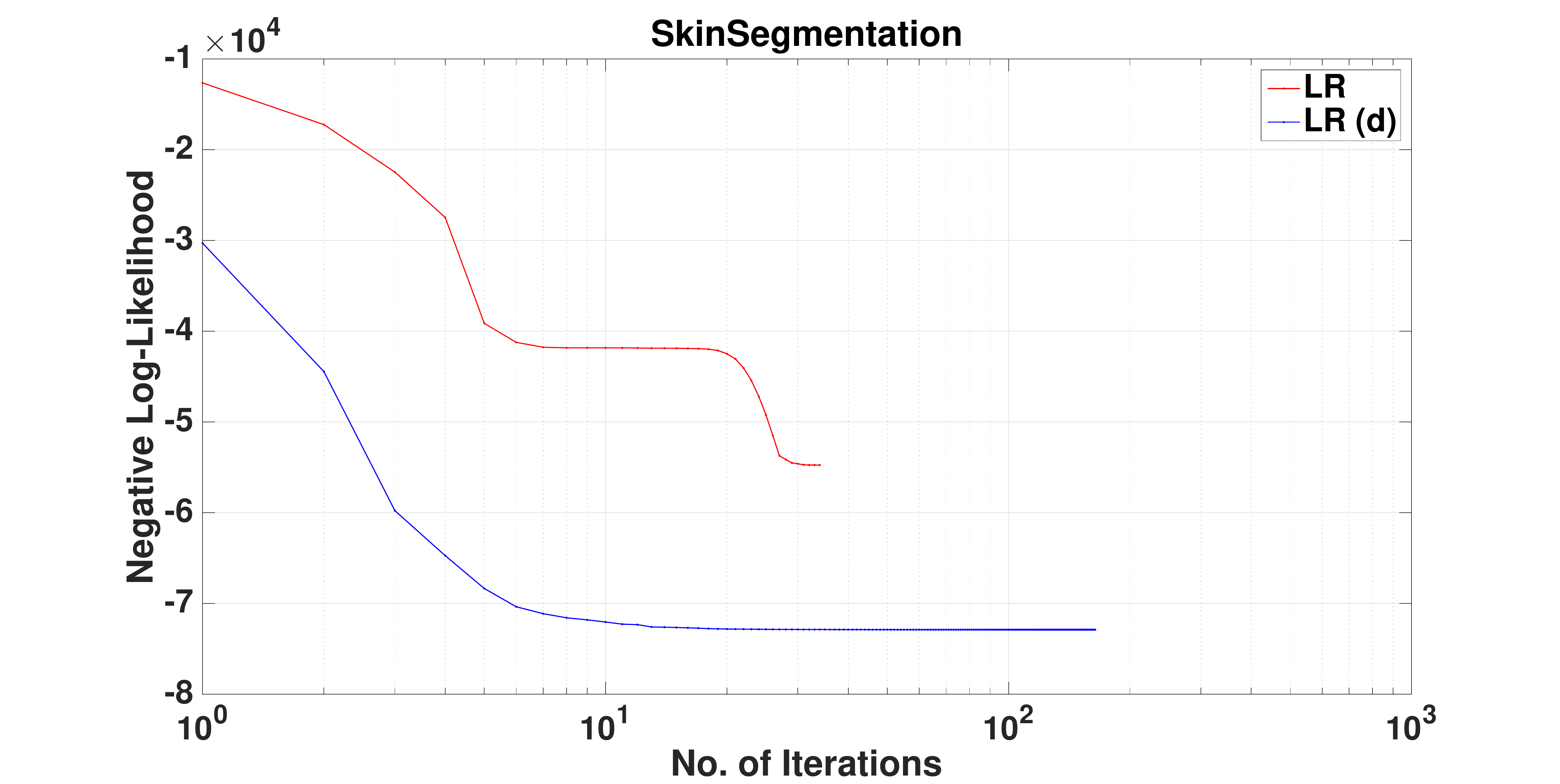}

\includegraphics[width=50mm,height=40mm]{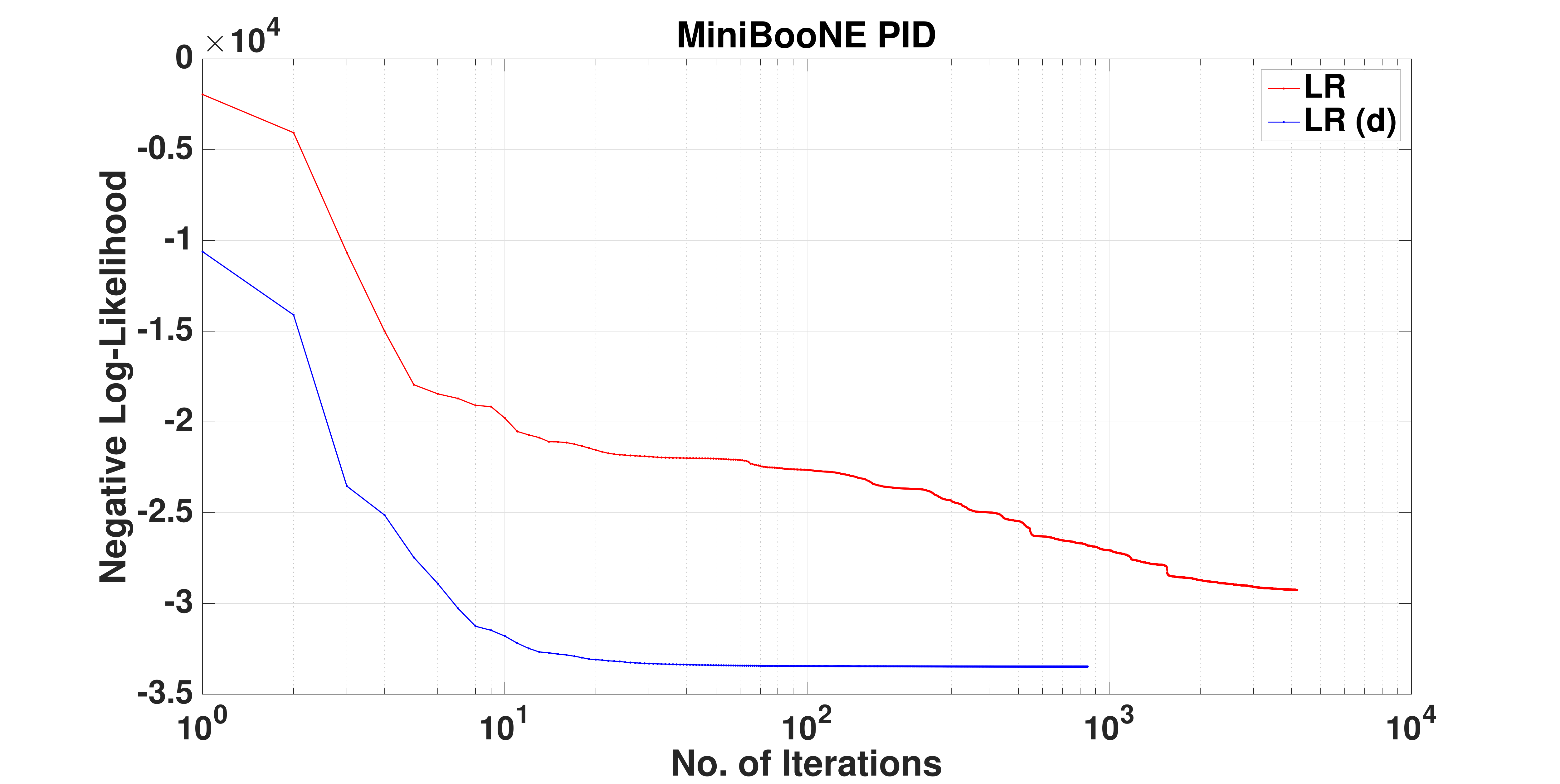}
\includegraphics[width=50mm,height=40mm]{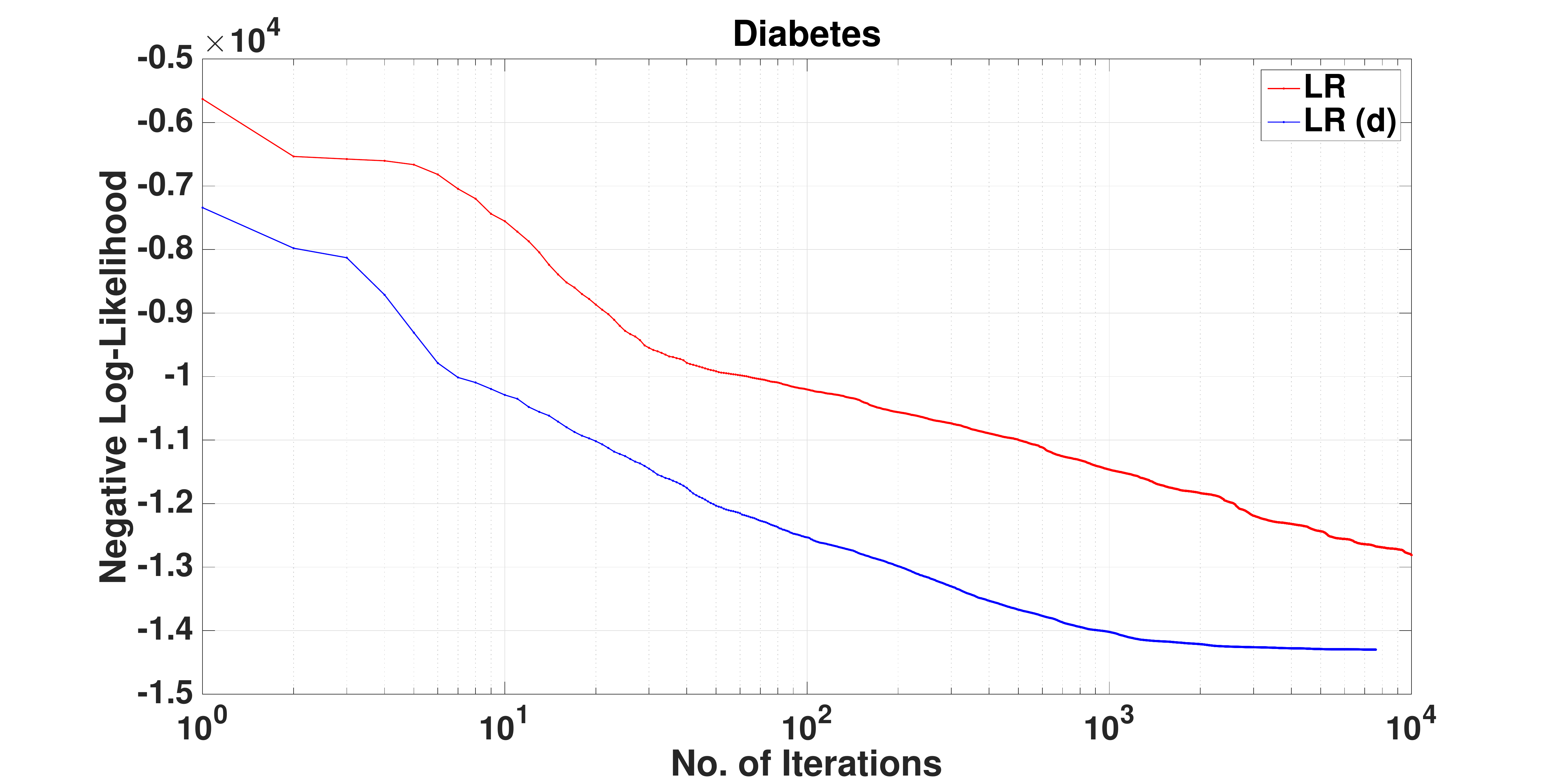}
\includegraphics[width=50mm,height=40mm]{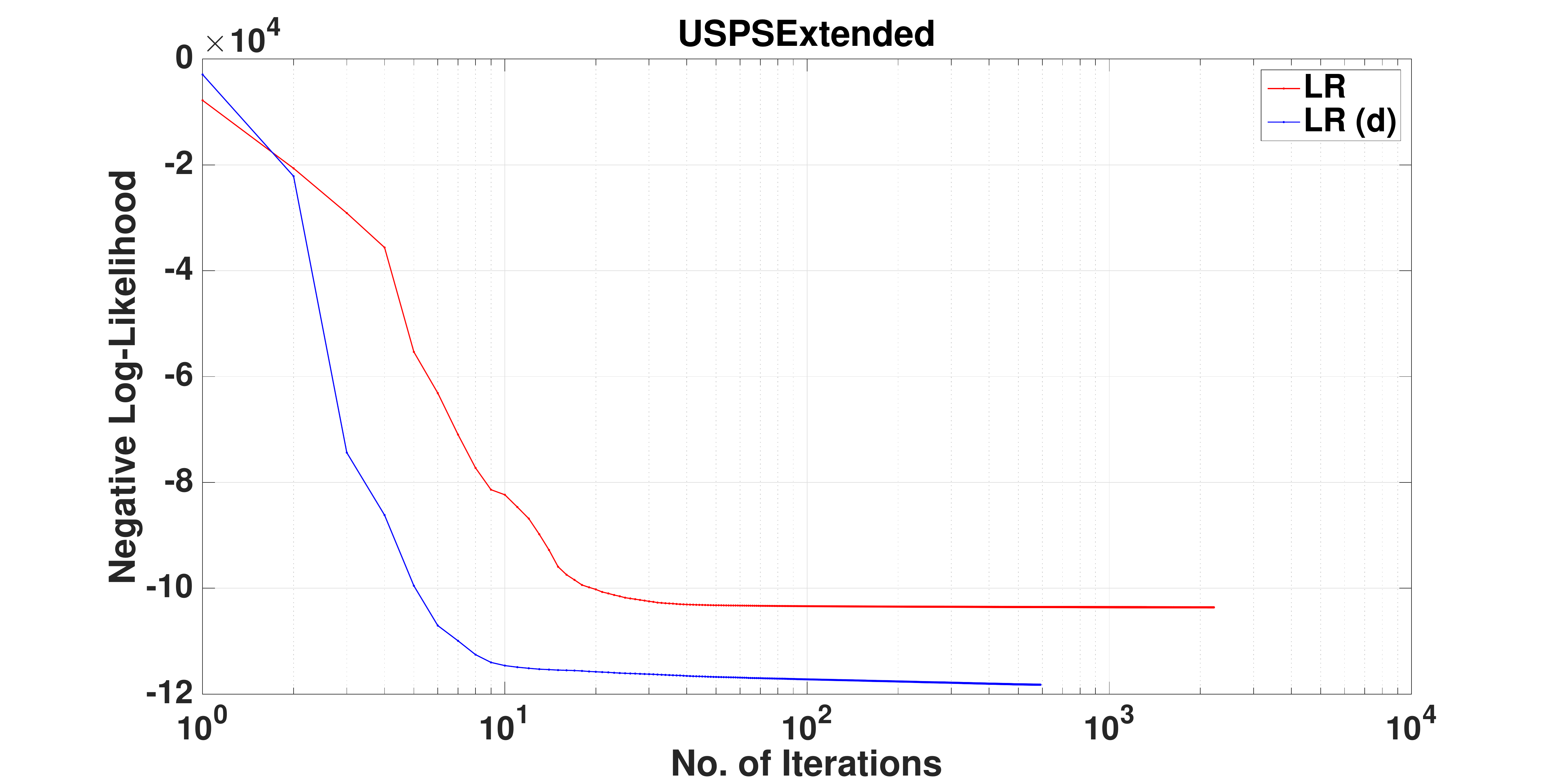}

\caption{\small Comparison of the rate of convergence of LR and LR(d) on nine sample datasets. The X-axis is on log scale.}
\label{fig_CC_CLL}
\end{figure}

A comparison of the variation in NLL objective function for LR and LR(d) is shown in Figure~\ref{fig_CC_CLL}. It can be seen that LR(d) has steeper curve -- that is, it asymptotes to its global minimum much quickly. It is also important to see that LR(d) leads to much lower NLL. Better accuracy of LR(d) is the result of this much lower NLL.

\begin{figure}[t]
\centering
\hspace{-0.1in}
\includegraphics[width=50mm,height=40mm]{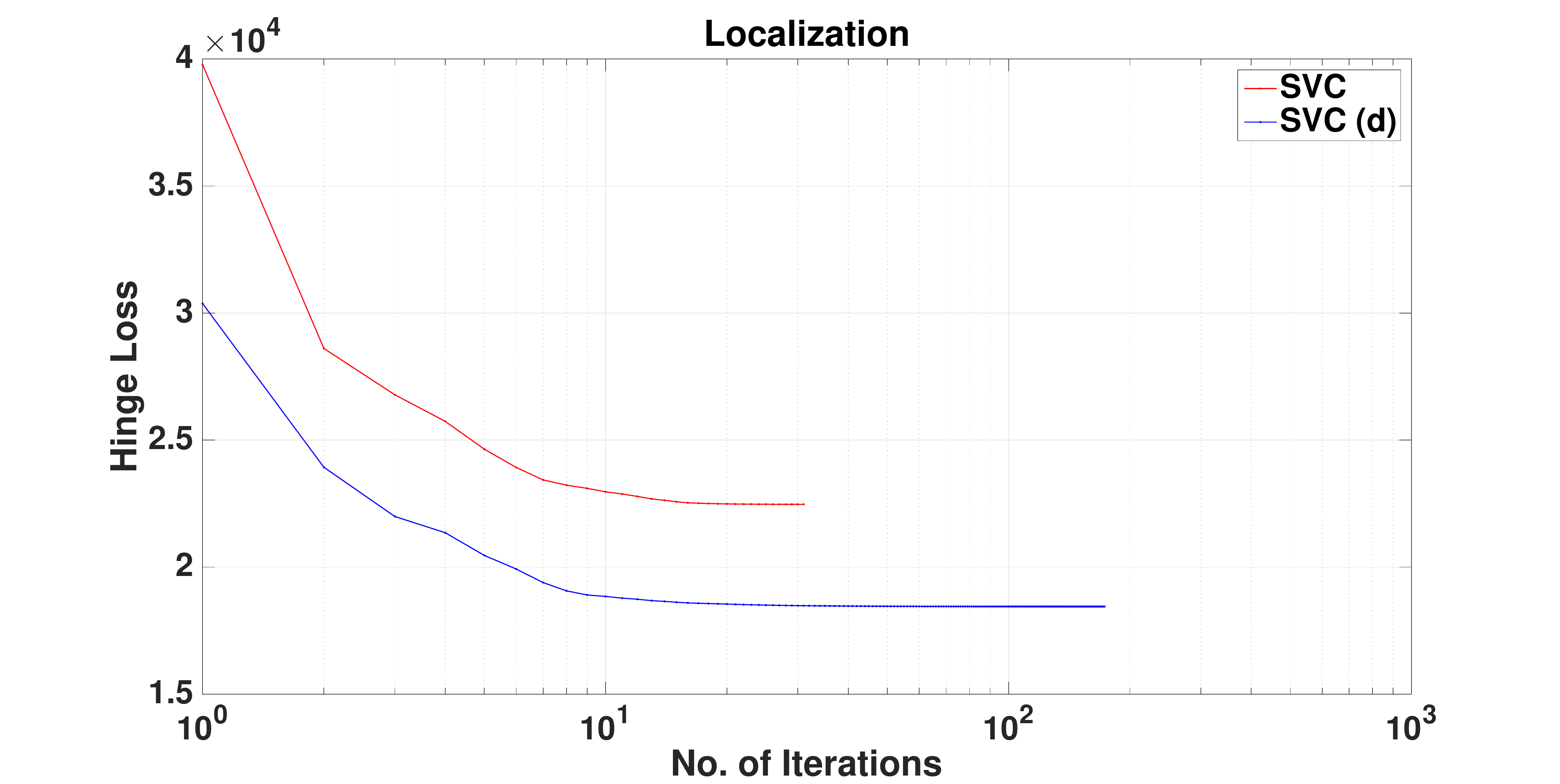}
\includegraphics[width=50mm,height=40mm]{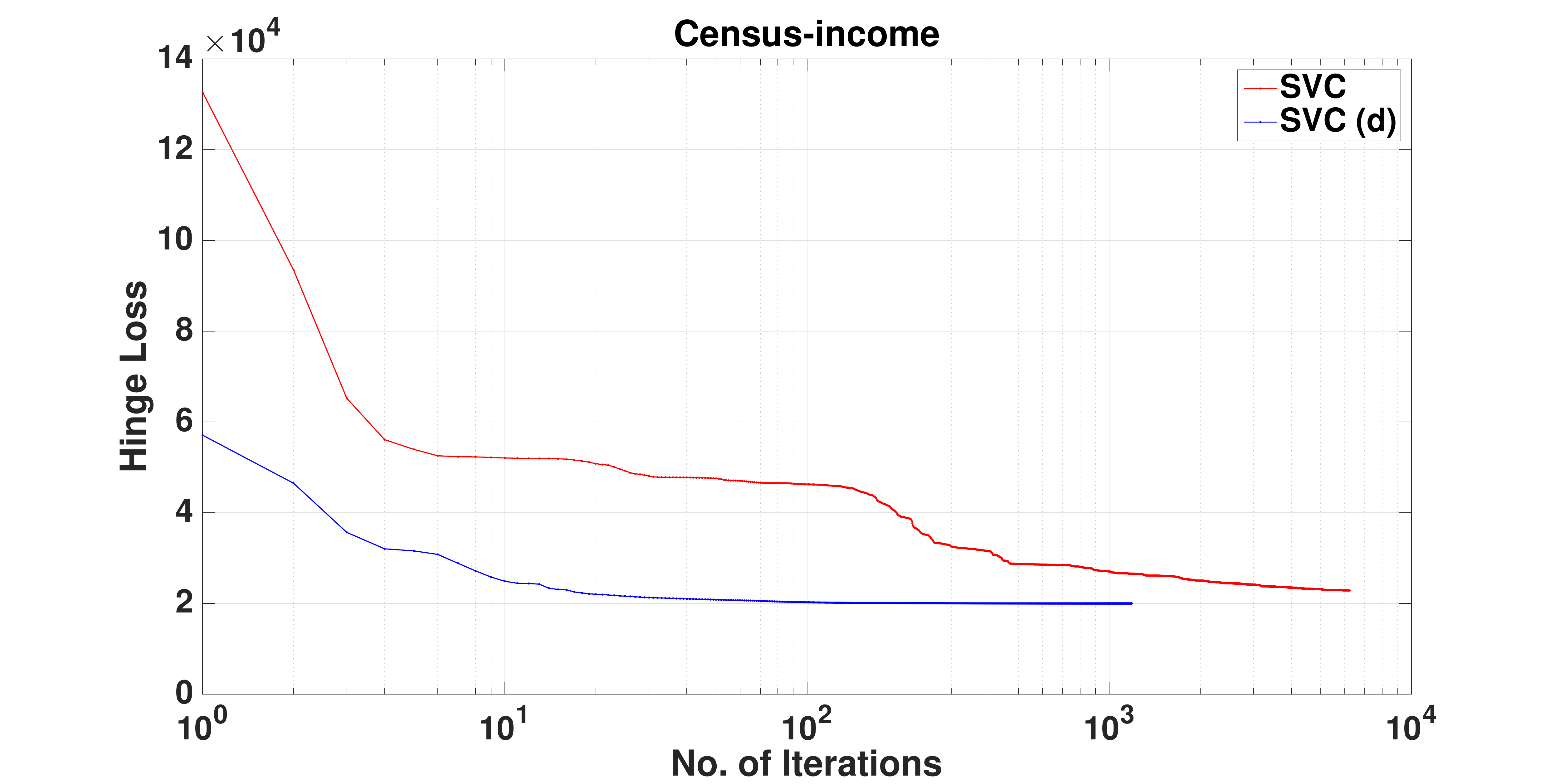}
\includegraphics[width=50mm,height=40mm]{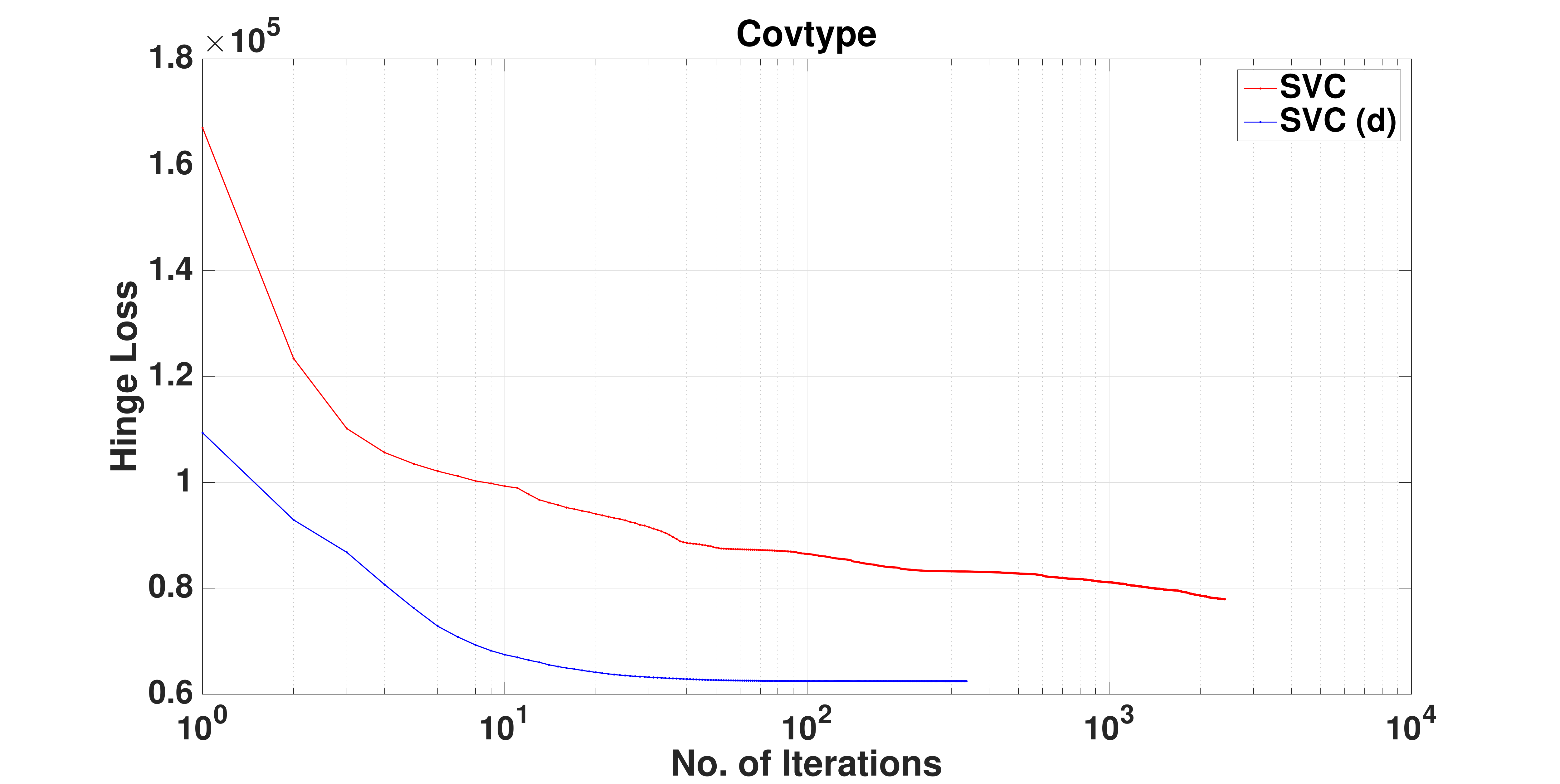}

\includegraphics[width=50mm,height=40mm]{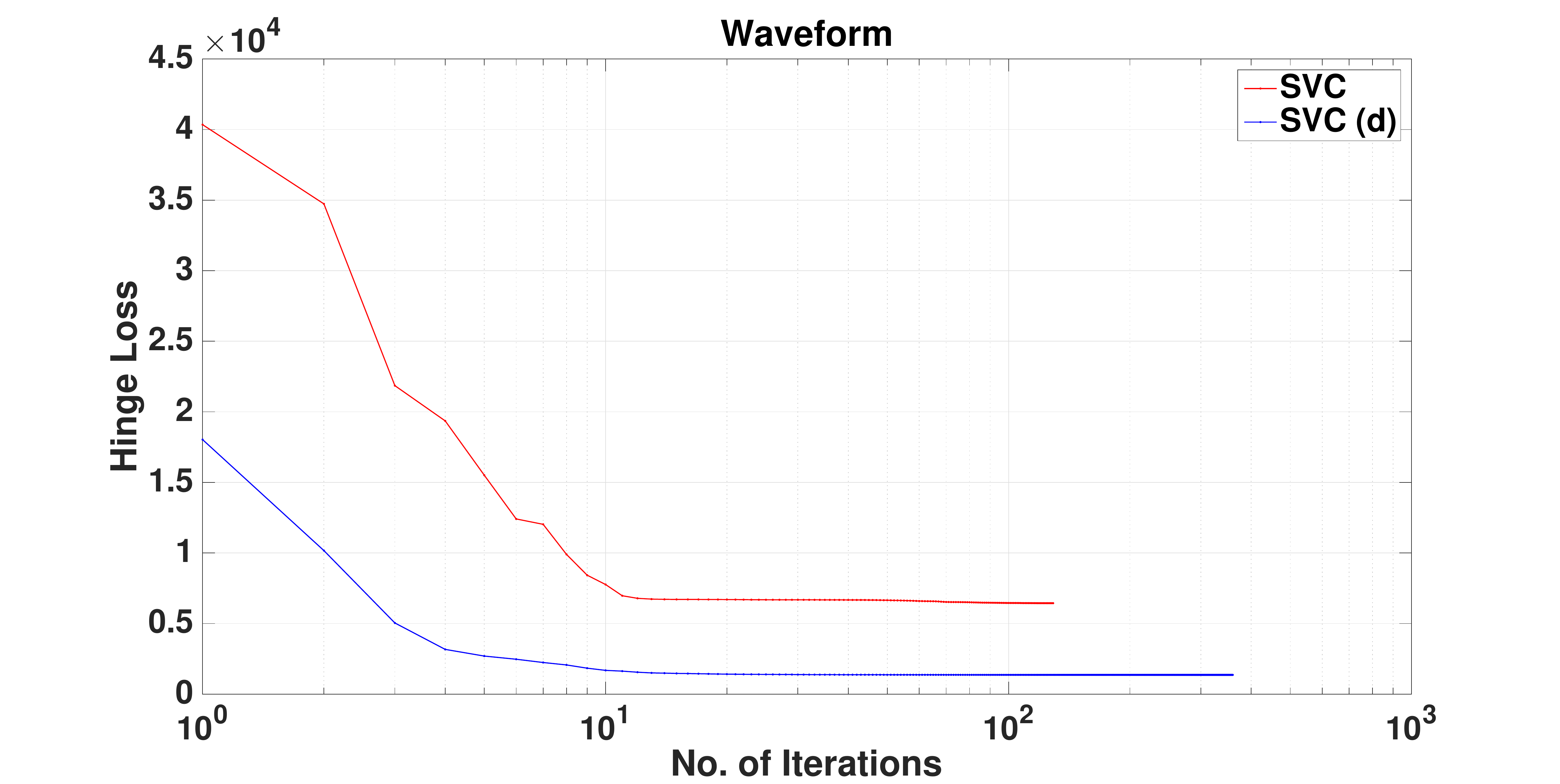}
\includegraphics[width=50mm,height=40mm]{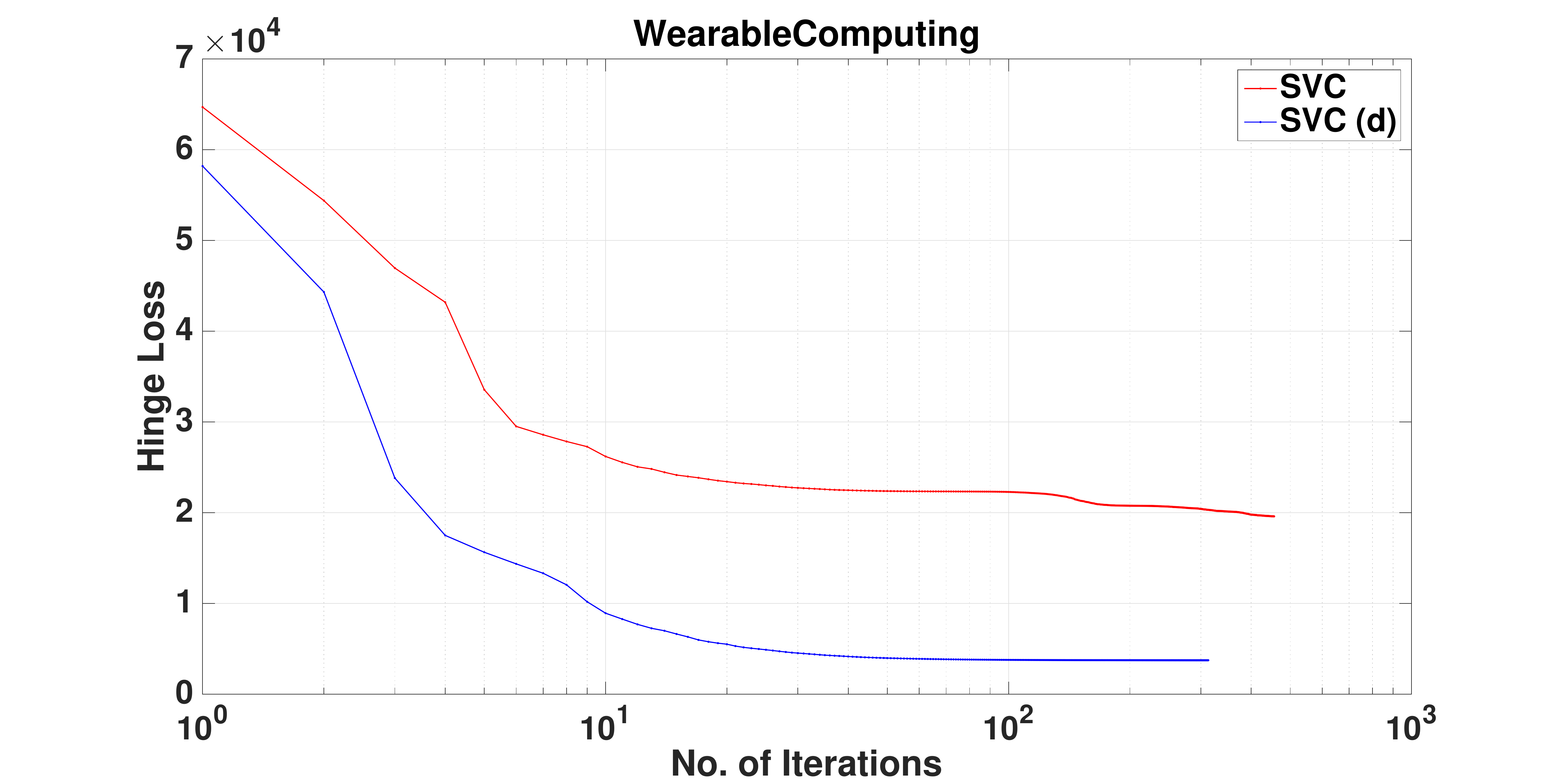}
\includegraphics[width=50mm,height=40mm]{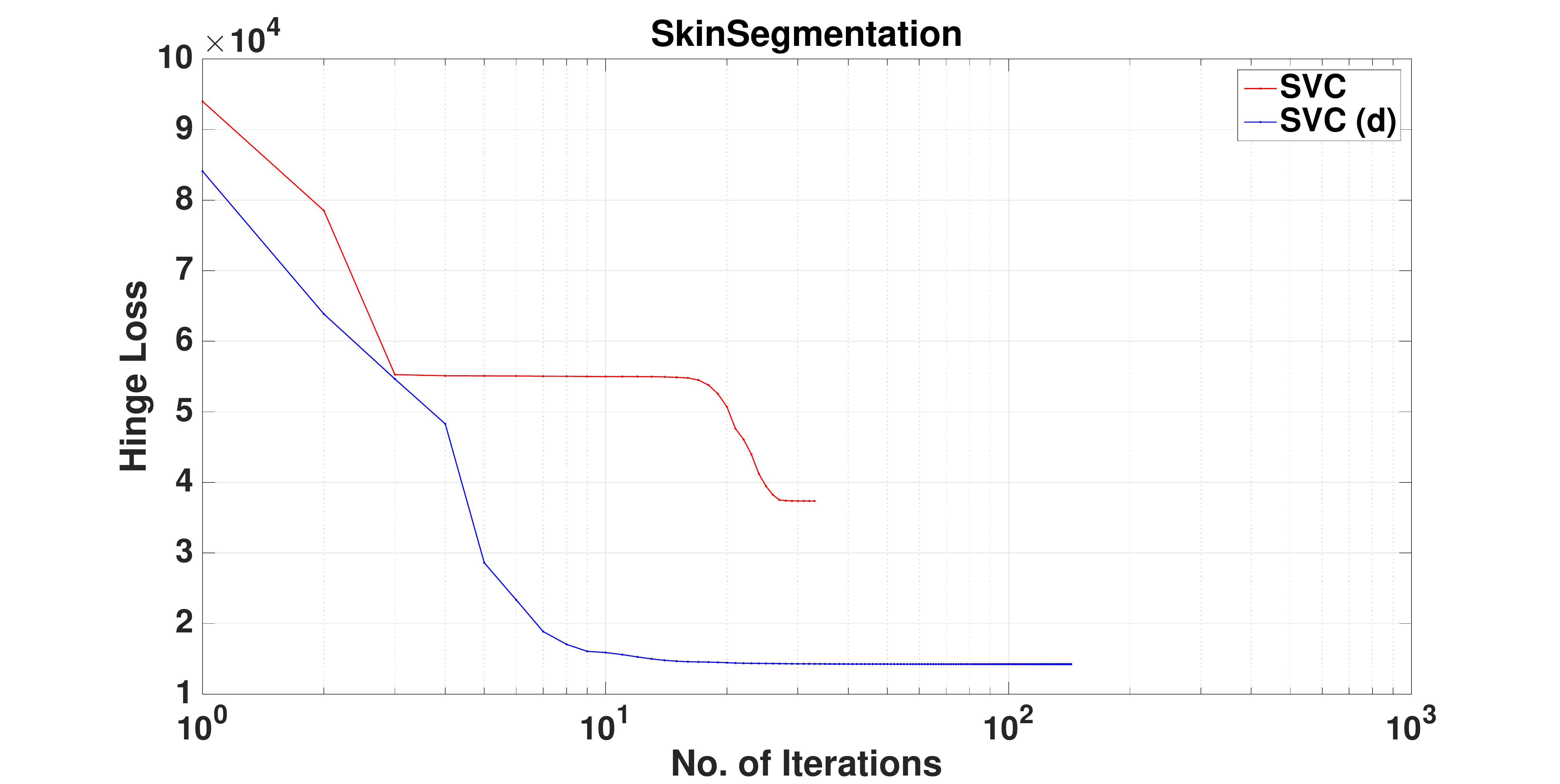}

\includegraphics[width=50mm,height=40mm]{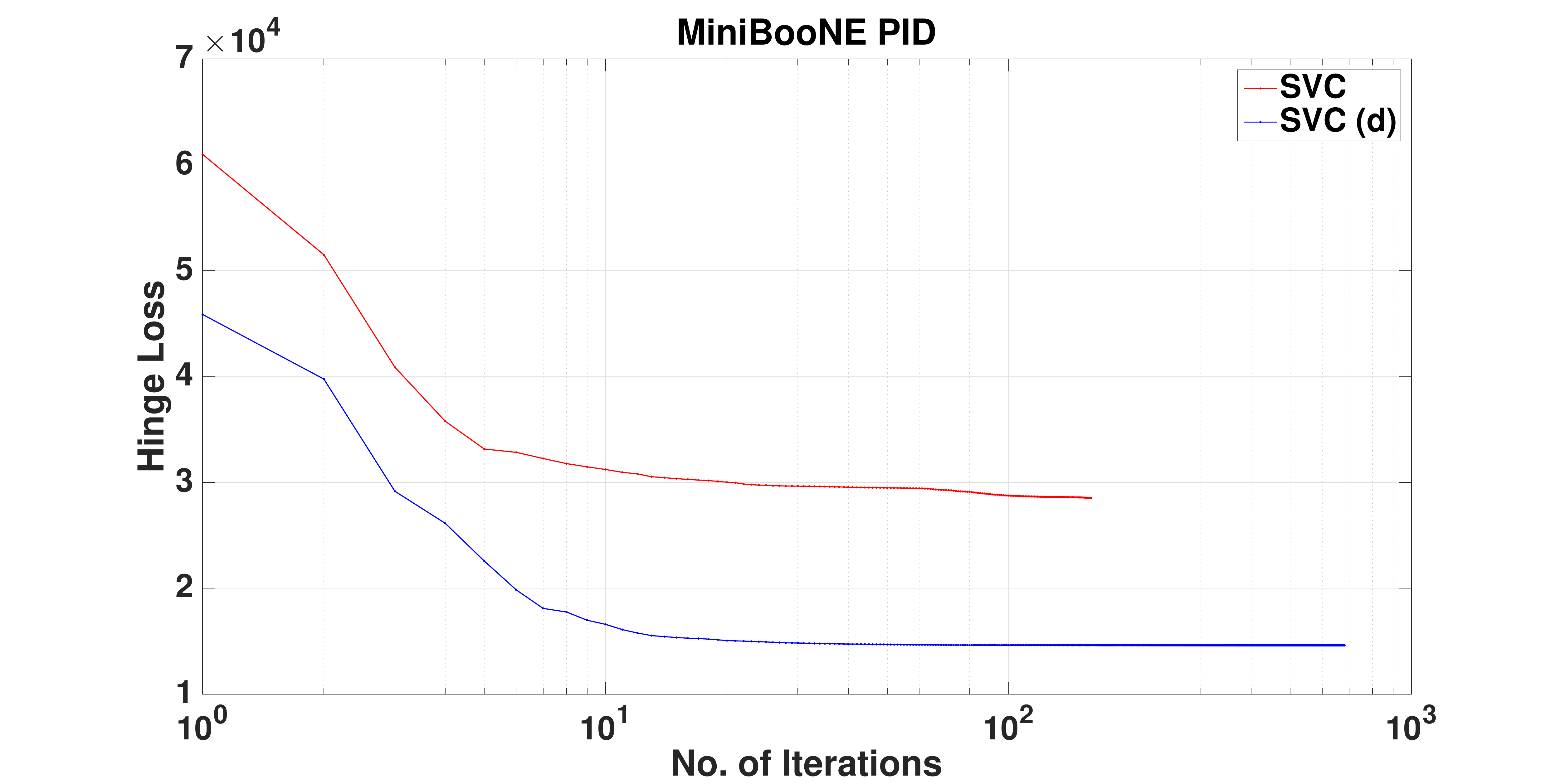}
\includegraphics[width=50mm,height=40mm]{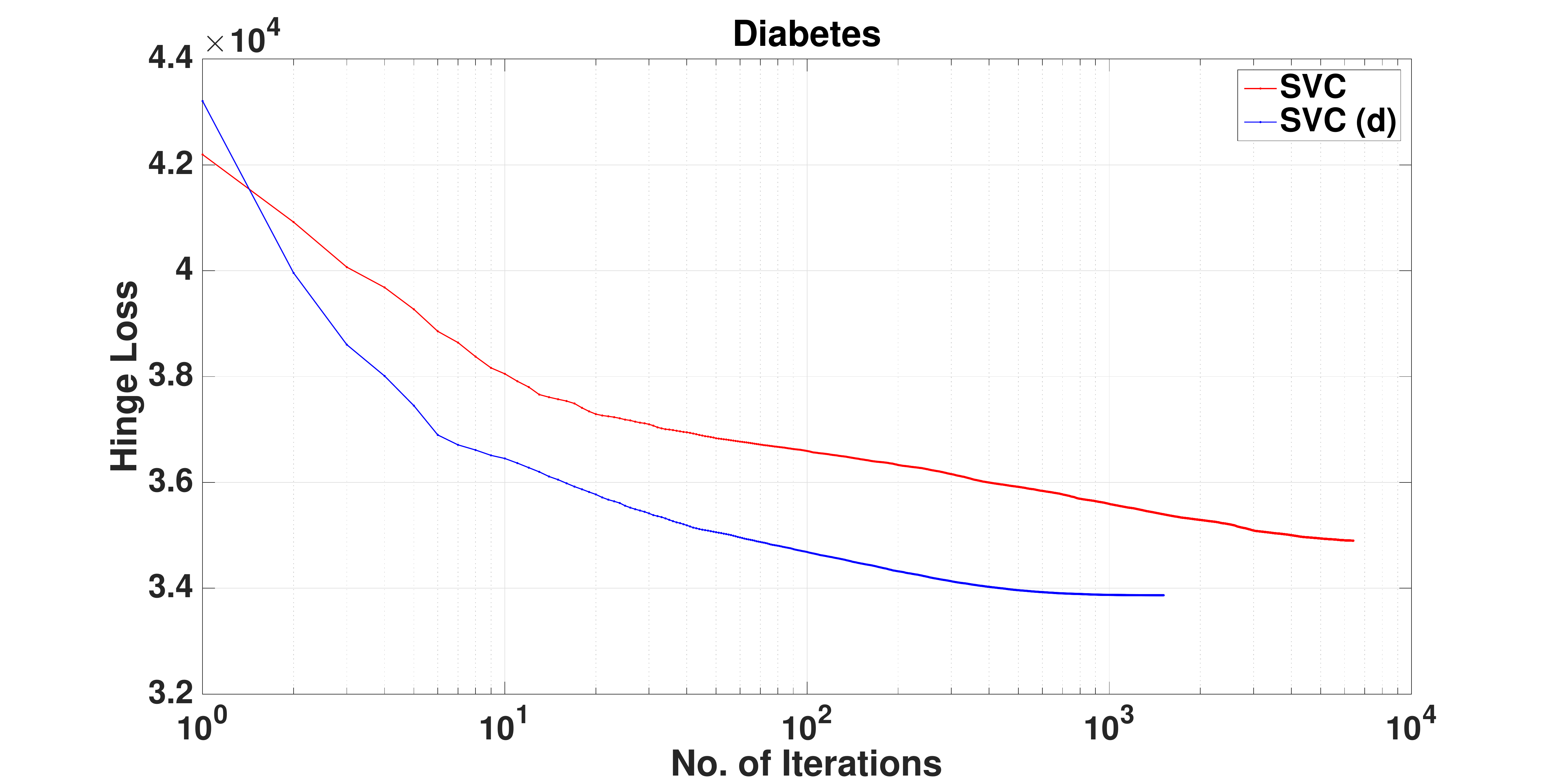}
\includegraphics[width=50mm,height=40mm]{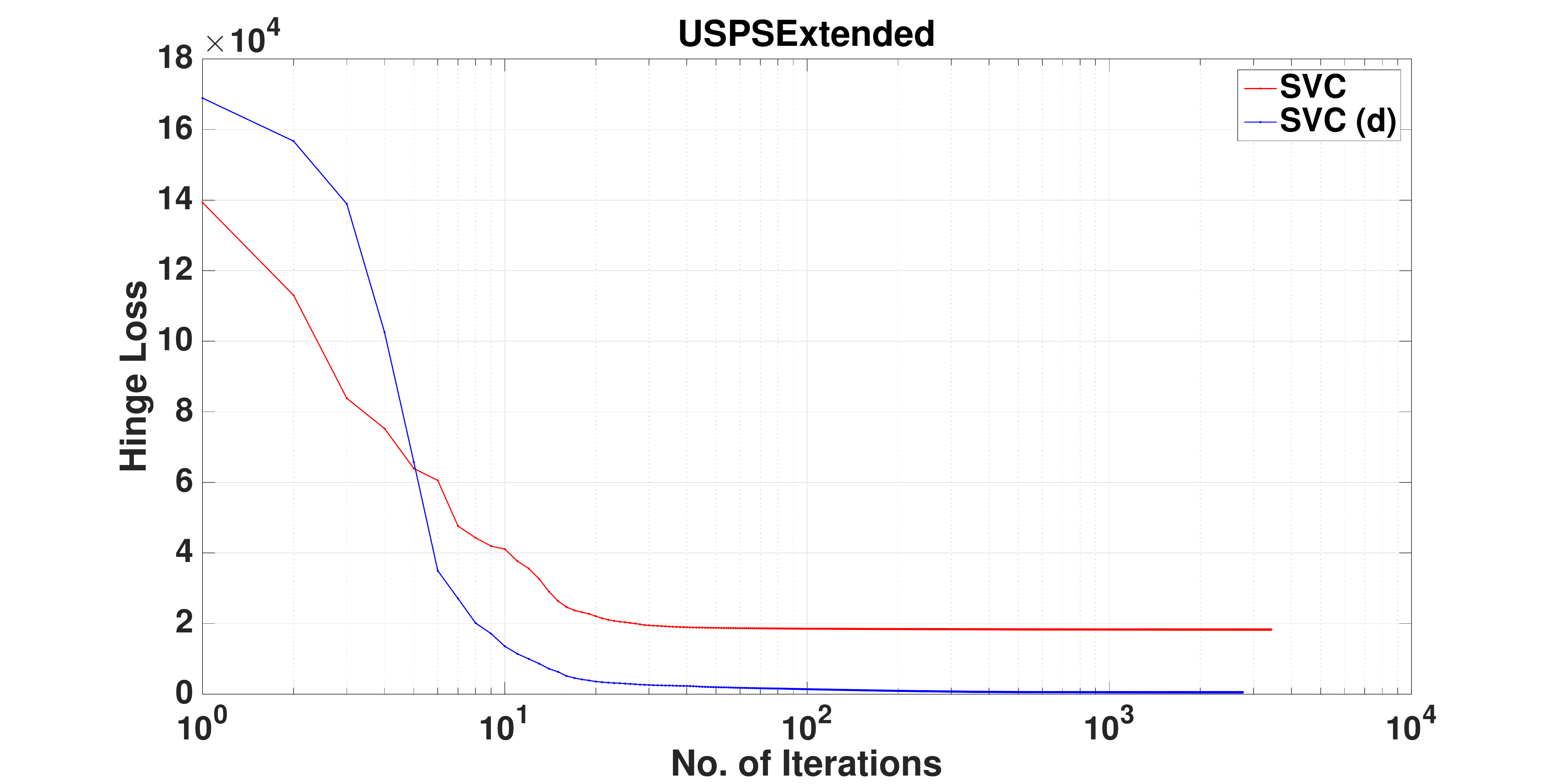}

\caption{\small Comparison of the rate of convergence of SVC and SVC(d) on nine sample datasets. The X-axis is on the log scale.}
\label{fig_CC_HL}
\end{figure}

\begin{figure}[t]
\centering
\hspace{-0.1in}
\includegraphics[width=50mm,height=40mm]{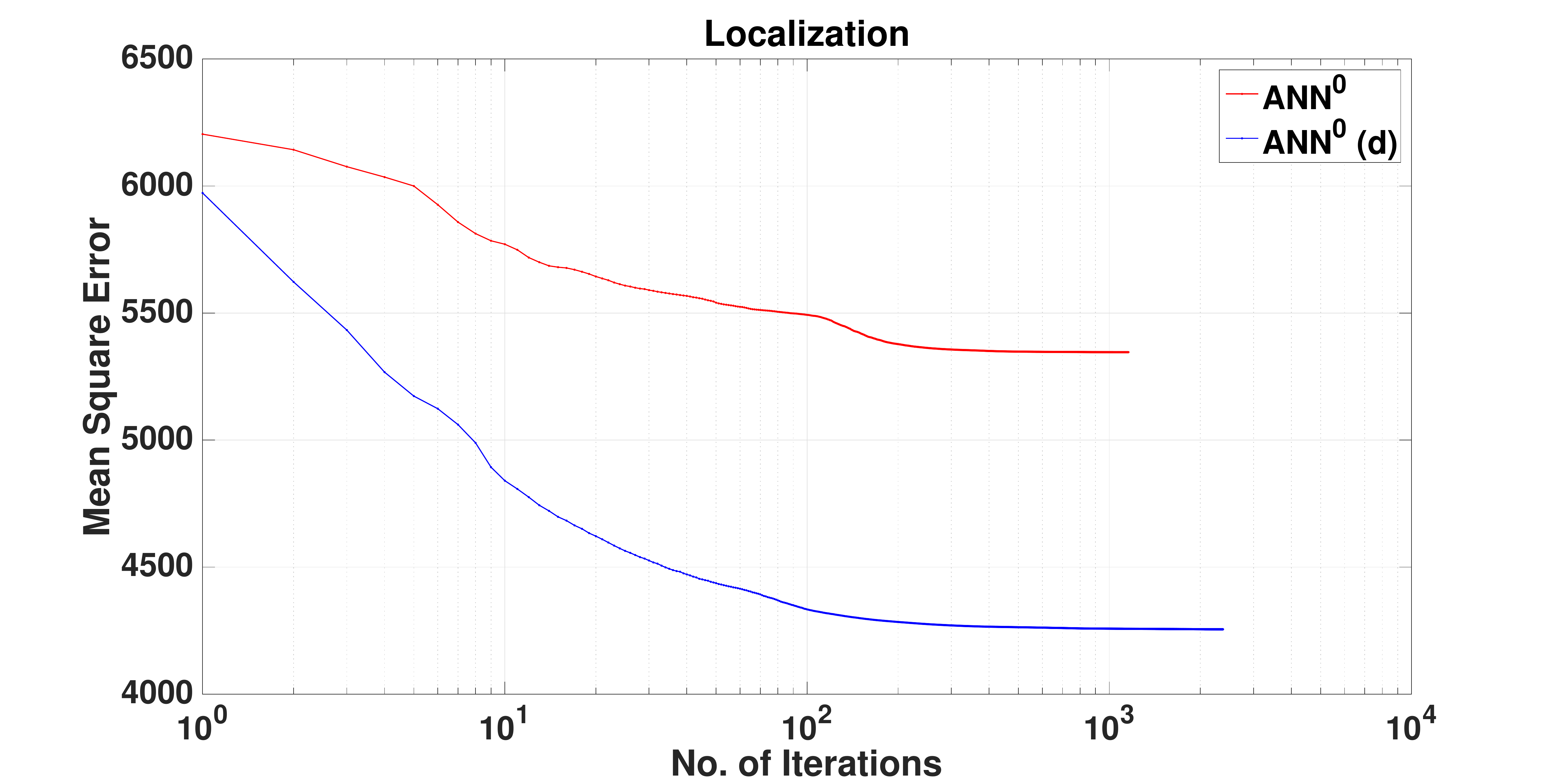}
\includegraphics[width=50mm,height=40mm]{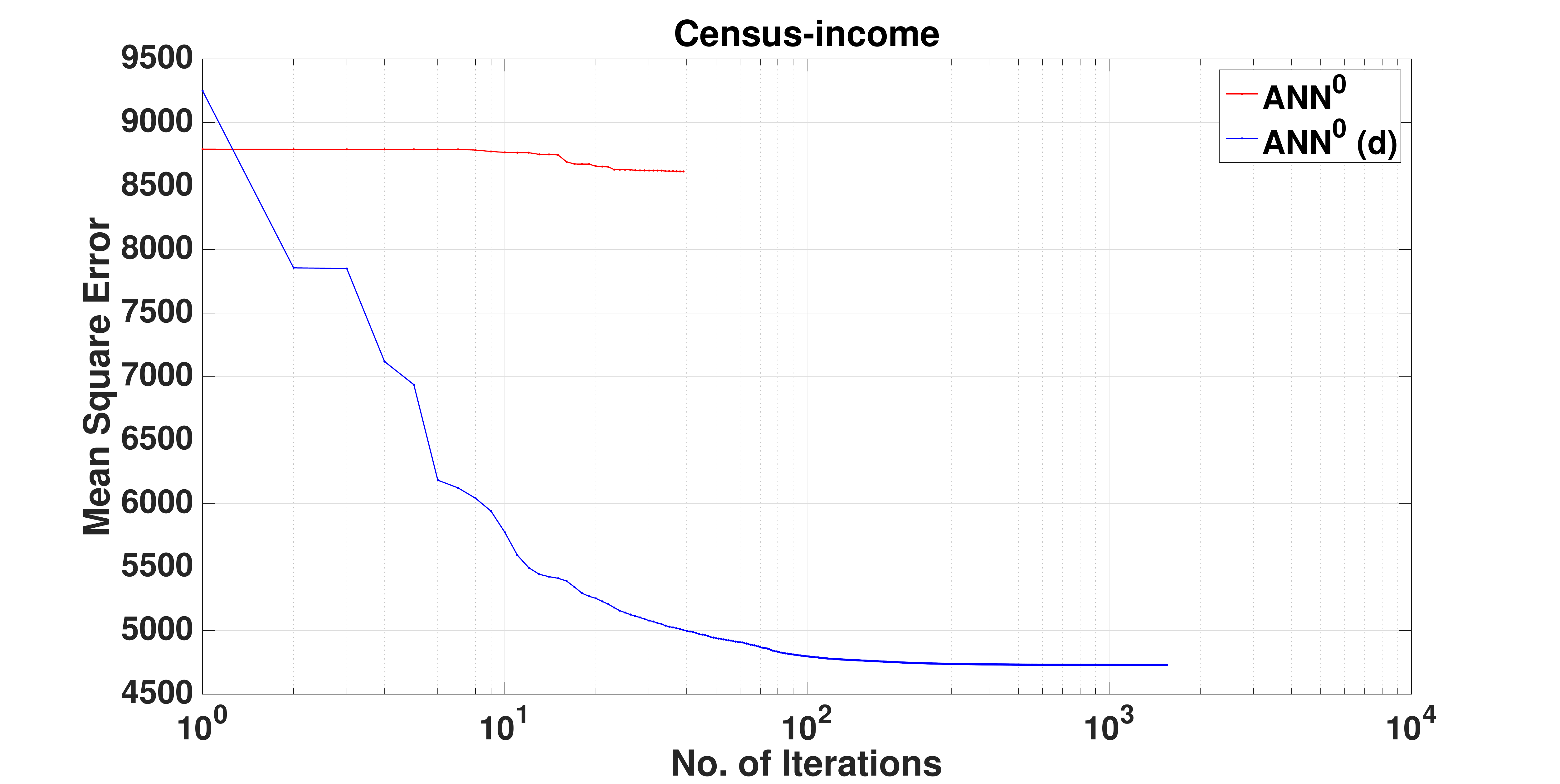}
\includegraphics[width=50mm,height=40mm]{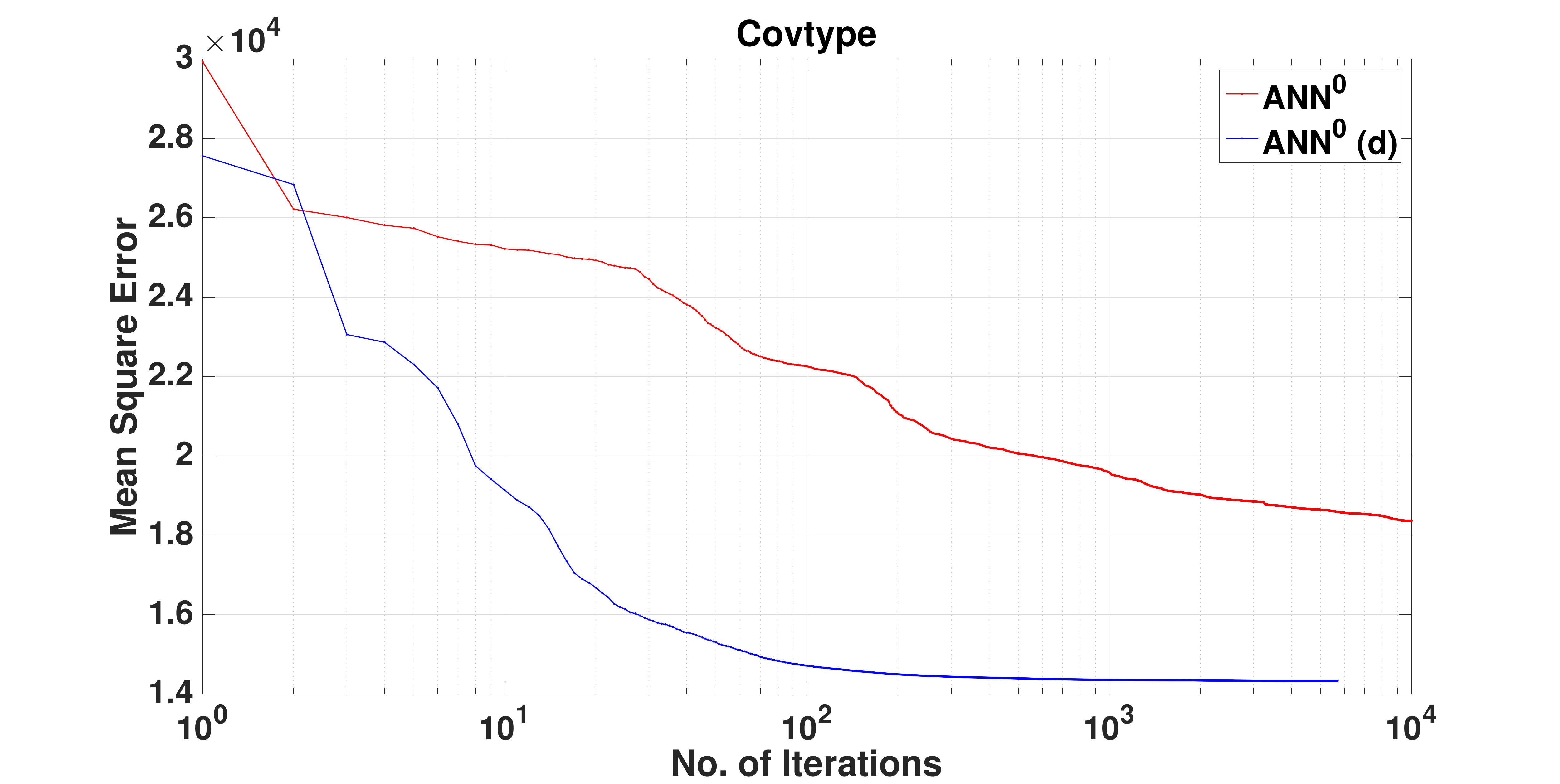}

\includegraphics[width=50mm,height=40mm]{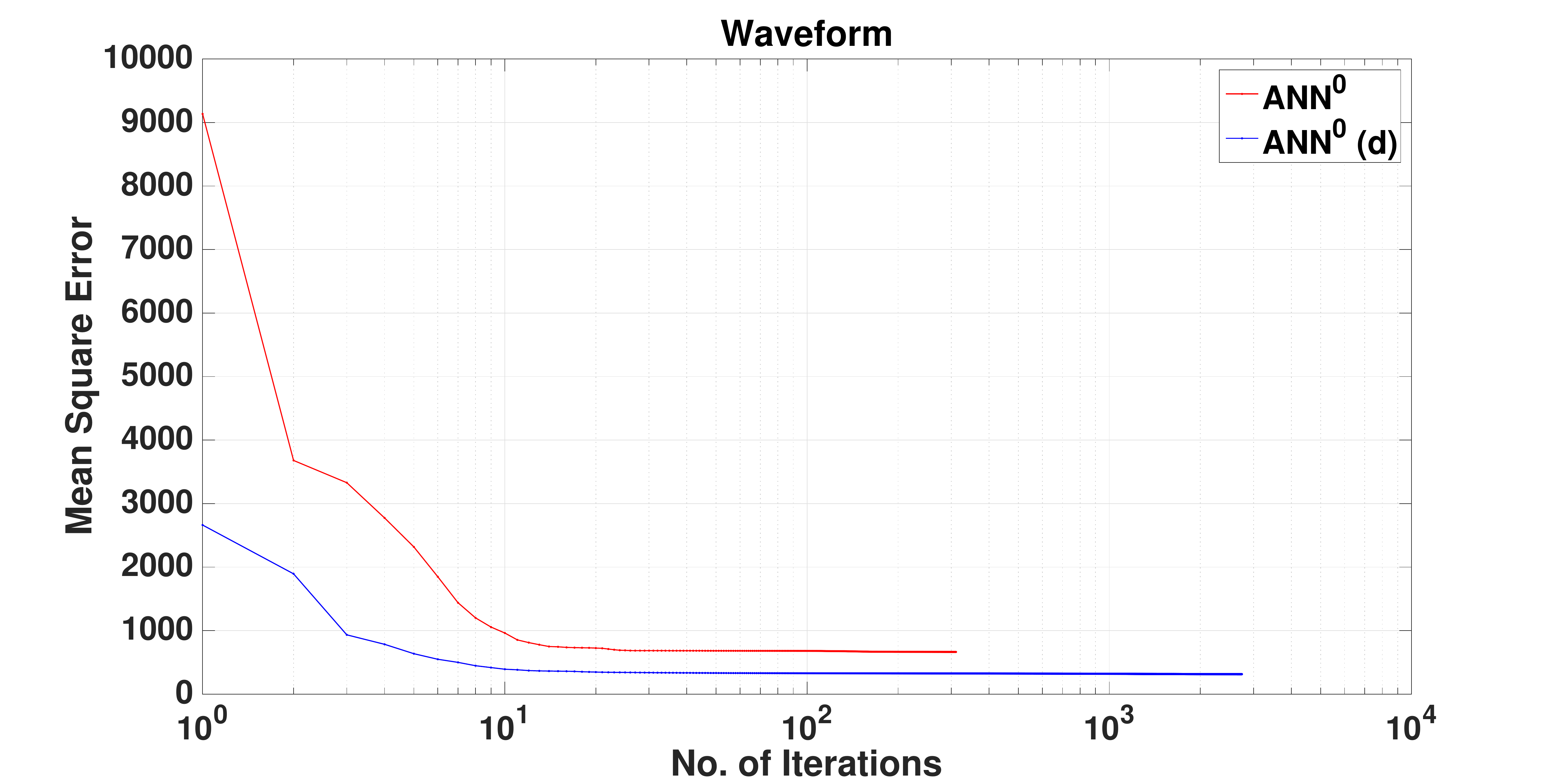}
\includegraphics[width=50mm,height=40mm]{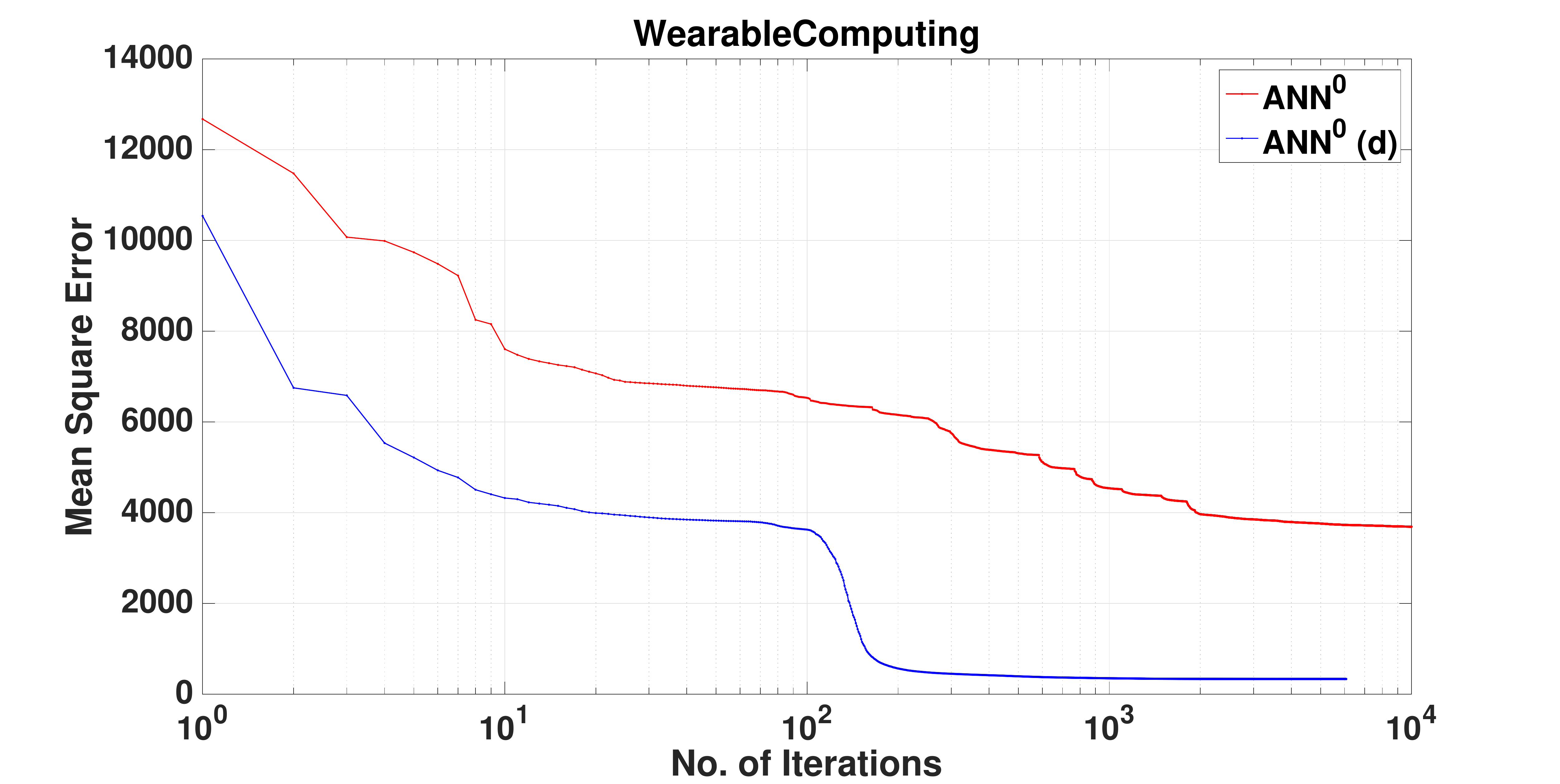}
\includegraphics[width=50mm,height=40mm]{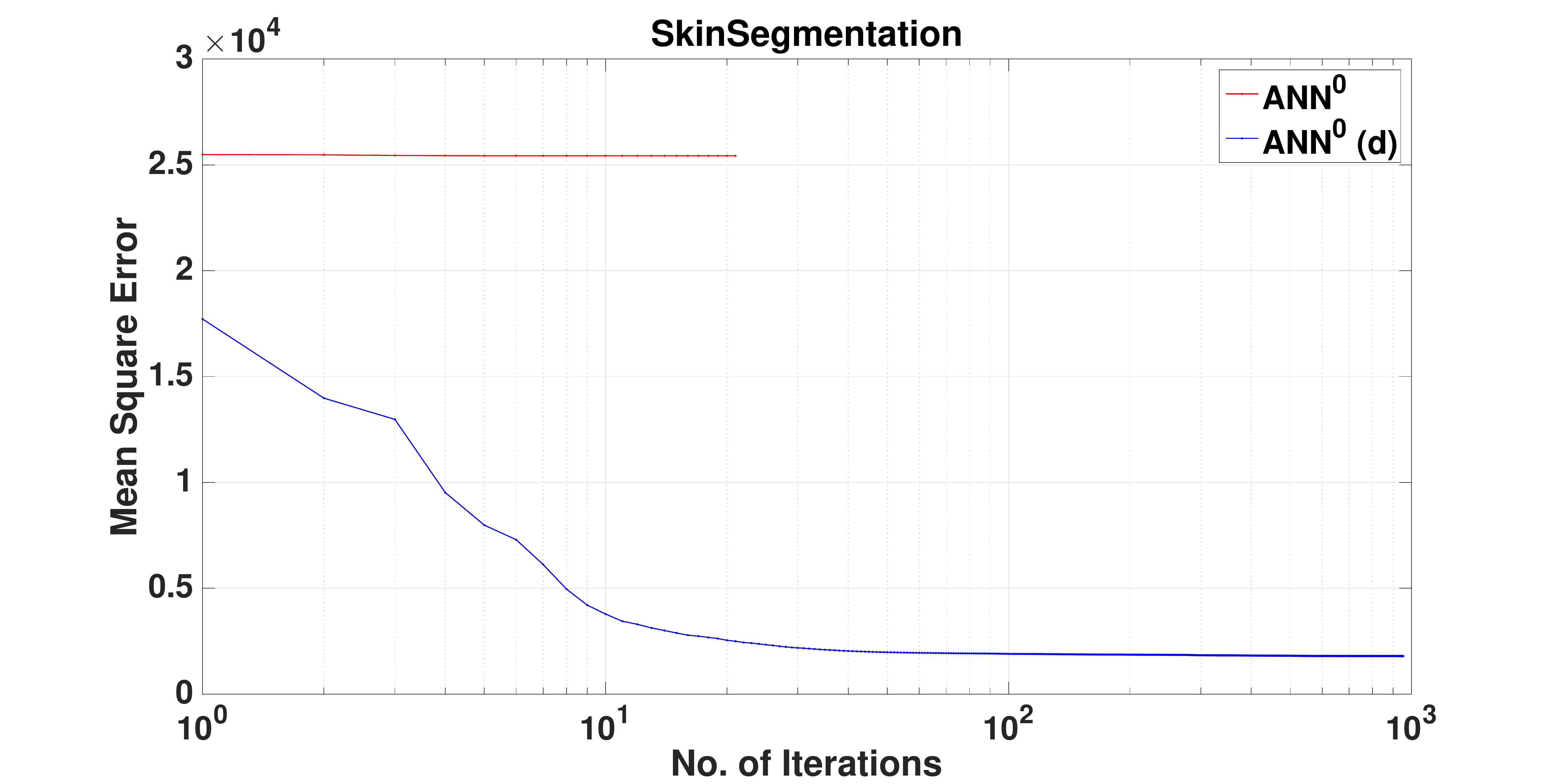}

\includegraphics[width=50mm,height=40mm]{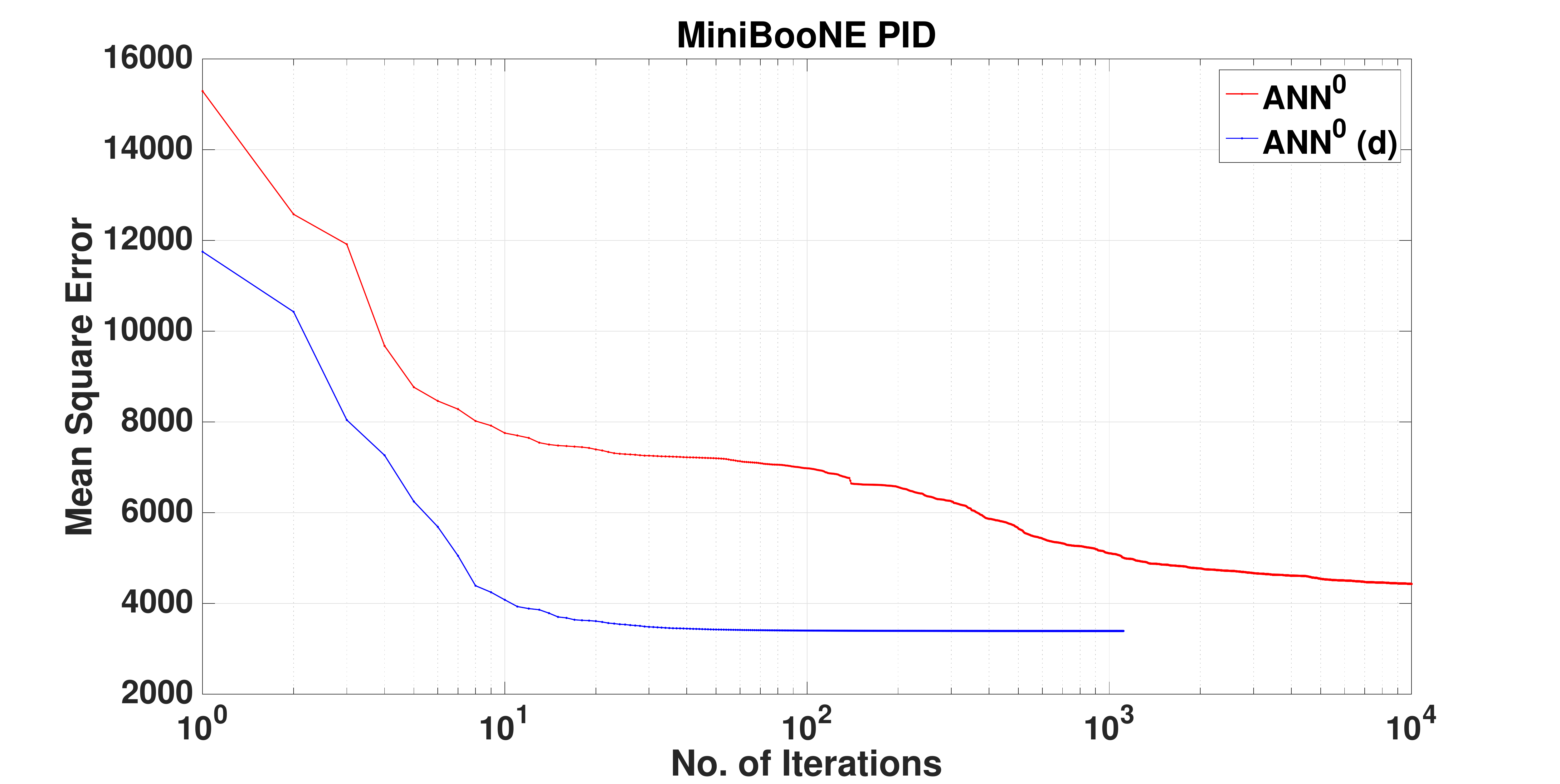}
\includegraphics[width=50mm,height=40mm]{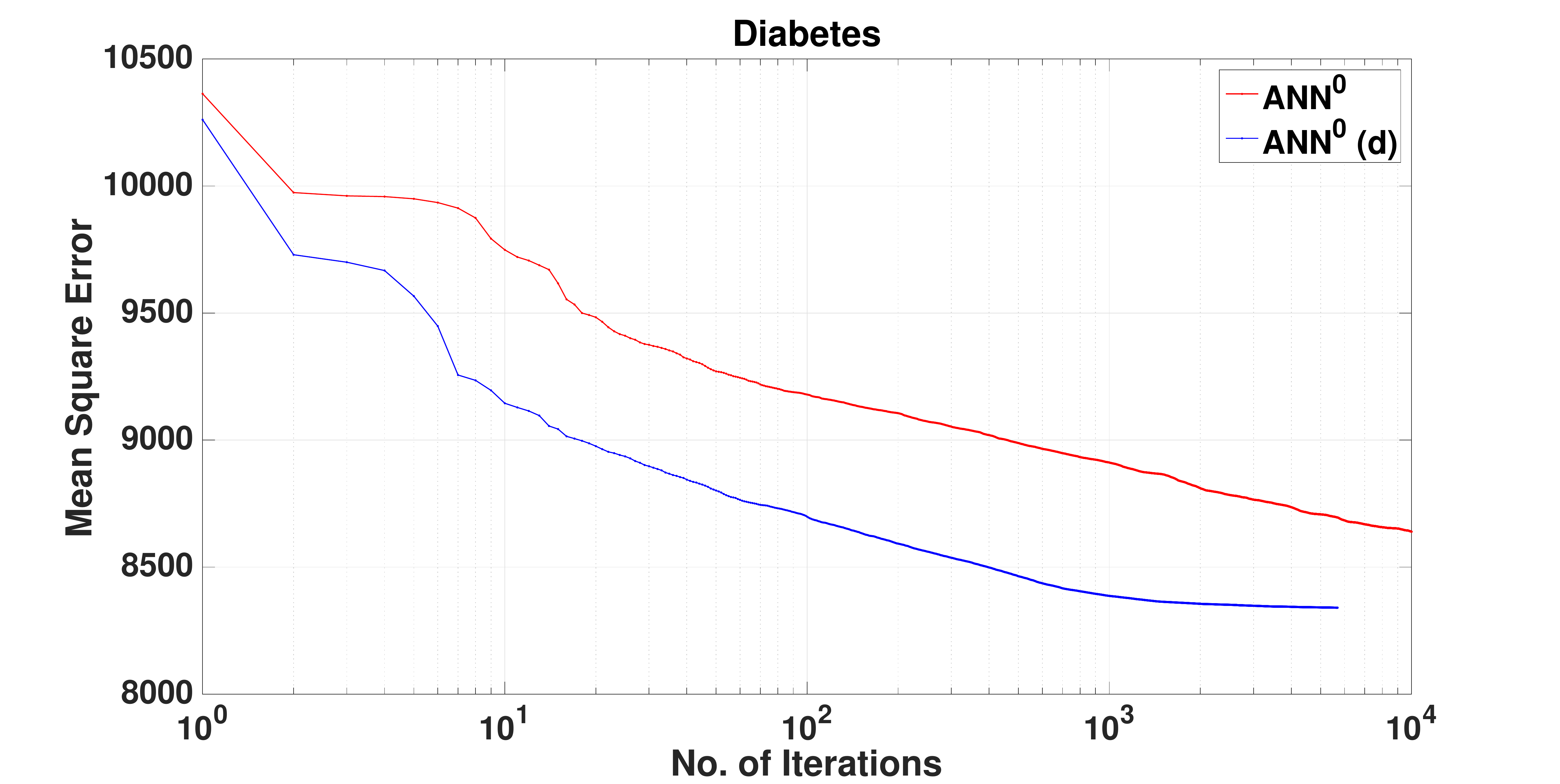}
\includegraphics[width=50mm,height=40mm]{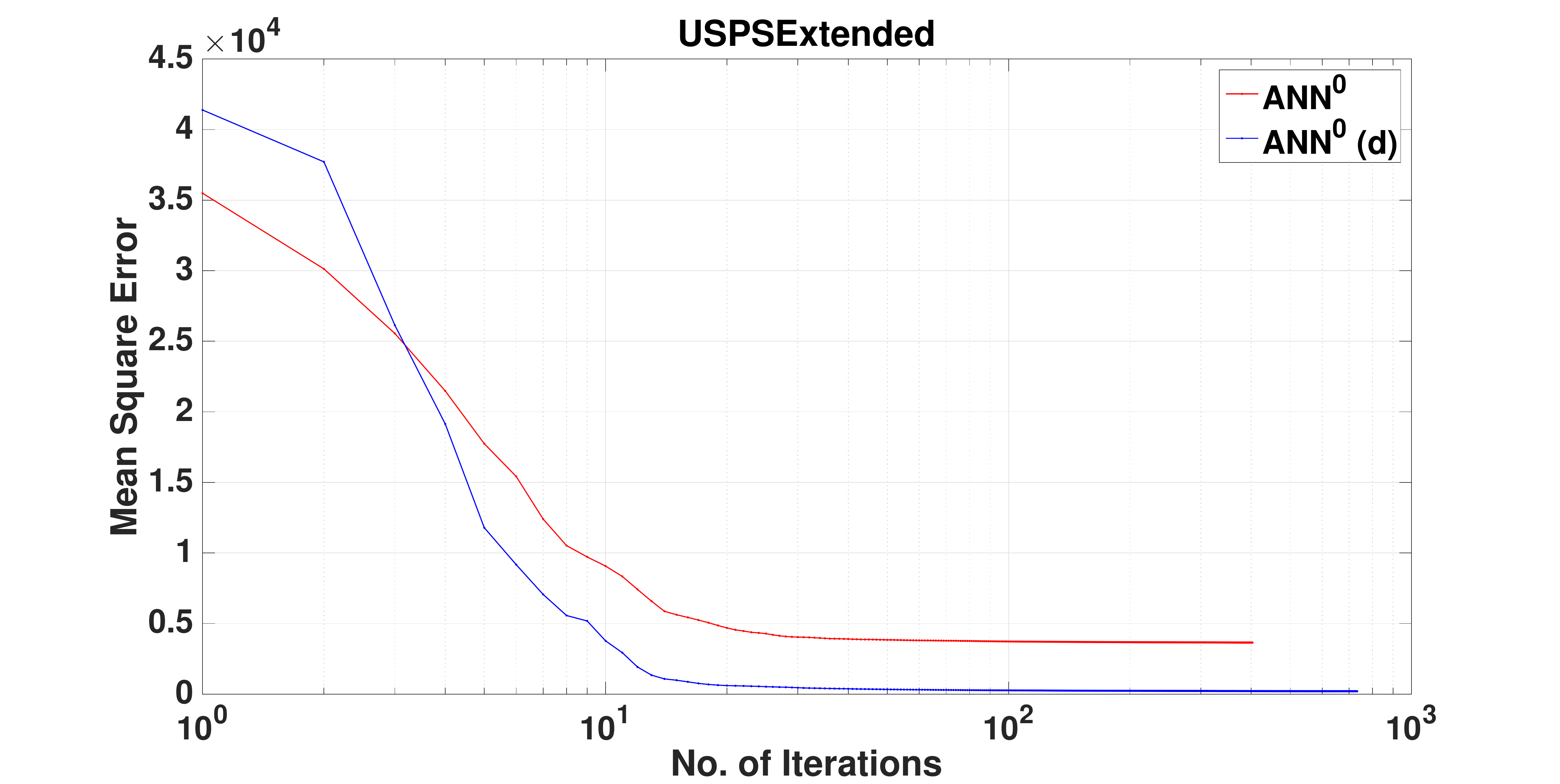}

\caption{\small Comparison of rate of convergence of $\ANN$ and $\ANNd$ on nine sample datasets. The X-axis is on the log scale.}
\label{fig_CC_MSE}
\end{figure}

Figures~\ref{fig_CC_HL} shows the variation in HL for SVC and SVC(d) whereas, Figure~\ref{fig_CC_MSE} shows the variation in MSE for $\ANNd$ and $\ANN$. A similar trend to NLL can be seen, that is SVC(d) and $\ANNd$ leading to a better value of the objective function while converging more rapidly.

\subsection{Comparison of the Learning and Classification Time $\dLC$ and $\LC$}

\begin{figure}[t] 
\centering
\hspace{-0.20in}
{\includegraphics[width=55mm,height=55mm]{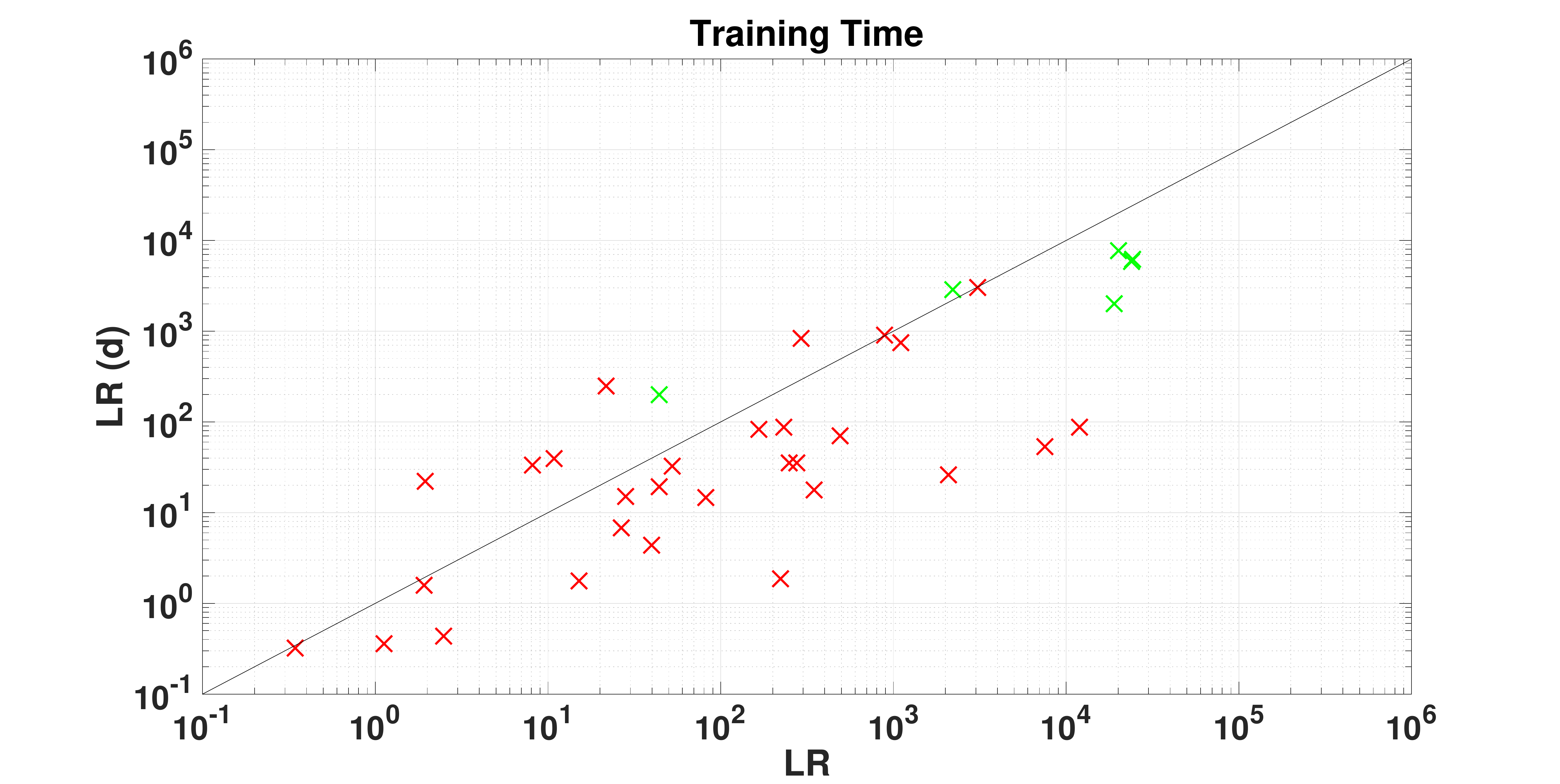}}\hspace{-0.20in}
{\includegraphics[width=55mm,height=55mm]{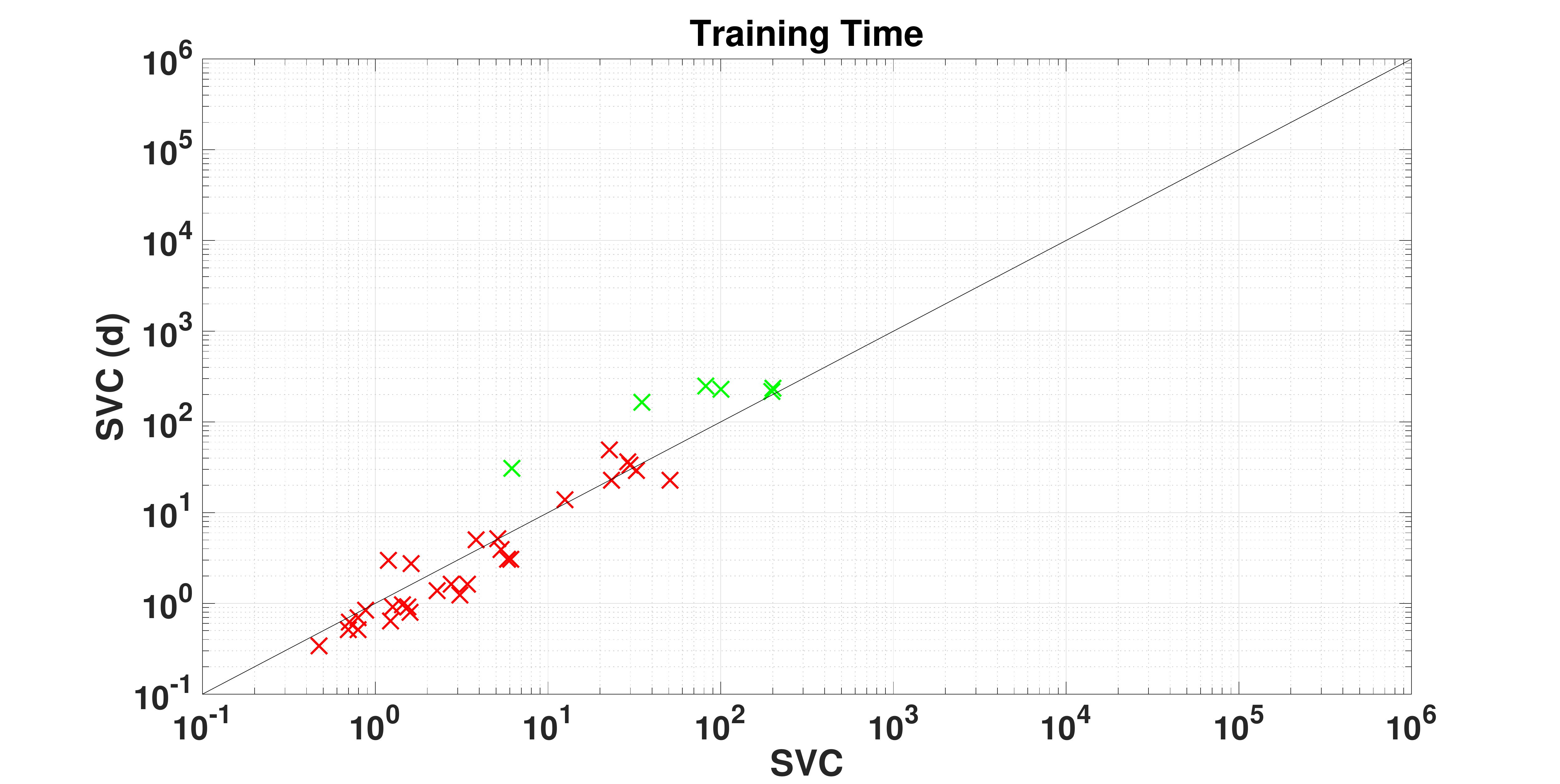}}\hspace{-0.20in}
{\includegraphics[width=55mm,height=55mm]{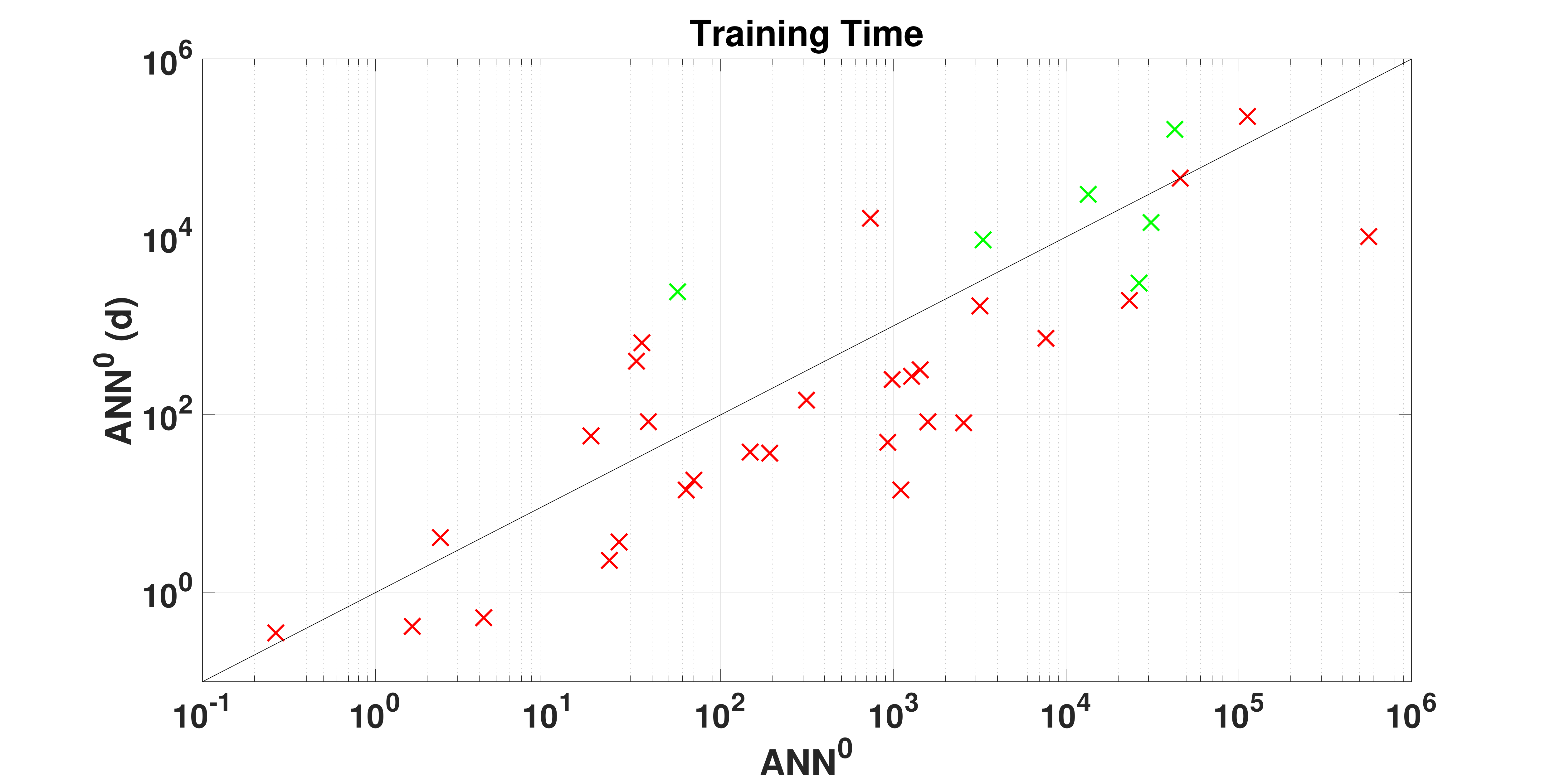}}\hspace{-0.20in}

\vspace{-0.15in}
\caption{\small 
Comparative scatter of \texttt{Training-time} results for $\LC$ and $\dLC$ classifiers. $\LC$ are on the X-axis whereas $\dLC$ are on the Y-axis. 
For points below the diagonal line, $\dLC$ is faster. 
Results on $Big$ datasets are shown in green, whereas results on $Little$ datasets are shown in red.}
\label{fig_LRvsoLR_train}
\end{figure}

\begin{figure}[t] 
\centering
\hspace{-0.20in}
{\includegraphics[width=55mm,height=55mm]{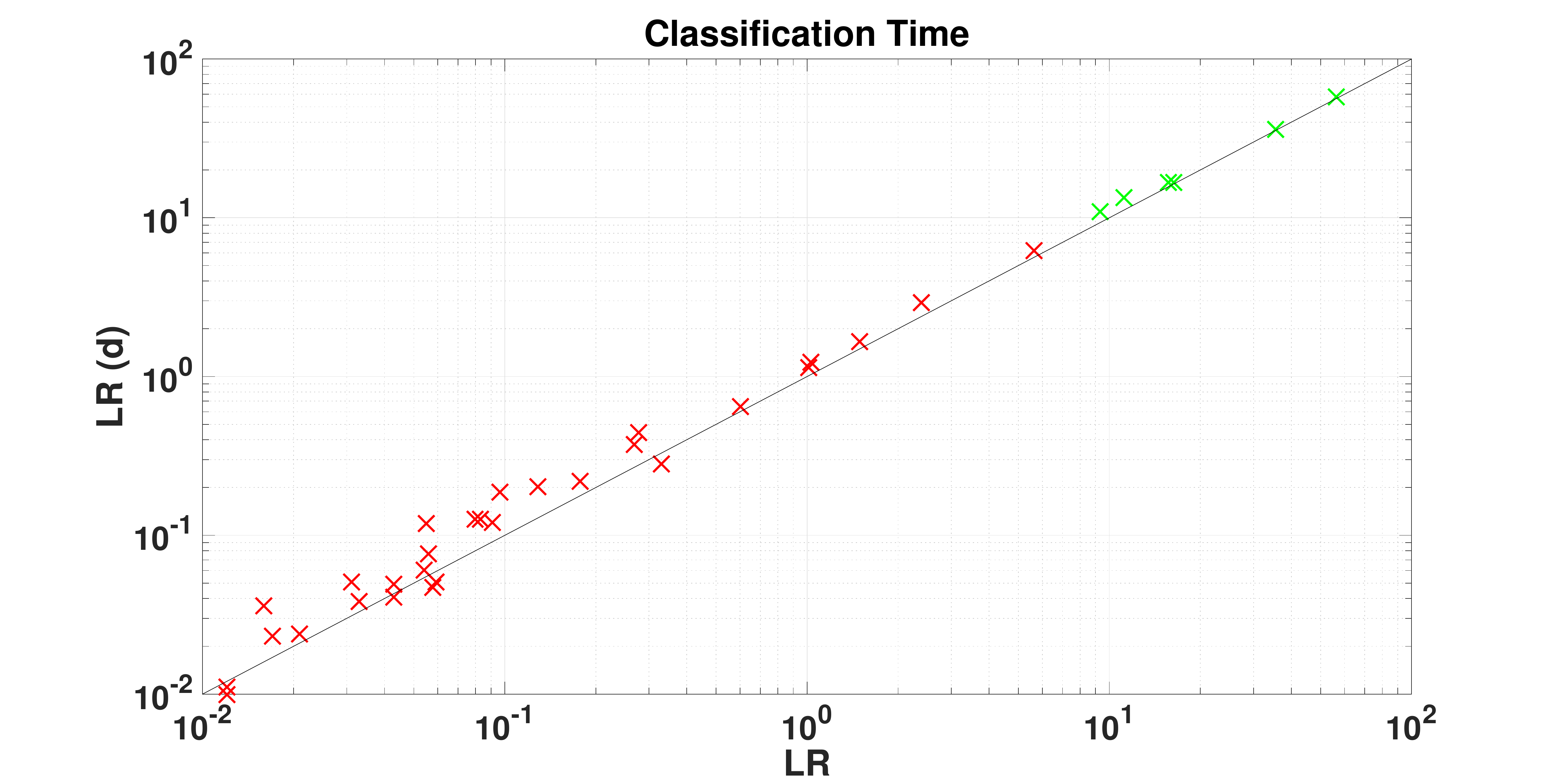}}\hspace{-0.20in}
{\includegraphics[width=55mm,height=55mm]{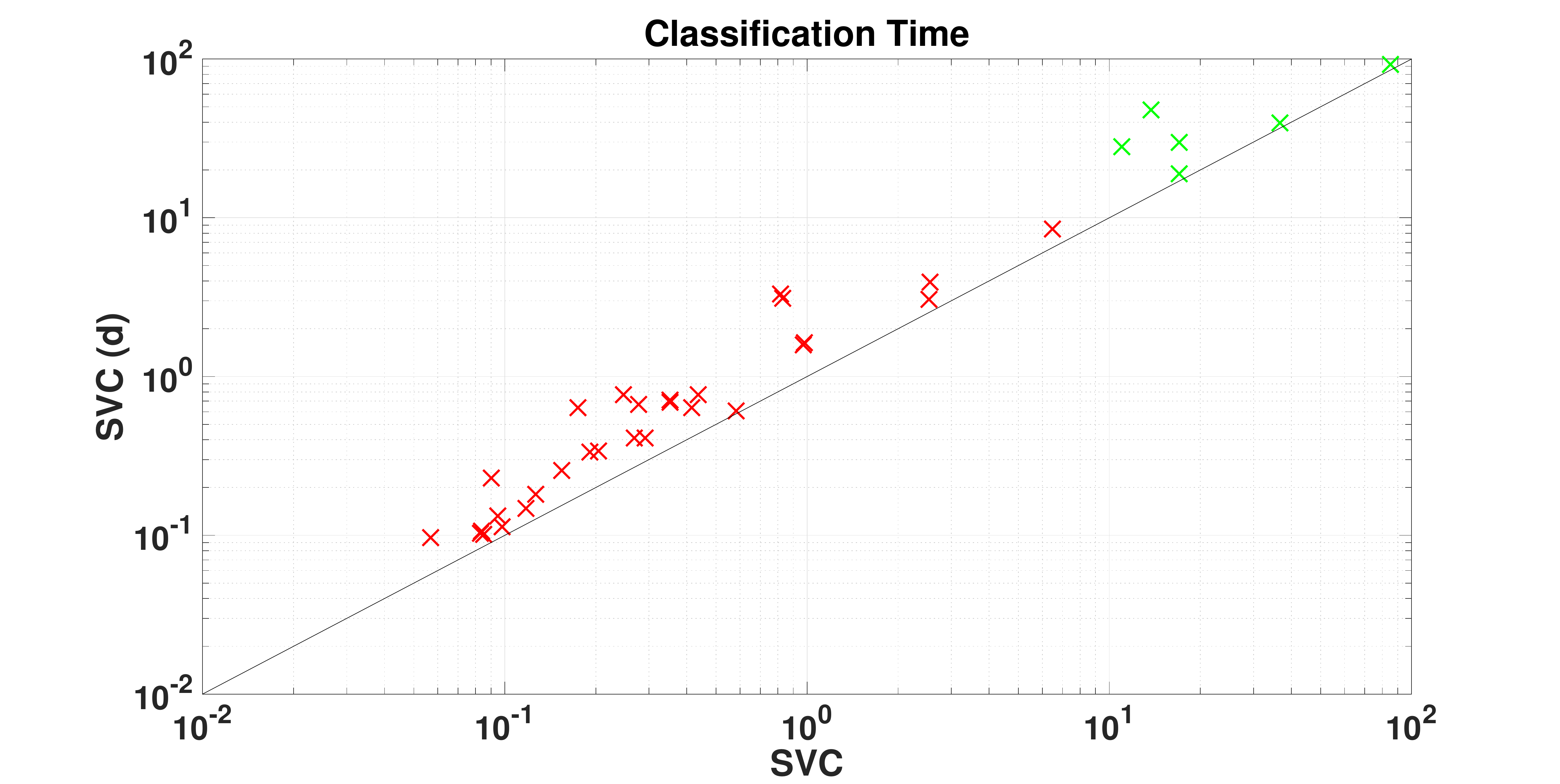}}\hspace{-0.20in}
{\includegraphics[width=55mm,height=55mm]{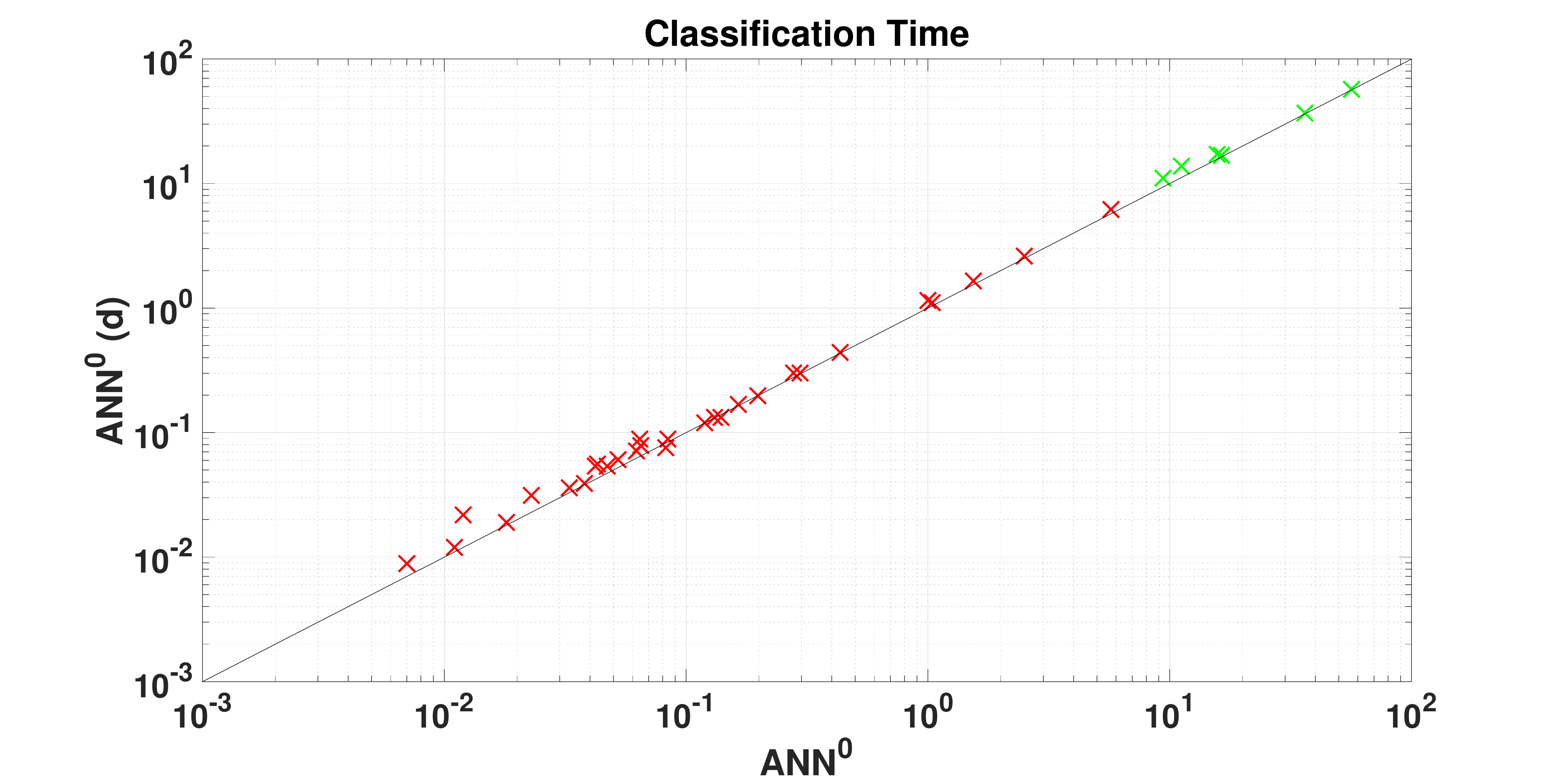}}\hspace{-0.20in}

\vspace{-0.15in}
\caption{\small 
Comparative scatter of \texttt{Classification-time} results for $\LC$ and $\dLC$ classifiers. $\LC$ are on the X-axis whereas $\dLC$ are on the Y-axis. 
For points below the diagonal line, $\dLC$ is faster. 
Results on $Big$ datasets are shown in green, whereas results on $Little$ datasets are shown in red.}
\label{fig_LRvsoLR_test}
\end{figure}

In this section, we compare the training and classification time of $\dLC$ and $\LC$. 
It can be seen from Figure~\ref{fig_LRvsoLR_train} that LR(d) and $\ANNd$ are slightly faster than LR and $\ANN$ respectively (majority of points below the diagonal line), whereas SVC(d) and SVC have similar training-time profiles. 
We already have seen the superior classification performance of $\dLC$ classifiers. These training-time results are extremely encouraging as they suggest that $\dLC$ can result in much better classification accuracy without compromising computational performance. 
We have reported the results only on four big datasets. This is because, the results were obtained by running the jobs on a local-desktop computer (i.e., a controlled set-up), rather than the heterogeneous cluster-computing environment in which most of the experimentation was performed. 
The scatter plots of classification time results for $\LC$ and $\dLC$ are presented in Figure~\ref{fig_LRvsoLR_test}. It can be seen that $\LC$ has slightly faster classification time than $\dLC$. However, in most cases the difference in the magnitude is small.

\section{Conclusion and Future Works} \label{sec_conclusions}

In this paper, we study the role of discretization for linear classifiers in machine learning. Current practice is primarily to apply discretization only when the learner requires  qualitative data. Overall, there exists some aversion to discretization as it loses information.
We argue that discretization -- despite losing information, can help  model non-linear relationships in the data and, therefore, can help reduce the bias of a learner that uses linear models. 
A linear classifier trained on discretized data is not linear any more, which has the potential to help in modeling non-linear decision boundaries which might otherwise require the use of kernels and multi-layer networks.

We show that discretization can greatly reduce the error of logistic regression and other discriminative linear classifiers optimizing Hinge Loss and Mean-square-error especially on large datasets. 
We compare the performance of linear classifiers trained with both qualitative and quantitative attributes (denoted as $\LC$) with LR trained with qualitative attributes only (denoted as $\dLC$), where quantitative attributes were discretized first. 
Our empirical analysis on $52$ datasets showed that $\dLC$ led to a low-bias model and, therefore, it resulted in significantly better 0-1 Loss and RMSE performance on large datasets. Quite surprisingly, it also reduced training time and had more desirable convergence, converging more rapidly to models that better fit the data.  These substantial benefits come at a cost of a minor increase in classification time.

Given the surprising gains from discretization, it is tempting to include both the original quantitative and derived discretized features in the data. Doing so avoids losing any information due to discretization. We undertook some preliminary experiments with this approach. They suggested that while it led to slight lower bias, they did not produce any improvement (in terms of error or convergence) over using only discretized-quantitative features. Further investigation of this research direction has been left as a future work.
 
With faster training, better convergence and low-bias we believe that discretization is worth consideration in any context where linear classifiers are learned from quantitative data.

\section{Code} \label{sec_code}

The details of the software library \texttt{fastLC} is given in Appendix~\ref{app_fastLC}. The library along with running instructions can be downloaded from Github: \url{https://github.com/nayyarzaidi/fastLC.git}.

\section{Acknowledgments} \label{sec_ack}

This research has been supported by the Australian Research Council (ARC) under grant DP140100087, and by the Asian Office of Aerospace Research and Development, Air Force Office of Scientific Research under contract FA2386-15-1-4007.

\appendix 
\section{\texttt{fastLC} -- Linear Classifiers Library} \label{app_fastLC}

The library can handle both quantitative and qualitative attributes. There is no need to do a one-hot-encoding for qualitative attributes, as the LR model built can actually handle the data types.

\sloppy
One can execute the code in the library by issuing the following command for LR:

\texttt{>> java -cp /fastLC.jar fastLC.BVDcrossvalx -t /dataset.arff -i 2 -x 2 -W LR.LRClassifier --  -V -O "Tron"}. For SVC, use the following command:

\texttt{>> java -cp /fastLC.jar fastLC.BVDcrossvalx -t /dataset.arff -i 2 -x 2 -W SVC.SVCClassifier -- -V -O "Tron"}. For $\ANN$, use the following command:

\texttt{>>java -cp /fastLC.jar fastLC.BVDcrossvalx -t /dataset.arff -i 2 -x 2 -W ANN.ANNClassifier -- -V -O "Tron"}.

Note, \texttt{-i 2 -x 2} flags specify two rounds of two-fold cross-validation.
 \texttt{-V} is the verbosity flag, whereas, \texttt{-O} specifies the solver. One can choose from the following list of solvers: $\{\textrm{GD}, \textrm{QN}, \textrm{CG}, \textrm{Tron}, \textrm{SGD}\}$, that is -- gradient descent, conjugate gradient, truncated Newton and stochastic gradient descent.

For SVC, the dataset has to be binary (i.e, the number of classes are only two). For non-binary dataset use the following command:

\texttt{>> java -cp /fastLC.jar fastLC.BVDcrossvalx -t /dataset.arff -i 2 -x 2 -W onevsAllSVCclassifier -- -V -O "Tron"}.


For computing results on discretized data, either pre discretize the dataset, or use the \texttt{-D} flag to convert quantitative attributes into qualitative one by the learner.

\section{Details of Datasets} \label{app_datasets}

\tiny

\tabulinesep = 1mm
\begin{center}
\begin{longtabu} to 140mm {|X[2.3c]|X[1c]|X[0.5c]|X[0.5c]|X[0.5c]|X[0.5c]|X[3.5c]|}

\everyrow{\tabucline{1-} }
\hline 
\multicolumn{1}{|c|}{\textbf{Domain}} & 
\multicolumn{1}{c|}{\textbf{Case}} & 
\multicolumn{1}{c|}{\textbf{NomAtt}} & 
\multicolumn{1}{c|}{\textbf{NumAtt}} & 
\multicolumn{1}{c|}{\textbf{Class}} & 
\multicolumn{1}{c|}{\textbf{MissVal}} & 
\multicolumn{1}{c|}{\textbf{Description}} \\ 
\hline 
\endfirsthead

\multicolumn{7}{c}%
{{\bfseries \tablename\ \thetable{} -- Continued from previous page}} \\
\hline 
\multicolumn{1}{|c|}{\textbf{Domain}} & 
\multicolumn{1}{c|}{\textbf{Case}} & 
\multicolumn{1}{c|}{\textbf{NomAtt}} & 
\multicolumn{1}{c|}{\textbf{NumAtt}} & 
\multicolumn{1}{c|}{\textbf{Class}} & 
\multicolumn{1}{c|}{\textbf{MissVal}} & 
\multicolumn{1}{c|}{\textbf{Description}} \\ 
\hline 
\endhead

\hline \multicolumn{7}{|r|}{{Continued on next page}} \\ \hline
\endfoot

\hline
\endlastfoot

\hline
HIGGS 		& 11000000 & 0     & 28    & 2     & N     & This dataset is generated using Monte Carlo simulation, related to separating particle-producing collisions from a background source, including 21 kinematic properties and 7 high level attributes.  \\

HEPMASS 		& 10500000 & 0     & 27    & 2     & N     & This dataset is generated using Monte Carlo simulation, representing information about separating particle-producing collisions from a background source, including 22 low level attributes and 5 high level attributes. \\

kddcup 		& 5209460 & 7     & 34    & 40    & N     & This dataset used for the Third International Knowledge Discovery and Data Mining Tools Competition \\
SUSY  		& 5000000 & 0     & 18    & 2     & N     & This dataset is generated using Monte Carlo simulation, representing information about separating particle-producing collisions from a background source, including 8 low level attributes and 10 high level attributes. \\
Watch\_accelerometer & 3540962 & 0     & 3     & 7     & N     & This dataset contains the readings of Accelerometer in smart watches. Two time related attributes are removed from the original dataset.\\
Watch\_gyroscope & 3205431 & 0     & 3     & 7     & N     & This dataset contains the readings of Gyroscope in smart watches. Two time related attributes are removed from the original dataset. \\
Phone\_gyroscope(40\%) & 2786526 & 0     & 3     & 7     & N     & The dataset contains the readings of Gyroscope in smartphones. Two time related attributes are removed from the original dataset. \\
Phone\_accelerometer(40\%) & 2612495 & 0     & 3     & 7     & N     & The dataset contains the readings of Accelerometer in smartphones. Two time related attributes are removed from the original dataset. \\
satellites(25\%) & 2176290 & 0     & 138   & 24    & Y     & This dataset describes satellite image time series. The classes are colours representing various land covers. \\
PAMAP2(25\%) & 962626 & 1     & 53    & 19    & N     & The dataset is generated by nine subjects wearing three colibri wireless activity sensors and one heart rate monitor. Timestamp attributes are removed from the original dataset.\\
MITFaceSetC & 839330 & 0     & 361   & 2     & N     & Each face in MITFaceSetB is rotated between $???20^{\circ}$ and $20^{\circ}$, in increments of $2^{\circ}$.\\
covertype & 581012 & 44    & 10    & 7     & N     & This dataset describes information about using cartograpic variables to identify forest cover type. \\
MITFaceSetB & 489410 & 0     & 361   & 2     & N     & This dataset is generated by making the original training face blurred and then adding to MITFaceSetA.\\
MITFaceSetA & 474101 & 0     & 361   & 2     & N     & This dataset is composed of 477366 new non-faces and the original MIT face dataset.\\
USPESExtended & 341462 & 0     & 675   & 2     & N     & The original training set has 1,005 zeros and 1,194 ones, while the test set has 359 zeros and 264 ones.  To better study the scaling behavior, we extend this data set by first converting the resolution from 16??16 to 26??26, and then generate new images by shifting the original ones in all directions for up to five pixels. \\
census-income(KDD) & 299285 & 32    & 9     & 2     & N     & This data set contains weighted census data extracted from the 1994 and 1995 Current Population Surveys conducted by the U.S. Census Bureau. The data contains 41 demographic and employment related variables \\
SkinSegmentation & 245057 & 0     & 3     & 2     & N     & The skin dataset is collected by randomly sampling B,G,R values from face images of various age groups, race groups, and genders obtained from FERET database and PAL database.  \\
WearableComputing & 165632 & 2     & 15    & 5     & N     & This dataset contains 5 different activities, gathered from 4 subjects wearing accelerometers mounted on their waist, left thigh, right arm, and right ankle. \\
localization & 164860 & 2     & 3     & 11    & N     & People used for recording of the data were wearing four tags (ankle left, ankle right, belt and chest). Each instance is a localization data for one of the tags. The tag can be identified by one of the attributes. \\
TwitterAbsoluteSigma500 & 140607 & 0     & 76    & 2     & N     & The objective of this dataset is to determine whether or not these time-windows are followed by buzz events. In this dataset, each instance covers seven days of observation for a specific topic. Considering the couple day following this initial observation; If there is at least 500 additional active discussions by day then, the predicted attribute Buzz is True. \\
MiniBooNE\_PID & 130065 & 0     & 50    & 2     & N     & This dataset is used for classifying signal and background event based on 50 particle ID variables. \\
TVNewsChannelCommercial & 129685 & 0     & 4124  & 2     & N     & This dataset is for automatic identification of commercial blocks in news videos finds a lot of applications in the domain of television broadcast analysis and monitoring.  \\
Diabetes 		& 101766 & 52    & 9     & 3     & Y     & The dataset represents 10 years (1999-2008) of clinical care at 130 US hospitals and integrated delivery networks. It includes over 50 features representing patient and hospital outcomes.  \\
waveform 		& 100000 & 0     & 20    & 3     & N     &  UCI\\
shuttle 		& 58000 & 0     & 9     & 7     & N     &  The dataset describes shuttle status log. \\ 
adult 			& 48842 & 8     & 6     & 2     & N     &  The dataset is for predicting if a person can make over 50K a year.\\
letter-recog 	& 20000 & 0     & 16    & 24    & N     &  The dataset is for predicting 26 capital English alphabet letters.\\
magic 		& 19020 & 0     & 10    & 2     & N     &  The dataset is generated by Monte Carlo simulation, for signal and background event classification using images from a Cherenkov gamma telescope. \\
sign  			& 12546 & 0     & 8     & 3     & N     &  UCI\\
pendigits 		& 10992 & 0     & 16    & 10    & N     & A pen-based handwritten digits dataset, collected from UCI. \\
pioneer 		& 9150  & 7     & 29    & 57    & N     &  UCI\\
satellite 		& 6435  & 0     & 35    & 6     & N     &  The dataset has records that describe different sub-areas of scenes converted from the satellite image.\\
optdigits 		& 5620  & 0     & 64    & 10    & N     &  This dataset is for predicting handwritten digits.\\
page-blocks 	& 5473  & 0     & 10    & 5     & N     & UCI \\
wall-following 	& 5456  & 0     & 24    & 4     & N     &  UCI\\
phoneme 		& 5438  & 0     & 7     & 50    & N     &  UCI\\
waveform-5000 	& 5000  & 0     & 40    & 3     & N     &  The dataset is collected from UCI\\ 
spambase 		& 4601  & 0     & 57    & 2     & N     & UCI \\
abalone 		& 4177  & 0     & 8     & 3     & N     &  UCI\\
segment 		& 2310  & 0     & 19    & 7     & N     &  UCI\\
mfeat-mor 		& 2000  & 2     & 3     & 10    & N     &  UCI\\
volcanoes 		& 1520  & 0     & 3     & 4     & N     &  UCI\\
yeast 			& 1484  & 1     & 7     & 10    & N     &  UCI\\
vowel 			& 990   & 3     & 10    & 11    & N     &  UCI\\
vowel-context 	& 990   & 1     & 10    & 11    & N     &  UCI\\
vehicle 		& 946   & 0     & 18    & 4     & N     &  UCI\\
anneal 		& 798   & 32    & 6     & 3     & N     &  UCI\\
pid   			& 768   & 0     & 8     & 2     & N     &  UCI\\
syncon 		& 600   & 0     & 60    & 6     & N     &  UCI\\
musk1 		& 476   & 0     & 166   & 2     & N     & UCI \\
new-thyroid 	& 215   & 0     & 5     & 3     & N     &  UCI\\
wine  			& 178   & 0     & 13    & 3     & N     &  UCI\\

\end{longtabu}
\end{center}

\normalsize

\bibliography{jmlr_subArXiv}

\end{document}